\documentclass{report}

\usepackage[textsize=tiny]{todonotes} 

\usepackage{footmisc}

\usepackage{graphicx}
\usepackage{dirtytalk}
\usepackage{url}
\usepackage{hyperref}
\usepackage{amsthm}
\usepackage{amsmath}
\usepackage{amssymb}
\usepackage{authblk}

\usepackage{times}
\usepackage{paralist}
\usepackage{color}

\definecolor{dkgreen}{rgb}{0,0.6,0}
\definecolor{gray}{rgb}{0.5,0.5,0.5}
\definecolor{mauve}{rgb}{0.58,0,0.82}

\newcommand{\vader}{VAD$^2$ER}

\usepackage{listings}
\usepackage{algorithm}
\usepackage{algpseudocode}

\usepackage[utf8]{inputenc}
\usepackage{graphicx}
\usepackage{listings}
\newcommand{\furl}[1]{\footnote{\scriptsize \url{#1}}}
\usepackage{amsmath}
\usepackage{mathtools}
\DeclarePairedDelimiterX\set[1]\lbrace\rbrace{#1}
\usepackage{comment}
\usepackage{todonotes}

\usepackage{commath}
\usepackage{caption}

\theoremstyle{definition}

\theoremstyle{definition}
\newtheorem{example}{Example}
\newcommand{\chapterauthor}[1]{%
  {\parindent0pt\vspace*{-25pt}%
  \linespread{1.1}\large\scshape#1%
  \par\nobreak\vspace*{35pt}}
  
}
\usepackage[backend=biber,sorting=nyt,bibencoding=latin1,abbreviate=false,mincrossrefs=3,style=numeric,maxnames=30]{biblatex}

\begin{document}

\begin{titlepage}

\newcommand{\HRule}{\rule{\linewidth}{0.5mm}} 

\center 
 

\includegraphics[scale=.3]{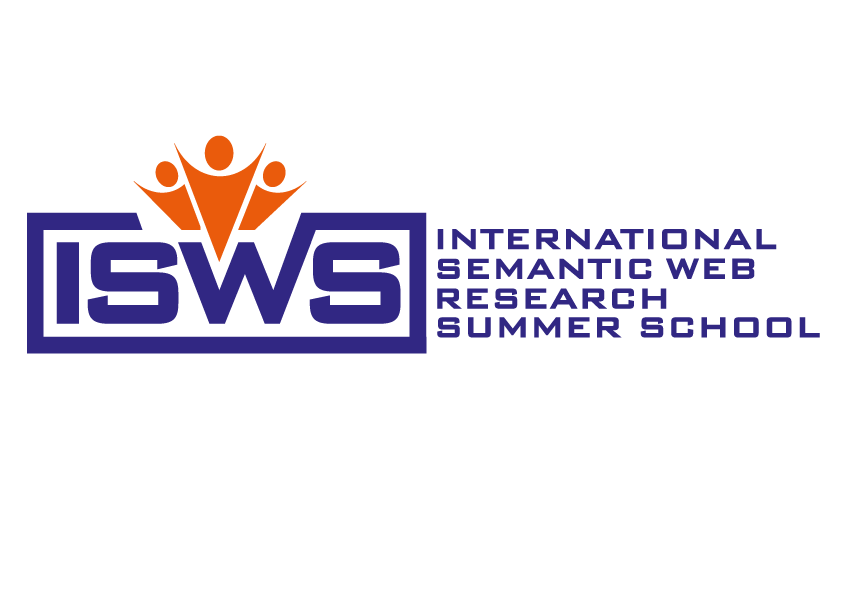}\\[1cm] 


\HRule \\[0.4cm]
{ \huge \bfseries Knowledge Graphs Evolution and Preservation}\\[0.2cm] A Technical Report from ISWS 2019\\[0.4cm]
\HRule \\[1.5cm]
 




{\large \today}\\[1cm] 
\large{Bertinoro, Italy}
\vfill 

\end{titlepage}

%
%
%
%
%

\chapter*{Authors}
\textbf{Main Editors}\\
Valentina Anita Carriero, Semantic Technology Lab, ISTC-CNR, IT\\
Luigi Asprino, Semantic Technology Lab, ISTC-CNR, IT\\
Russa Biswas, FIZ Karlsruhe\\



\noindent \textbf{Supervisors}\\
Claudia d'Amato, University of Bari, IT\\
Irene Celino, Cefriel, Milan, IT\\
John Domingue, KMi, Open University and 
President of STI International, UK\\
Michel Dumontier, Maastricht University, NL\\
Marieke van Erp, 
DHLab, KNAW Humanities Cluster, NL\\
Heiko Paulheim, University of Mannheim, DE\\
Axel Polleres, Vienna University of Economics and Business, AT\\
Valentina Presutti, Semantic Technology Lab, ISTC-CNR, Rome, IT\\
Sebastian Rudolph, TU Dresden, DE\\
Harald Sack, FIZ - Karlsruhe, Karlsruhe Insitute of Technology, AIFB, DE\\

\noindent \textbf{Students}\\
Nacira Abbas, INRIA\\
Kholoud Alghamdi, The University of Southampton\\
Mortaza Alinam, University of Genoa\\
Francesca Alloatti, University of Torino, CELI - Language Technology\\
Glenda Amaral, Free University of Bozen-Bolzano\\
Martin Beno, Vienna University of Economics and Business\\
Felix Bensmann, GESIS - Leibniz Institute for the Social Sciences\\
Ling Cai, University of California, Santa Barbara\\
Riley Capshaw, Link{\"o}ping University\\
Amine Dadoun, Eurecom\\
Stefano De Giorgis, University of Bologna\\
Harm Delva, Ghent University, IDLab, imec\\
Vincent Emonet, Institute of Data Science, Maastricht University\\
Paola Espinoza Arias, Universidad Polit{\'e}cnica de Madrid\\
Omaima Fallatah, University of Sheffield\\
Sebastián Ferrada, Institute for Foundational Research on Data, University of Chile\\
Marc Gallofr{\'e} Oca\~{n}a, University of Bergen\\
Michalis Georgiou, University of Huddersfield\\
Genet Asefa Gesese, FIZ Karlsruhe\\
Frances Gillis-Webber, University of Cape Town\\
Francesca Giovannetti, University of Bologna\\
Mar{\'i}a Granados Buey, everis/NTT Data\\
Ismail Harrando, Eurecom - Sorbonne University\\
Ivan Heibi, Digital Humanities Advanced Research Centre, University of Bologna\\
Vitor Horta, Dublin City University, Insight Centre\\
Laurine Huber, LORIA\\
Federico Igne, University of Oxford\\
Mohamad Yaser Jaradeh, L3S Research Center, Leibniz University Hannover\\
Neha Keshan, Rensselaer Polytechnic Institute\\
Aneta Koleva, TU Dresden, Siemens Munich\\
Bilal Koteich, French naval academy research institute\\
Kabul Kurniawan, TU - Wien\\
Mengya Liu, University of Southampton\\
Chuangtao Ma, E{\"o}tv{\"o}s Lor{\'a}nd University\\
Lientje Maas, Utrecht University\\
Martin Mansfield, The University of Liverpool, CSols Ltd.\\
Fabio Mariani, University of Bologna\\
Eleonora Marzi, University of Bologna\\
Sepideh Mesbah, Delft University of Technology\\
Maheshkumar Mistry, University of Koblenz (Institute of Web Science)\\
Alba Catalina Morales Tirado, The Open University\\
Anna Nguyen, Karlsruhe Institute of Technology\\
Viet Bach Nguyen, University of Economics, Prague\\
Allard Oelen, L3S Research Center, TIB Hannover, Leibniz University Hannover\\
Valentina Pasqual, University of Bologna\\
Margherita Porena, ICCD - Istituto centrale per il catalogo e la documentazione\\
Jan Portisch, SAP SE, University of Mannheim\\
Kader Pustu-Iren, L3S Research Center, TIB Hannover, Leibniz University Hannover\\
Ariam Rivas Mendez, L3S Research Center, Leibniz University Hannover\\
Soheil Roshankish, Universit{\`a} della Svizzera Italiana, Institute of Digital Technologies for Communication\\
Ahmad Sakor, L3S Research Center\\
Jaime Salas, University of Chile\\
Thomas Schleider, Eurecom\\
Meilin Shi, University of California, Santa Barbara\\
Gianmarco Spinaci, University of Bologna\\
Chang Sun, Institute of Data Science - Maastricht University\\
Tabea Tietz, FIZ Karlsruhe\\
Molka Tounsi Dhouib, University Cote d'Azur, INRIA, CNRS, I3S, Silex\\
Alessandro Umbrico, Institute of Cognitive Sciences and Technologies (ISTC-CNR)\\
Wouter van den Berg, TNO\\
Weiqin Xu, Universit{\'e} Paris Est Marne la Vall{\'e}e
\chapter*{Abstract}
One of the grand challenges discussed during the Dagstuhl Seminar ``Knowledge Graphs: New Directions for Knowledge Representation on the Semantic Web''~\cite{DBLP:journals/dagstuhl-reports/BonattiDPP18} and described in its report is that of a:
\begin{quote}
Public FAIR Knowledge Graph of Everything: We increasingly see the creation of knowledge graphs that capture information about the entirety of a class of entities. For example, Amazon is creating a knowledge graph of all products in the world and Google and Apple have both created knowledge graphs of all locations in the world. This grand challenge extends this further by asking if we can create a knowledge graph of “everything” ranging from common sense concepts to location based entities. This knowledge graph should be “open to the public” in a FAIR manner democratizing this mass amount of knowledge.\end{quote}

Although linked open data (LOD) is \emph{one} knowledge graph, it is the closest realisation (and probably the only one) to a public FAIR Knowledge Graph (KG) of everything. Surely, LOD provides a unique testbed for experimenting and evaluating research hypotheses on open and FAIR KG.

One of the most neglected FAIR issues about KGs is their ongoing evolution and long term preservation. We want to investigate this problem, that is to understand what preserving and supporting the evolution of KGs means and how these problems can be addressed. Clearly, the problem can be approached from different perspectives and may require the development of different approaches, including new theories, ontologies, metrics, strategies, procedures, etc.   

This document reports a collaborative effort performed by nine teams of students, each guided by a senior researcher as their mentor, attending the International Semantic Web Research School (ISWS 2019). Each team provides a different perspective to the problem of knowledge graph evolution substantiated by a set of research questions as the main subject of their investigation. In addition, they provide their working definition for KG preservation and evolution.


\tableofcontents
\listoffigures
\listoftables


\part{Machine learning techniques for detecting Knowledge Graphs evolution}
\label{part1}
\chapter{Towards an Automatic Detection of Evolution in Knowledge Graphs}
\label{sec:dragons}
\chapterauthor{Francesca Alloatti, Riley Capshaw, Molka Dhouib, Mar\'ia G. Buey, Ismail Harrando, Jaime Salas, Claudia d'Amato}

\section{Introduction}
\label{sec:intro-dragon}


A Knowledge Graph (KG) could be considered as a collection of interlinked descriptions of entities (e.g. objects, people, events, situations, or concepts) possibly enriched information 
acquired and integrated within ontologies~\cite{gruber1993translation} thus making possible to 
apply a reasoner and derive new knowledge~\cite{ehrlinger2016towards}. Several KGs are available, including DBpedia\footnote{https://wiki.dbpedia.org/}, Wikidata\footnote{https://www.wikidata.org/}, YAGO\footnote{https://www.mpi-inf.mpg.de/yago-naga/yago/}, or DBLP\footnote{https://dblp.uni-trier.de/}. 
These KGs may be continuously maintained by an open community, by adding new entities and relationships, or modifying the existing ones, 
in order to represent a constantly changing reality and a permanent flow of new knowledge. 
As a consequence, 
the conclusions 
that can be inferred may also 
change over time. 
An example is provided by DBpedia,  
releasing a new version each year as well as 
a snapshot each month.

Despite this need to constantly maintain KGs, another important problem is to understand the motivation(s) for (some) changes in a piece of knowledge and possibly how these changes should be interpreted. 
For instance, in the scholarly data domain (DBLP), KGs are often updated due to new publications, changes in the affiliation of authors, or the introduction of a new research area. By considering different snapshots of DBLP for a researcher, his/her evolution could be ideally traced. Similarly, the change and/or evolution of a research community (e.g. the semantic web community) over a collection of years could be detected (and possibly tracked), which may turn out useful to beginners and practitioners interested in knowing what are the current trends in a research field.


Our contribution focuses on a methodological proposal for capturing different types of evolution within a KG. Specifically, 
three different kinds of evolution are targeted: (i) \textbf{atomic evolution}, focusing on the analysis of atomic operations at resource level (entities and links), (ii) \textbf{local evolution}, studying the evolution of a resource within its community and (iii) \textbf{global evolution}, focusing on the detection of communities in the whole graph in order to also understand  
the general behaviour of the network. 
Therefore, the main research question that we focused on in this work is: \textit{Would it be possible to automatically capture evolution within KGs?}
To try answering this question, we addressed the following sub-questions:
\begin{enumerate}
    \item How can we automatically capture this evolution?
    \item Which Machine Learning (ML) approach can be used to achieve that? Are ML existing solutions sufficient for addressing our goal?
\end{enumerate}

As a use case, we focused on exploring scholarly data 
but we also show how 
the envisioned methodology can be generalized to other domains. 


The rest of the paper is organized as follows. Section~\ref{sec:approach-dragon} presents our proposed approach to detect the three types of evolution on a KG that we have defined. Section~\ref{sec:evaluation-dragon} describes the 
evaluation protocol. Section~\ref{sec:related-dragon} 
overviews 
related approaches. Section~\ref{sec:conclusions-dragon} draws 
conclusions and discuss our perspectives for these challenges.


\section{Proposed approach}
\label{sec:approach-dragon}



Evolution in KGs is an open research direction that can be approached from different perspectives. 
We focus on capturing evolution at three different levels: \textit{atomic}, \textit{local}, and \textit{global} by tracking these changes in different snapshots of a KG. 

Given the scale of KGs we have usually to deal with, we leverage unsupervised and  graph-mining methods for our analysis. Nevertheless, graph-mining methods mostly focus on the graph structure, that is they focus on the graph in terms of typeless nodes and edges. However, as briefly discussed in Section~\ref{sec:intro-dragon}, KGs are usually endowed with ontologies acting as background knowledge. 
In order to exploit the additional semantics therein, background knowledge about the content of the graph can be introduced by querying specific subsets of the KG that carry a homogeneous meaning and then use the proposed methods on them. It is also worth mentioning that since our goal is to automate the process of capturing KG evolution, we only focus on the data within the KG, while 
the evolution of the meta-data (the ontology/scheme) is out of our scope. 

In the following, we first present our hypotheses, hence 
we illustrate our proposed methodology to address the evolution in KGs.

\subsection{Hypotheses}
\begin{itemize}
	\item We assume that KGs have a community structure\footnote{\label{footnote1}https://en.wikipedia.org/wiki/Community\_structure}, meaning that related resources tend to fall into highly linked structures that we call \textit{communities}.
	\item Evolution in KG can be observed at several levels of granularity. We theorize that there are three such levels: atomic (evolution on a \textit{resource} level), local (evolution of a resource 
	within its \textit{community}), and global (evolution of \textit{communities} in the whole knowledge graph). 
	\item We may be able to understand the evolution of a KG by extracting some features on each level of abstraction, both explicit ones (e.g. number of resources of a certain type), and latent ones (e.g. connected of a community features).
\end{itemize}

\subsection{Overview}
We surmise that evolution in a knowledge graph can manifest on multiple levels: 
\begin{itemize}
	\item \textbf{Atomic:} evolution on a resource level (entities and links). This refers to the atomic transactions performed at resource level thus making the KG evolving over time, i.e. insertions, updates and deletions of entities and/or links. 
	\item \textbf{Local:} evolution on a community level. We borrow the notion of community from Graph Theory as a ``set of nodes which is densely connected internally''\footref{footnote1}. This refers to detecting when a set of atomic transactions within a community determines an actual and significant change/evolution of the community itself. 
	\item \textbf{Global:} evolution on the whole graph, which is observed through the evolution of communities, such as the emergence of a new community, the splitting of some community etc. 
\end{itemize}

Our hypothesis is that each kind of evolution can be detected by collecting multiple snapshots of a KG. 
%
Precisely, a \textit{snapshot} of a KG is a dump of all or a part of its content at a given timestep. Given two snapshots $S_i$ and $S_j$ of some Knowledge Graph such that $i < j$, we define the following sets:
\begin{align*}
    A_{ij} &= \left\{triplet\ t_k \in S_j \mid t_k \notin S_i \right\} \\
    D_{ij} &= \left\{triplet\ t_k \in S_i \mid t_k \notin S_j \right\} \\
    U_{ij} &= A_{ij} \cup D_{ij}
\end{align*}

\noindent
where $A_{ij}$ denotes the set of atomic additions performed between snapshot $S_i$  and $S_j$
and $D_{ij}$ denotes the set of deletions done between snapshots $S_i$  and $S_j$. 
Modification operations will be considered as a sequence of a deletion of the original triplet and an addition of the modified version.  
We 
group all updates 
into a single group: $U_{ij}$. $A_{ij}$ and $D_{ij}$ 
materialize the atomic evolution of the graph through time. 

However, it is arguably more interesting to capture the more implicit aspects of evolution, i.e. the ones that are not expressed in terms of entities or relations but rather in the underlying structure of the graph, such as the apparition of a new community. 
To do so, we define "communities" as an intermediate level between the resource and the graph level. A community 
is subset of highly linked 
nodes within the graph.

\todo[disable]{Figure missing}

\subsection{Atomic Evolution}
\label{atomic}

At this level, we rely on the set of updates $U_{ij}$, which contains the atomic operations performed between two snapshots $S_i$ and $S_j$, to extract a first-order idea 
of the evolution of the KG. This is manifested in two aspects:
\begin{itemize}
    \item \textbf{Evolution description:} it quantifies the detected changes between the two snapshots. It can be expressed as a statistical analysis of the observed updates as an \textit{expectation of the evolution}. The simplest way to model these changes is to assume a normal distribution and quantify each feature we want to track in terms of its \textit{mean} and \textit{variance}. The features on a resource level can be related to:  resources types (e.g. number of inserted/deleted resources of a given type $T$), relations (e.g. number of insertion/deletions of a property $p$), and resource-relation counting (e.g. number of insertion/deletions of a property $p$ for each resource of type $T$). 
    In the considered scholarly domain, we can quantify the average number of publications by an author, or the number of citation a paper gets, etc.
    \item \textbf{Noteworthy changes:} given the \emph{evolution description}, 
    a statistical analysis can be performed to recognize potentially meaningful changes that happened between two snapshots e.g. by identifying the features which diverge from the expected evolution, 
    given a 
    threshold that would depend on the studied KG.
\end{itemize}

Formally, given $F_{r, i}$ as the vector of features of some resource $r$ in a snapshot $S_i$, $\mu$ and $\delta^2$ as the mean and variance respectively, for the change between two snapshots in those features across the knowledge graph and a sensitivity threshold $\theta$, we flag a resource as \textit{noteworthy} iff:

\begin{equation}
    p(\Delta_{i,j}F_r | \mu, \delta^2) < \theta
\end{equation}

For instance, 
in the scholarly domain, a publication can be flagged as noteworthy if it received an exceptionally high number of new citations, or a conference getting many new submissions, etc.

For more efficiency, the features should be computed from the constructed set of updates $U_{ij}$, disregarding the static part of our KG. 

\subsection{Local Evolution}
\label{sec:localEvolution}

As for local evolution, i.e. studying the evolution of a resource within its community, 
we compute, for each resource, the same features as illustrated in Section~\ref{atomic}, 
but instead of considering the whole graph, 
the community to which the resource belongs to is solely taken into account for computing the expected statisics ($\mu_i$ and $\delta_i^2$ for each community $c_i$) and identifying any \textit{noteworthy} resource. 
E.g. in the scholarly domain, a publication may not be that \textit{noteworthy} with respect to the whole Scholarly KG, but it stands out within its community (considering that a community may be a subfield of research). This is due to the fact that the expected evolution statistics can be vary significantly across communities. 

Importantly, at this level we may also consider community level features such as \textit{graph density}, number of connected nodes, 
etc, and compute the statistics to identify \textit{noteworthy} communities in a similar fashion as we did with the resources within the graph.

It is also worth noting that we can carry over this local analysis as many times as deemed fit by defining multiple levels of abstraction, i.e. by repeating the process of subdividing the communities into sub-communities and study their evolution on increasingly lower granularities e.g. fields, subfields, research groups, etc.

\subsection{Global Evolution}
The detection of communities is an important task for the analysis of networks, because it may provide insight on the general behaviour of the different entities that belong to that network. 
KGs can be considered a type of (multi-)community structure, where entities are connected to each other based on their (multiple) relations. 
Because of this, we are bound to find communities of nodes that are more densely connected by some metric. Furthermore, 
(multiple) updates on individual entities may affect the communities they belong to. Since we are interested in the evolution of a KG and whether the changes brought by the evolution are significant, we aim at identifying 
 communities in the KG in order to analyse the impact of 
changes on their entities.

Several algorithms have been proposed to detect communities in networks. The study of Porter et al. \cite{porter2009communities} provides a comparison of a variety of community detection algorithms, including their advantages and disadvantages, emphasizing 
the effectiveness in different fields of study. 
We 
consider the Girvan-Newman Algorithm \cite{girvan2002community} 
as the most useful one for our purpose since 
it is based on the \emph{betweenness} of nodes in a graph, and the analysis of Porter et al. \cite{porter2009communities} mentions betweenness-based approaches as the ones that yield the most intuitive communities\footnote{A possible drawback to the Girvan-Newman Algorithm is the fact that the algorithm may perform slowly on large graphs, and results may be poor if the graphs are too dense. However, since the scholarly KG we are mostly considering  
is not particularly dense, we believe it will work adequately}.

Betweenness is a notion of closeness in graphs that denotes the number of shortest paths that go through a certain node. Intuitively, the more times a node $n$ appears in the shortest path between nodes $n_i$ and $n_j$ in a graph, the closer node $n$ is to those nodes.

Given a community detection method $\kappa$ and a snapshot $S_i$, we define:
\[C_{i} = \left\{\text{ Communities } c_k \mid c_k \in \kappa(S_i) \right\}\,,\]
as the set of communities within the graph snapshot $S_i$, where $\kappa(S_i)$ denotes the computation of the method $\kappa$ over the snapshot $S_i$, whose output is a set of sub-graphs of $S_i$ called communities. 

Given two timesteps $i$ and $j$ such that $i < j$, and the sets of communities $C_i$ and $C_j$ generated from the snapshots $S_i$ and $S_j$, 
we aim at detect 
the following types of phenomena:
    a) Emergence / disappearance of a community;
    b) Merging / splitting of a community; 
    c) Persisting communities. 

Particularly, to detect persisting communities (communities which carry over between the snapshots), 
we measure the overlap (overlapping triplets, specifically in terms \textit{intersection over union}) 
of communities.
Formally, given a threshold
$\omega_{overlap}$, two communities $c_m \in C_i$ and $c_n \in C_j$ are persistent if: 
%
\begin{equation}
persisting(c_m, c_n) = True \Rightarrow \frac{|S_i \cap S_j|}{|S_i \cup S_j|} > \omega_{overlap}
\end{equation}

So we consider a community $c_m$ of $C_i$ to be the \textit{same} in as a community $c_n$ of $C_j$ if there is a high enough ratio (above the fixed threshold) 
of common to different elements between the two. 

Using this definition, we can define the phenomenon of \textit{emergence} of a community $c_{new}$ of $C_j$ as a community that is not persisting from the previous snapshot (that is does not have a counterpart in $C_i$), that is 
\begin{equation}
\centering
c_{new} \in C_j \text{ is an emerging community} 
\Leftrightarrow 
\forall c_k \in C_i : persisting(c_{new}, c_k) = False  
\end{equation}

Similarly, we define the \textit{disappearance} of a community as the inverse phenomenon, i.e. by finding communities in the older snapshot that don't exist in the newer one.

To investigate the phenomenon of merging communities, we use the 
definition of persistence, this time considering a union of multiple communities as a potential new merged community. Namely, we detect a \textit{merging} event if we can find a subset of $C_i$ that is the \textit{same} as some community $c_{merged}$ in $C_j$:
\begin{equation}
\begin{split}
    c_{merged} \in C_j \text{ is a merged community} \Leftrightarrow \\
\exists c_l, \ldots, c_n \in C_i : persisting(\cup_{i \in l,\ldots,n} c_i, c_{merged}) 
\end{split}
\end{equation}
The splitting of two communities is defined as the inverse phenomenon of merging, i.e. by inverting the snapshot indices.

\section{Evaluation Protocol}
\label{sec:evaluation-dragon}

\noindent
In this section we describe our protocol to evaluate the proposed solution. 
As a first step, 
a series of snapshots from a KG needs to be considered, e.g. DBLP 
\footnote{https://dblp.org/} over 4 years. 
DBLP collects scholalry data specialized on computer science bibliography. DBLP listed more than 3.66 million journal articles, conference papers, and other publications on computer science in July 2016, up from about 14,000 in 1995. Our choice to use this dataset is motivated by two main reasons:
    1) The regular 
    snapshots (monthly since 2015)
that will allow us to better analyze the evolution of data, and
    2) The nature of the data exposed by DBLP which we are more familiar with.



In order to evaluate our solution, two experimental settings are designed: 
(i) controlled and (ii) random.

\vspace{0.2cm}
\textbf{Controlled Experiment:}
The aim is to create artificial cases of evolution and check if the solution is able to detect such cases. Specifically, given a KG, a subsets of nodes / link within an homogeneous sub-graphs are removed.  
%
%

This experiment may be realised with varying levels of granularity by taking smaller samples of the total set of differences between the graphs. Additionally, it may be repeated by fixing the size of the samples, but computing different samples. Since this will produce different sets of intermediary snapshots, it will allow for a variety of possible evolutionary steps between snapshots $S_1$ and $S_2$.


\vspace{0.2cm}
\textbf{Random Experiment:}
In this section, we describe the settings for the evaluation of our solution using random changes in the knowledge graphs.

Similar to the the case of controlled experiment, 
given a snapshot $S_1$, we compute a series of $n$ snapshots following each of the following actions: 
%
    1) Addition of a fraction $0 < \alpha < 1$ of the values 
    of $m$ entities in the snapshot.
    2) Deletion of a fraction $0 < \alpha < 1$ of the values 
    of $m$ entities in the subset.
    3) Update of a fraction $0 < \alpha < 1$ of the values 
    of $m$ entities in the subset.
    4) All of the previous items combined.

These random changes will differ in the controlled set of experiments in that we have no guarantee that the final result will be meaningful or even consistent. By doing this, we will evaluate the performance of our proposed approach on a more unrestricted environment. Moreover, because parameters $m$ and $\alpha$ can be adjusted without affecting the evaluation process in a higher level, we will increase the scope of the possible evolutionary paths we may evaluate. In particular, some of these evolutionary paths may be impractical and unlikely to occur in real life, but they are important to test the completeness of the designed method. 


\vspace{0.2cm}
\textbf{Atomic Evolution}

In this section, we describe the evaluation of our method for detecting changes at the atomic level on our set of snapshots.

Given that the 
set of snapshots, we 
create pairs of consecutive snapshots. In the case of the controlled snapshots, we know exactly how many changes have been made. For the random snapshots, we may have to compute the number of changes between snapshots. 

Once we know exactly how many changes have been made between snapshots, we may compute all of the noteworthy changes between the snapshots, as defined in Section \ref{atomic}.


\vspace{0.2cm}
\textbf{Local Evolution}

In this section, we describe how to evaluate the performance of our method for detecting changes at a local level.

At this point, we need to find out the communities in each of the snapshots we have generated. The result 
will be a set of sets of communities.

Following this, we may proceed in a similar fashion as in the previous section, only instead of calculating the number of changes and the noteworthy changes between snapshots, we will be computing the same values, but for communities.

\vspace{0.2cm}
\textbf{Global Evolution}

In this section, we describe how to evaluate the performance of our method for detecting changes at a global level, focusing on communities.

By analyzing the communities of each of the intermediary snapshots, we will be able to determine points in time when communities merge, split or disappear entirely. We proceed as follows:
    a) If the number of communities is greater than in the previous snapshot, we may conclude that more communities have either split or found than merged or deleted.
    b) Conversely, if the number of communities is less than in the previous snapshot, we may conclude that more communities have either merged into other communities, or a have been deleted than split or created.
    c) If the number of communities has not changed, it is possible that no changes have occurred.


In addition, if we have identified communities that persist among the intermediary snapshots, we may study their local evolution as specified in Section \ref{sec:localEvolution}. This will allow us to examine the evolution in individual communities.

\section{Related Work}
\label{sec:related-dragon}

\todo[disable]{I would suggest to put this as the last section of the paper. Also I do assume that you are going to create a narrative on the reason why this related works are relevant for your purpose. Additionally, it is important to highlight what are the limitations of these works with respect to your research goal(s) and how are you going to address and/or solve these limitations. Also I did not check the references, I do assume they are fine. I don't know if you plan to extend them. For sure extending them is the last action to be considered for the current purpose, provided that you set up a reasonable collections of relevant and related works.}

Evolution within KGs has not been studied as thoroughly as many other topics regarding KGs and structured data in general.
Bonatti et al.~\cite{bonatti} present in their report an overview of the major current topics in research related to KGs, with a strong focus on capturing various aspects of evolution and understanding it.
One recent effort to study this phenomenon includes Esteban et al.~\cite{esteban}, who also explored the evolution of clusters within the DBLP dataset.
They focus specifically on predicting the evolution of KGs based on latent representations of the model, and the event model as inputs.
However, their use cases cover specific domains unrelated to DBLP we focused on. 

Also of interest is the work by Chakrabarti et al.~\cite{chakrabarti} on evolutionary clustering within homogeneous networks.
They define a method for clustering that is smoothed across timesteps in order to preserve clusters where possible.
They discuss the drawbacks to traditional, semi-randomized clustering approaches, which constitutes an important foundation. 
Similarly, in order to capture the fundamental changes that induce evolution in a graph, we need to capture both explicit and latent changes within the graph.
Singh et al.~\cite{singh2018delta} present Delta-LD, a technique for detecting changes between different versions of a linked dataset.
Their work acts as a foundation for generating change sets that are necessary in order to explore evolution in KGs.
However, once we have those changes, we need to know what the clusters or communities are within our data.
For that, we leverage the techniques discussed by Cuvelier et al.~\cite{cuvelier2011graph}, who present an overview of graph mining and graph-based clustering techniques.


Finally, we wish to also look ahead and see for what purposes this sort of analysis can be used.
In order to motivate the solution, we examined both predictive tasks and summarization tasks as future work.
For change prediction, there are several such sources available that focus on predicting different aspects within KGs.
One such example is by Dasgupta et al.~\cite{dasgupta}, who focus on temporal features within a KG and how they may change over successive versions of the KG.
Chiang et al.~\cite{chiang} present a particularly important study since they also use the DBLP dataset.
However, their system predicts the probability of a value re-appearing over time.
These both point toward ways in which we might use our system of identifying evolution to begin predicting evolution instead.


Finally, as future work we explored the possibility to give \textit{intentions} or \textit{explanations} to sets of changes which bring about detected evolution.
One such approach is to summarize the changes that are seen such that a human might be able to deduce the intentions behind those changes.
For example, the work by Tasnim et al.~\cite{tasnim} adds descriptions to entities which have changed over time.
However, in our case we would hope to provide a description for changes that occur within a KG, rather than the entities themselves.

\section{Conclusions and Future works}
\label{sec:conclusions-dragon}

\noindent

In this work we proposed a methodology to capture significant evolution in Knowledge Graphs taken into consideration three types of evolution: (i) atomic level, (ii) local level and (ii) global level. We focused on 
the scholarly data domain even if 
the actual approach can be generalized to other domains. 

Evolution in Knowledge Graphs has yet to be widely explored by the academic community: our work contributes to the already existing studies, 
while further deepening the discussion about methodologies that can be used, expected results and ways to verify them. 

Many trajectories for future work can be 
identified, as well as variables in the data processing that could improve our study: (i) the first prosecution would be to provide an understandable summarizing for the changes that have been captured. This part would probably benefit by human intervention. A classification of the highlighted variations could also need a human contribution, to perform annotation.
To identify patterns of changes (change of affiliation, evolution of h-index,etc.) in a proficient way it is better to have many snapshots, taken with short time range between one and the following other. This may depend on the pre-conditions of the data-set. In the case where the KG is a proprietary one, experiments can be conducted managing the snapshots more freely. Then, it would be also possible to establish characteristics for relations by frequency of change: static, periodic, or frequent ones. (ii) The second prospective that we can take into consideration is the distinction between the valid evolution from the noisy one. We can resolve this by developing a classifiers. Furthermore, the elimination of noise may allow us to better understand the causes of evolution. (iii) An other prosecution of this work is to provide an explanation for a capture evolution phenomenon. It would be interesting to know why this evolution and what are the main causes of it, so that we can perhaps predict these changes for the future. (iv)Finally, we also consider important to address a possible meta evolution: it will mean comparing the changes across already defined inter-snapshot evolution. In addition, a different approach would need to be outlined in the case where two graph are to be linked.


\chapter{The Evolution of Knowledge Graphs: An Analysis using DBpedia}\label{sec:mordor}
\chapterauthor{Ling Cai, Stefano De Giorgis, Sebasti\'an Ferrada, Genet Asefa Gesese, Frances Gillis-Webber, Neha Keshan, Heiko Paulheim}

\section{Knowledge Graphs Evolution and Preservation}
\label{sec:def-mordor}
\noindent Knowledge graph (KG) evolution can be defined as the periodic or continuous updates to a KG, which involves deleting, adding, and updating data and/or the schema. 
Every change to the KG is typically in response to the state of the world changing for the domain which the knowledge graph represents. 
However, a KG may not necessarily be an accurate representation of said domain. 
This can be due in part to: 
\begin{enumerate}
    \item the oftentimes crowdsourcing nature of knowledge graphs, where subjects can be the object of vandalism,
    \item the inadvertent misrepresentation of information, and
    \item a lack of human resources (particularly for knowledge graphs maintained by small communities, such as some localised versions of DBpedia), as a result, the knowledge graph can be slow to react to changes in the domain for which it represents.
\end{enumerate}

The preservation of a knowledge graph pertains to its provenance -- that is, providing traceability for each change, be it an addition, update or deletion.

\section{Introduction}
\label{sec:intro-mordor}
\noindent 
Knowledge Graph (KGs) consist of a set of triples where a triple contains a subject $s$, a property $p$, and an object $o$ in the form of $<s, p, o>$. KGs have been used for the purpose of sharing linked data. For example, the typical KGs are DBpedia \cite{lehmann2015dbpedia}, Freebase \cite{bollackerfreebase}, the Wikidata \cite{vrandevcic2014wikidata}, etcetera. 
As the state of the world is constantly undergoing change, KGs can evolve in response to new information that is generated. 
KGs are thus intended to be dynamic structures: there are periodic or continuous updates that include new, refined, and redefined data and/or schema. 
We refer to these changes over time as the evolution of KGs. Studying the evolution of KGs is of significance for data-driven applications, such as event prediction~\cite{esteban2016predicting}, change verification~\cite{nishioka2018analysing}, entity summarisation~\cite{tasnim2019summarizing} and so on.

The following research questions were thus identified:

\begin{description}
	\item[RQ1] What are the characteristics of an evolving KG, using DBpedia as the use case?
	\item[RQ2] How can the evolution of KGs be exploited as a training signal?
\end{description}

The DBpedia community project is an effort to provide a structured version of the information contained in Wikipedia  \cite{lehmann2015dbpedia}. 
DBpedia periodically extracts the data within infoboxes and other relevant data from Wikipedia, with the infoboxes serving as the main source of information for DBpedia \cite{lehmann2015dbpedia}. 
DBpedia then provides both an RDF representation and a SPARQL endpoint for further querying \cite{lehmann2015dbpedia}. 
Using two datasets from DBpedia, we propose an approach to measure the \emph{volatility} of relations in an evolving KG, which can then be used to support further analysis. As a volatile relation, we understand a relation whose objects are likely to change over time, such as the population and the president of a country. In contrast, non-volatile relation is a relation with a low likelihood of such changes, such as the area and the capital of a country.

Potential use cases for evolution awareness have been identified, namely:

\begin{itemize}
	\item Aprosio et al.: their work tries to automatically compile Wikipedia infoboxes, using Relation Extraction trained on DBpedia properties, with the Distant Reading method applied to triples of subject-attribute-value \cite{aprosio2013extending}. 
	\cite{aprosio2013extending} could benefit from our work by having more awareness about volatility of properties.
	\item Bryl and Bizer: their work develops fusion policies for data from different KGs \cite{bryl2014learning}. 
	As the fusion policies are likely to be different depending on the volatility of a property (i.e. volatile and non-volatile), such approaches could  also benefit from our work.
\end{itemize}

\section{Related Work}
\label{sec:related-mordor}
\noindent Schema.org provides a vocabulary for marking up structured data in HTML pages \cite{meusel2015web}.
Meusel et al. conducted an analysis of the usage of Schema.org over time using a series of large-scale web crawls \cite{meusel2015web}. 
Although as a KG, the purpose of schema.org is very different to DBpedia, there were interesting phenomena found, such as schema abuse, fast and slow adoption of new schema elements in the data, semantic drift, that could be applicable to DBpedia.

With Wikidata as their focus, Tanon and Suchanek propose a system to make the edit history of Wikidata accessible via a SPARQL endpoint \cite{tanon2019querying}. 
Following each revision, both the diffs and the global state of the graph are indexed after said revision. 
The result is a system which allows more complex SPARQL queries, thus enabling, among other features, the contributions of users (automated or human) to be traced over time.

Nishioka and Scherp analyse the evolution of a KG, using Wikidata, focusing on topographical features \cite{nishioka2018analysing}. 
As a result, their analysis does not rely on editors' history, thus web allowing for changes by both new editors and bots to be included in the analysis. 
A change is deemed to be correct if it remains unreverted for ~4 weeks. 
This information is then used to predict the accuracy of a change on a KG. 
Their findings reveal that a KG’s evolutionary patterns follow that of social networks.

Gonzalez and Hogan propose the computation of lattices containing information about the characteristic sets of the entities of Wikidata \cite{gonzalez2018modelling}. 
They define an algebra over such lattices in order to characterise the evolution of a knowledge graph.
This difference is used to train a model capable to predict which properties will be added or removed in future versions of the KG.
Esteban et al. present a link prediction model, modelling changes in the knowledge base as events. 
They thus use the historic events in order to predict future ones\cite{esteban2016predicting}.

Noy et al. present an ontology-versioning environment which allows for the structural changes (instead of text presentation changes) of ontologies to be compared in a versioning system \cite{noy2004tracking}. 
Using an efficient version-comparison algorithm, a structural diff is then produced between ontologies. 
The result allows for users to analyse the changes within a user interface.

Taking a cue from \cite{tanon2019querying} and \cite{noy2004tracking}, we focus on the diffs, but our approach diverges in that the focus is primarily on analysis to derive insights, unlike that of \cite{tanon2019querying} and \cite{noy2004tracking}, who both provide a system in which the user is expected to derive their own insights.

\section{Resources}
\label{sec:resources-mordor}
\noindent We make use of sub-datasets from two datasets available for download from DBpedia for the periods 2014 and 2015-04, referred hereon as \textit{DS1} and \textit{DS2} respectively \cite{ref:db:2014}; \cite{ref:db:2015}. 
Although neither are recently published datasets, they were considered suitable for analysis of the research questions due to the fact that from version 2015-10 onward there have been many changes in the way the data was provided. 
This makes it difficult to compare and track changes across versions because it is no longer possible to distinguish organic updates from updates forced by the changes of the schema.

\textit{DS1} is comprised of three files from the 2014 dataset:

\begin{itemize}
	\item \textit{DS1-MP:} mapping-based properties from the English version \cite{ref:db:2014:mapping},
	\item \textit{DS1-IT:} instance types from the English version \cite{ref:db:2014:instances},
	\item \textit{DS1-OWL:} the OWL ontology \cite{ref:db:2014:owl}.
\end{itemize}

Similarly, \textit{DS2} is comprised of three files from the 2015-04 dataset:

\begin{itemize}
	\item \textit{DS2-MP:} mapping-based properties from the English version \cite{ref:db:2015:mapping},
	\item \textit{DS2-IT:} instance types from the English version \cite{ref:db:2015:instances},
	\item \textit{DS2-OWL:} the OWL ontology \cite{ref:db:2015:owl}.
\end{itemize}

For \textit{DS1}, the English version of the dataset contains 583 million triples which describe `facts' for 4.58 million `things' \cite{ref:db:wiki:2014}. 
For \textit{DS2}, the English version has increased to 737 million triples, with the `things' described now 5.9 million \cite{ref:db:wiki:2015}.

DBpedia also provides live updates, available for download on an hour-by-hour basis, for example \cite{ref:db:live}. 
Although it would be preferable to consider more recent data in our analysis, due to external constraints, this was not possible. 
As it was concluded that \textit{DS1} and \textit{DS2} served sufficiently for analysis, DBpedia's live updates could be considered for future work.

\section{Proposed approach}
\label{sec:approach-mordor}

\noindent
In this part, the methods proposed to solve the research questions defined in Section \ref{sec:intro-mordor} are presented. 
In subsection \ref{subsec:PropertyExp}, the patterns of changes in properties are explored by using basic statistical analysis as a solution for RQ1. 
Then, Section \ref{subsec:BasicAna} discusses the attempt that has been made to use the identified patterns to help predict the growth of properties in order to address RQ2.

\subsection{Basic Analysis}\label{subsec:BasicAna}
In order to capture the changes of properties over time in the studied datasets, the modified triples are classified into two categories, namely, `Added' triples and `Removed' triples, in terms of addition and removal of triples with these properties respectively. 
The combination of the triples from these two categories is referred to as `Edited' triples. 
These three derived data sets are used to explore the changes from different perspectives. 
In general, we look into the number of changes of properties, w.r.t to the absolute and the relative number of changes. 
Specifically, basic descriptive statistics is used to discern the most frequently changed properties and the least frequently changed properties. 

\subsection{Property Expansion Prediction} \label{subsec:PropertyExp}
Once the characteristics of the evolution of a KG has been analyzed, it is possible to anticipate future changes that can happen to the KG. 
Thus, in this section, an approach which can be used to predict the growth of properties is proposed. 
The growth of properties can be seen, for example, in terms of the degree of properties or in terms of the graph structure of properties. 
In this research, the focus lies on utilizing the graph structure of properties. 

As a first step, a series of snapshots of DBpedia over a period of time are generated for each property. 
Instead of treating entities as nodes as most KGs do, in our case, dual graphs for each property are constructed, in each of which properties/relations are represented as nodes and entities are treated as edges. 
In this way, the task is designed as a time-series prediction task on a graph data structure.
Formally, the problem can be defined as follows:
\\

\textit{Given a time-series graphs of properties, with a specified time scale, predict the graph structure of properties at the next time step.}
\\

\textbf{Proposed Solution:} Enlightened by the success of graph representation in the past few years, in this research, a Temporal Capsule Neural Network-based model, which extends the Capsule Neural Network \cite{sabour2017dynamic} to the temporal domain is proposed to solve the problem defined above.

\section{Evaluation and Results: Proof of concept - Experiments}
\label{sec:evaluation-mordor}



\subsection{Ranking Properties Based on Change Frequencies}

\begin{figure}[t]
    \centering
    \begin{minipage}[b]{0.4\textwidth}
    \includegraphics[width=\textwidth]{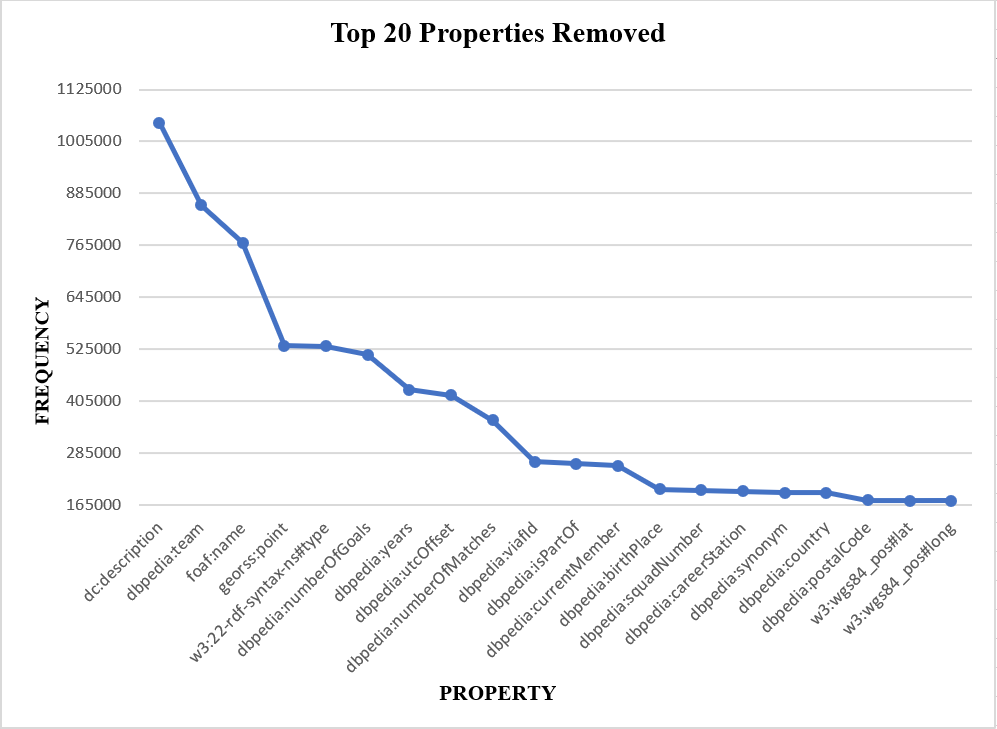}
    \caption{The twenty most frequently changed properties which were removed from the 2014 release}
    \label{fig:top-20-prop-removed}
    \end{minipage}
    ~
    \begin{minipage}[b]{0.4\textwidth}
    \includegraphics[width=\textwidth]{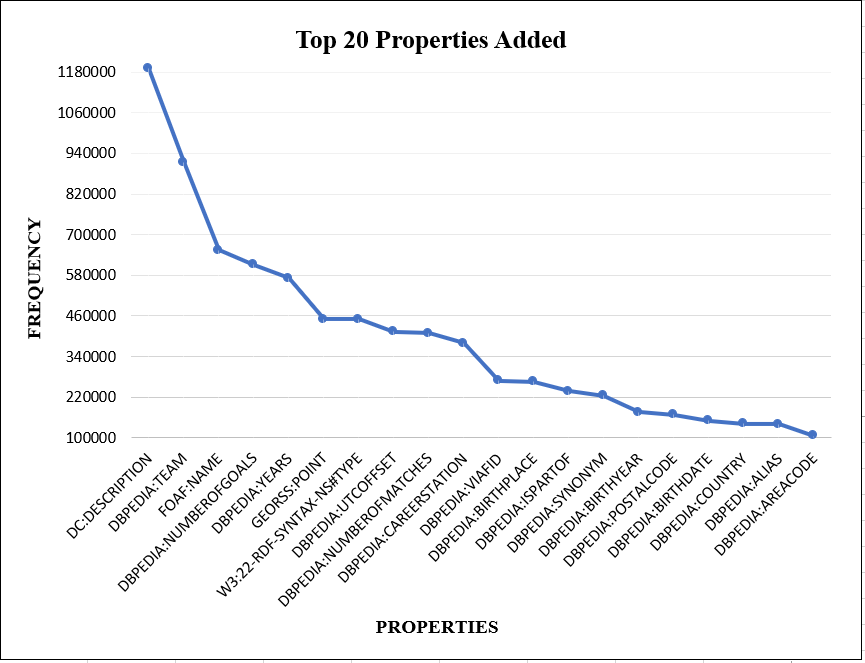}
    \caption{The twenty most frequently changed properties which were added in the 2015-04 release}
    \label{fig:top-20-prop-added}
    \end{minipage}
\end{figure}

\begin{figure}[t]
    \centering
    \includegraphics[width=.8\textwidth]{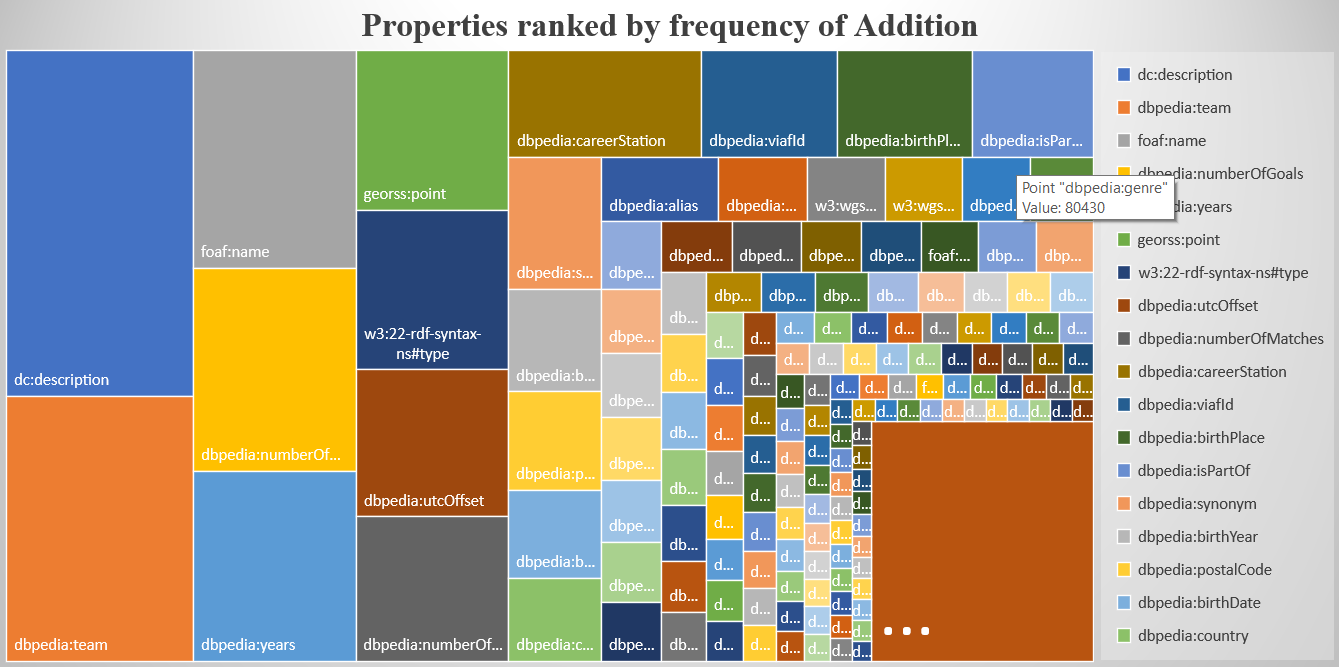}
        \caption{The properties which were added in the 2015-04 release based on their frequencies}
        \label{fig:prop-added}
\end{figure}

\begin{figure}
    \centering
    \includegraphics[width=.8\textwidth]{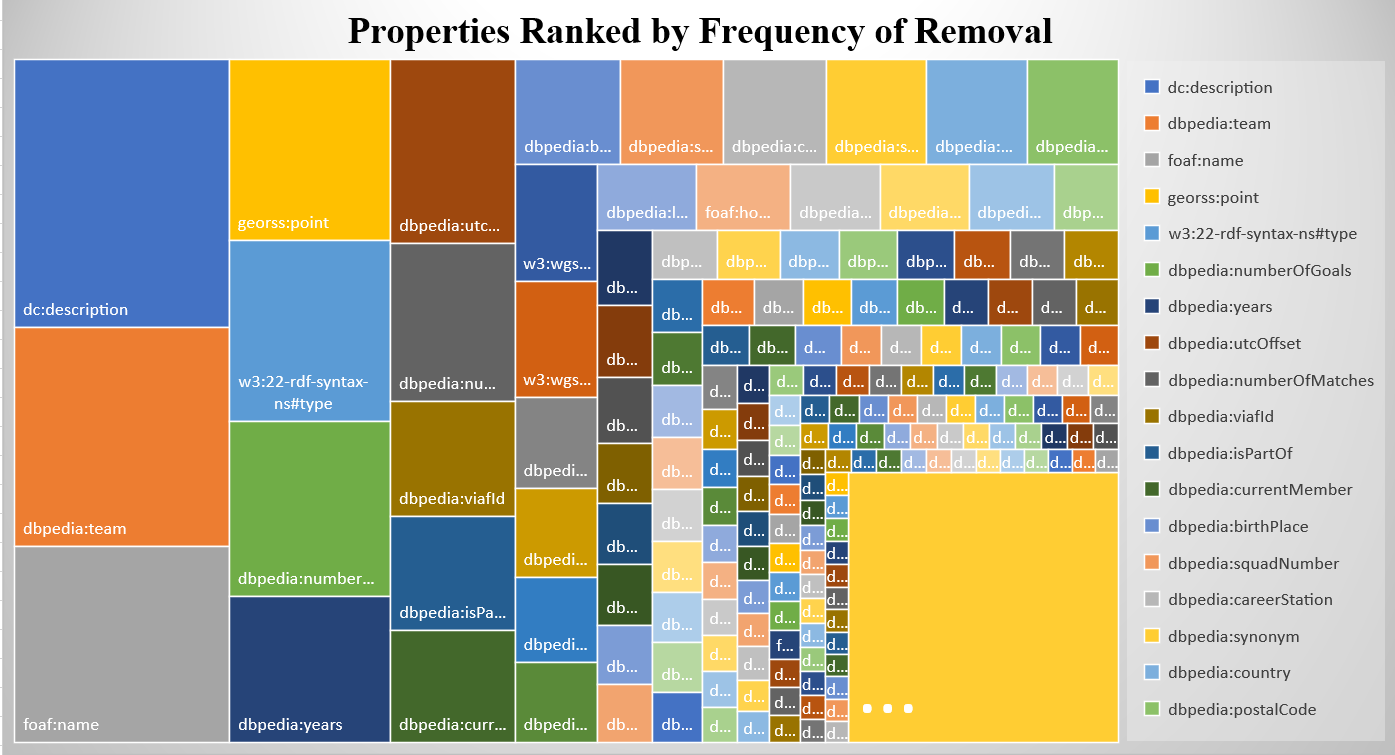}
    \caption{The properties which were removed from the 2014 release based on their frequencies}
    \label{fig:prop-removed}
\end{figure}

\begin{figure}[t]
    \centering
    \includegraphics[width=.6\textwidth]{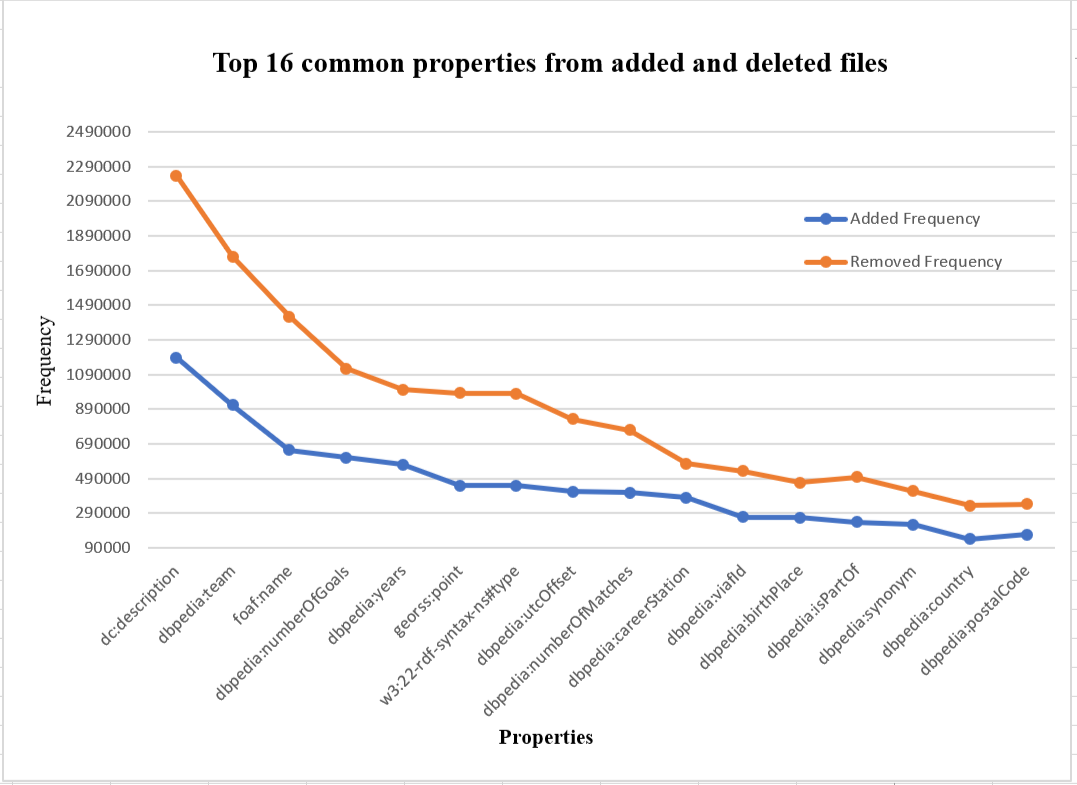}
    \caption{Frequently changed properties common to both the Added and the removed files}
    \label{fig:top-20-prop-common}
\end{figure}

The frequency of each property was calculated, irrespective of the subjects and objects connected through them for both the `Added' and `Removed' files. Each property was ranked based on its frequency in the `Added' Fig.~\ref{fig:prop-added}and the `Removed' Fig.~\ref{fig:prop-removed} files. This provided us with the top twenty most frequently added properties ( Fig.~\ref{fig:top-20-prop-added}) and removed properties (Fig.~\ref{fig:top-20-prop-removed}). This shows that at least fifteen properties are common between the frequently added and deleted properties Fig.~\ref{fig:top-20-prop-common}.

Fig~\ref{fig:prop-added} represents all the properties that were added to the \textit{DS2} version of DBpedia. The area of each property is based on its frequency in the `Added' file. The more frequently a property was added, the larger the area. At the bottom right corner we see a large area with three dots in it, representing the expansion of the graph to see less frequently added properties. Fig~\ref{fig:prop-removed} represents all the properties that were removed from \textit{DS1}. It can be interpreted similarly to the previous figure representing the added properties. 

From Fig~\ref{fig:top-20-prop-added} and Fig~\ref{fig:top-20-prop-removed}, we see that birth date, alias, area code and birth year were frequently added in \textit{DS1}, whereas these properties are not in the top twenty frequently removed properties. This is reasonable, as properties such as birth date and birth year do not change frequently, unless it requires a correction. On the other hand, properties such as the current member and squad number were among the top twenty removed properties. These represent the volatile properties as members join and leave and the system needs to be updated with the latest information. Moreover, not all frequently changed properties are volatile in nature, for example, the birth date. From the data, it can be seen that many names were added, suggesting the addition of their birth details, making birth date one of the most frequently changed properties.

\begin{figure}[t]
    \centering
    \begin{minipage}[b]{0.4\textwidth}
    \includegraphics[width=\textwidth]{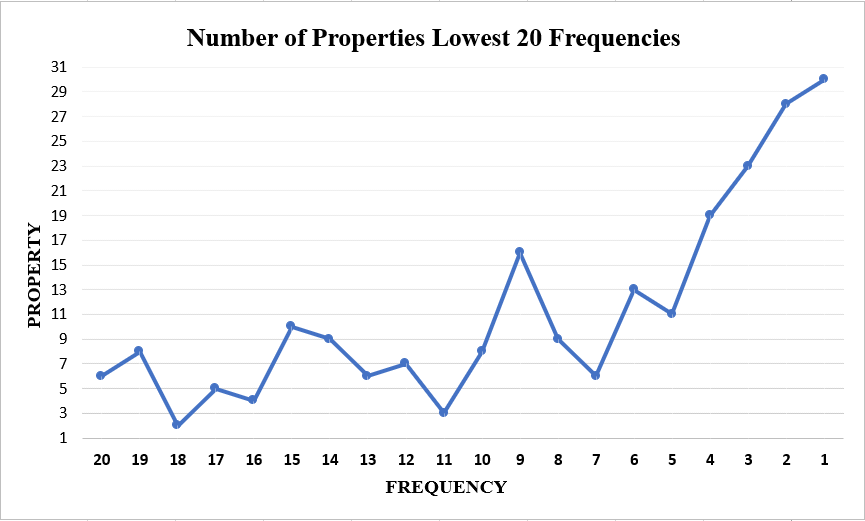}
    \caption{Infrequently changed properties from the added file}
    \label{fig:low-fre-add}
  \end{minipage}
 ~
  \begin{minipage}[b]{0.4\textwidth}
    \includegraphics[width=\textwidth]{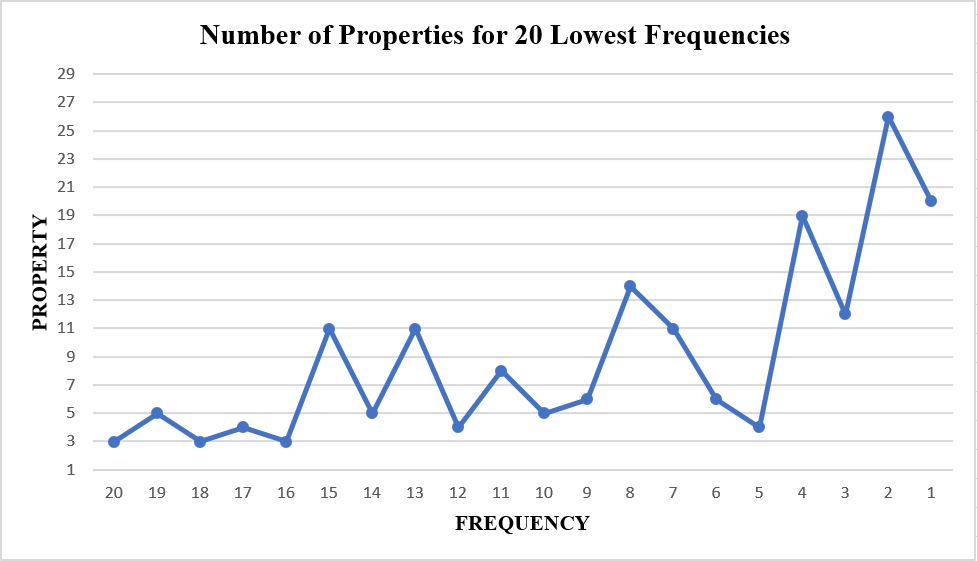}
    \caption{Infrequently changed properties from the removed file}
    \label{fig:low-fre-rem}
    \end{minipage}
\end{figure}

For the least updated properties, we decided to group the properties based on their frequencies and see the correlation between the number of properties and the frequencies for the properties changed once to twenty times. Fig~\ref{fig:low-fre-add} shows for the added properties while Fig~\ref{fig:low-fre-rem} portrays the least changed removed properties.

Now that the properties from both the "Added" and the "Removed" files were evaluated separately, the data from both the files were compared to provide some useful insights. The 

The above figure provides us a sense of frequently updated properties, serving as a starting point to find the most volatile and non-volatile properties. These inferences can help future researchers to design the schema of knowledge graph in a much concrete manner.


\subsection{Relative Frequency Analysis}
\begin{figure}[ht]
    \centering
    \begin{minipage}[b]{0.4\textwidth}
    \includegraphics[width=\textwidth]{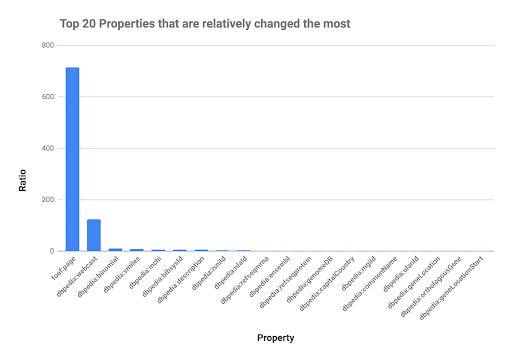}
    \caption{Properties that are relatively changed the most frequently.}
    \label{fig:top20_relative_freq}
  \end{minipage}
 ~
  \begin{minipage}[b]{0.4\textwidth}
    \includegraphics[width=\textwidth]{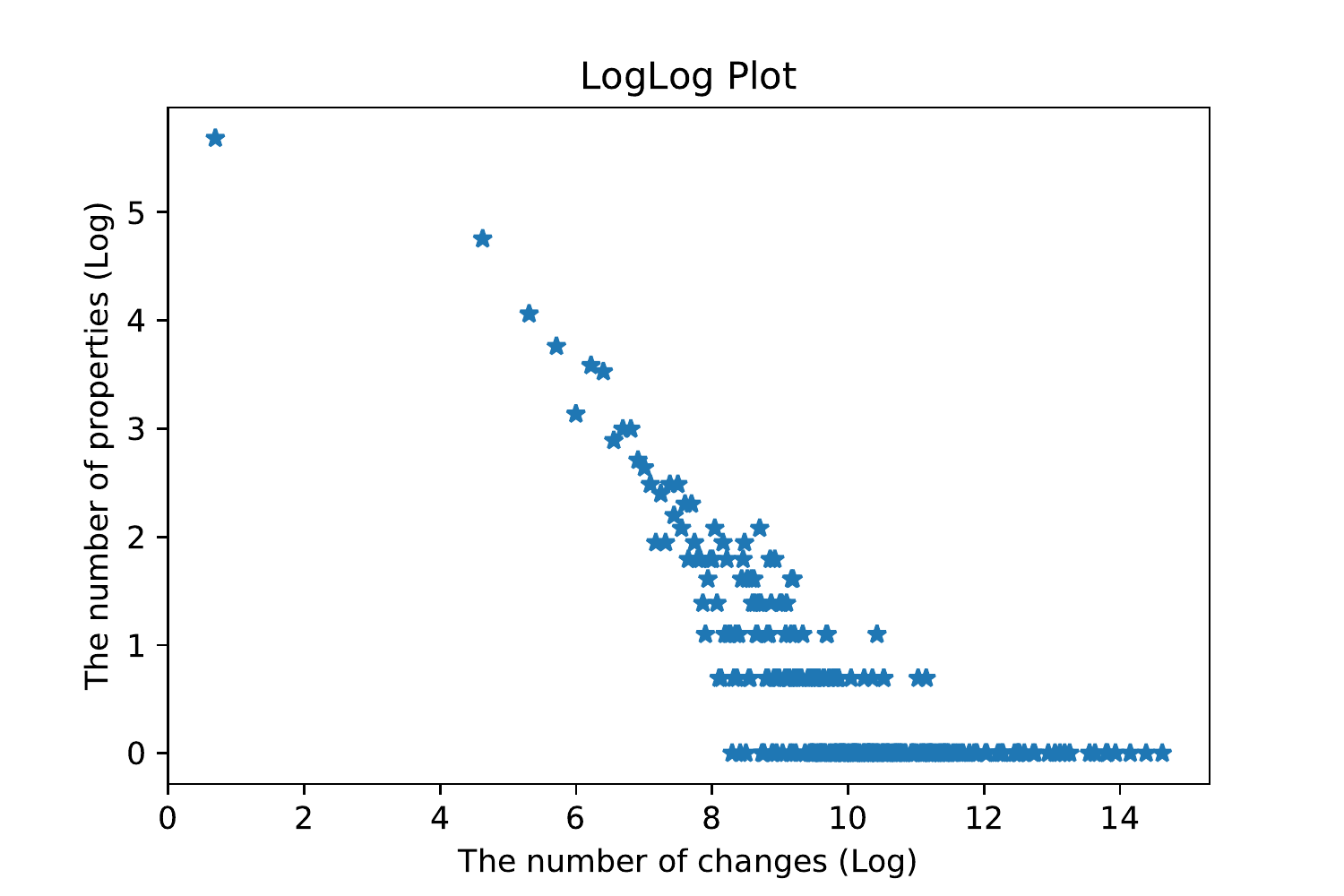}
    \caption{Relation between the number of properties (Log) and the number of changes (Log) per property.}
    \label{fig:loglog}
    \end{minipage}
\end{figure}

In this subsection, first the number of changes per property was considered and loglog plot was used to approach the basic network analysis. As shown in Figure.~\ref{fig:loglog}, it can be found that most of properties are changed a few times but a few properties are frequently changed. This result shows that the evolution of knowledge graph presents the same characteristics of common networks, which follow the power law distribution. Furthermore, the relative change rate for each of the property was calculated by using Eq.\ref{eq:change_rate}. From Figure~\ref{fig:top20_relative_freq}, it is observed that the property - "foaf:page" has the highest ratio, which means in 2015 pages have been edited way more than those in 2014. "DBpedia:webcast" was ranked the second. 

\begin{equation}
    ratio = \frac{\text{number of changes per property}}{\text{occurrence of the property in 2014}}
    \label{eq:change_rate}
\end{equation}

\subsection{Instance Type Dynamics}

When new classes appear in the schema, properties can become specialized in terms that they can have a more specific range or represent more accurately the reality by lifting schema imposed restrictions. As an example, we found that since the introduction of the class dbo:EthnicGroup, hundreds of entities had their dbo:nationality property updated from a dbo:Country to the new class thus being more accurate.

We analyze the way types of instances change throughout time and find if they become more specialized or more general. To achieve that, we queried the data from DS1-MP, DS1-IT, DS2-MP and DS2-IT to obtain the classes of the objects in the older and newer versions, for a fixed properties. Later, we counted the number of objects that register a type change on a given property in order to explore the dynamics present in the data. 

\begin{figure}[ht!]
    \centering
    \includegraphics[width=\textwidth]{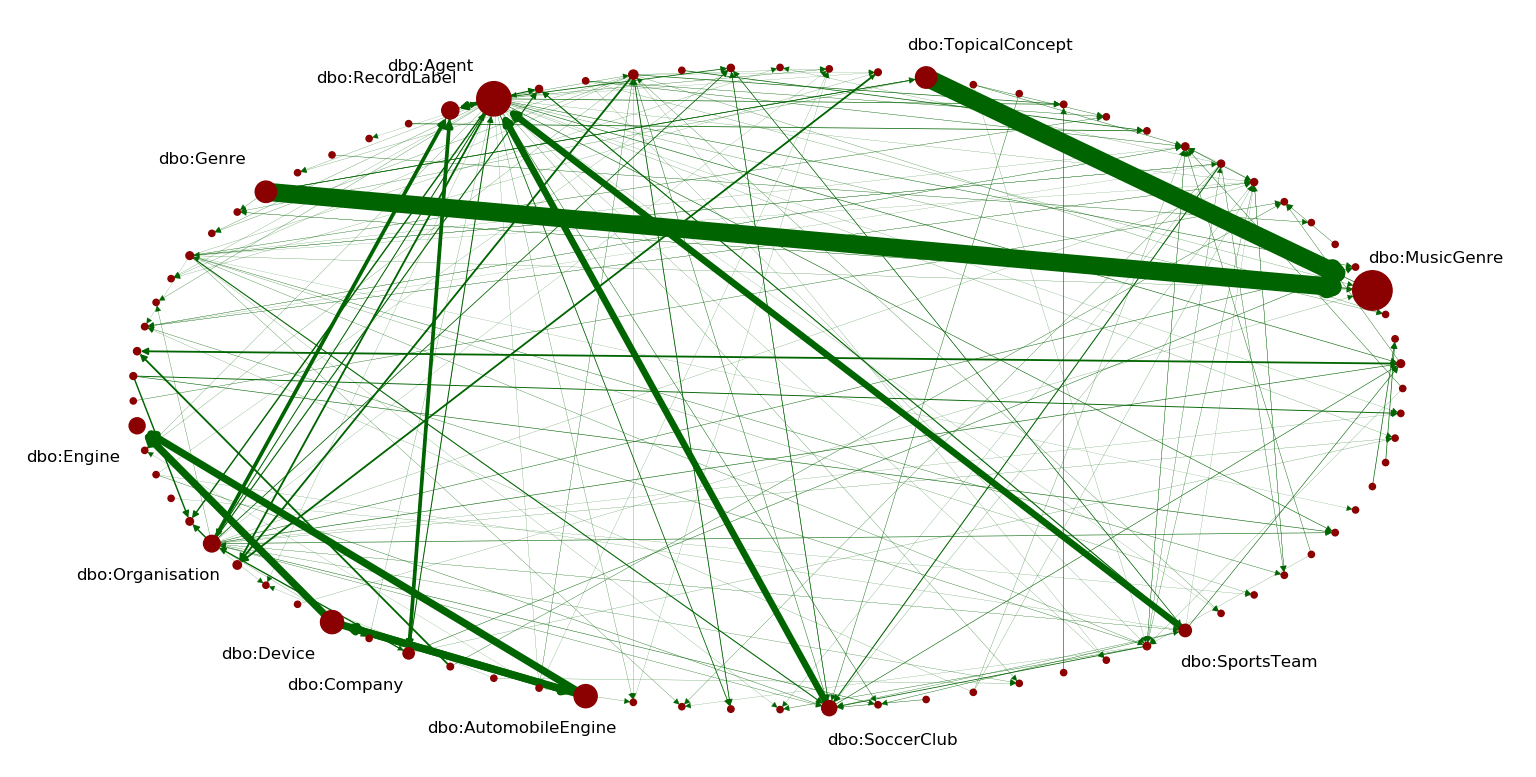}
    \caption{Type Evolution Graph. Nodes represent the classes involved and edges the properties whose objects changed from one class to another. Only the properties with a high number of changes are depicted.}
    \label{fig:vol-graph}
\end{figure}

Fig.~\ref{fig:vol-graph} shows a network representing the dynamics of the type of the entities in DBpedia. It contains the relationships of the classes that presented more than 100 changes from or to them. The size of the nodes is proportional to the total number of changes. The edges represent the properties that are affected by these type changes, their width represents the number of objects that changed their type. It can be appreciated that the class dbo:MusicGenre is an specialization of the classes dbo:Genre and dbo:TopicalConcept when using the property dbo:genre. Another specialization of the old classes dbo:Organization and dbo:Company can also be seen as dbo:RecordLabel is introduced as a type for the property dbo:recordLabel. Some other interesting changes happen with the properties dbo:birthplace and dbo:nationality, which use to have range dbo:PopulatedPlace and dbo:Country respectively, now they present a type migration that includes dbo:Settlement and dbo:EthnicGroup resp. thus representing reality in a more precise way.

\begin{figure}[ht!]
    \centering
    \includegraphics[width=\textwidth]{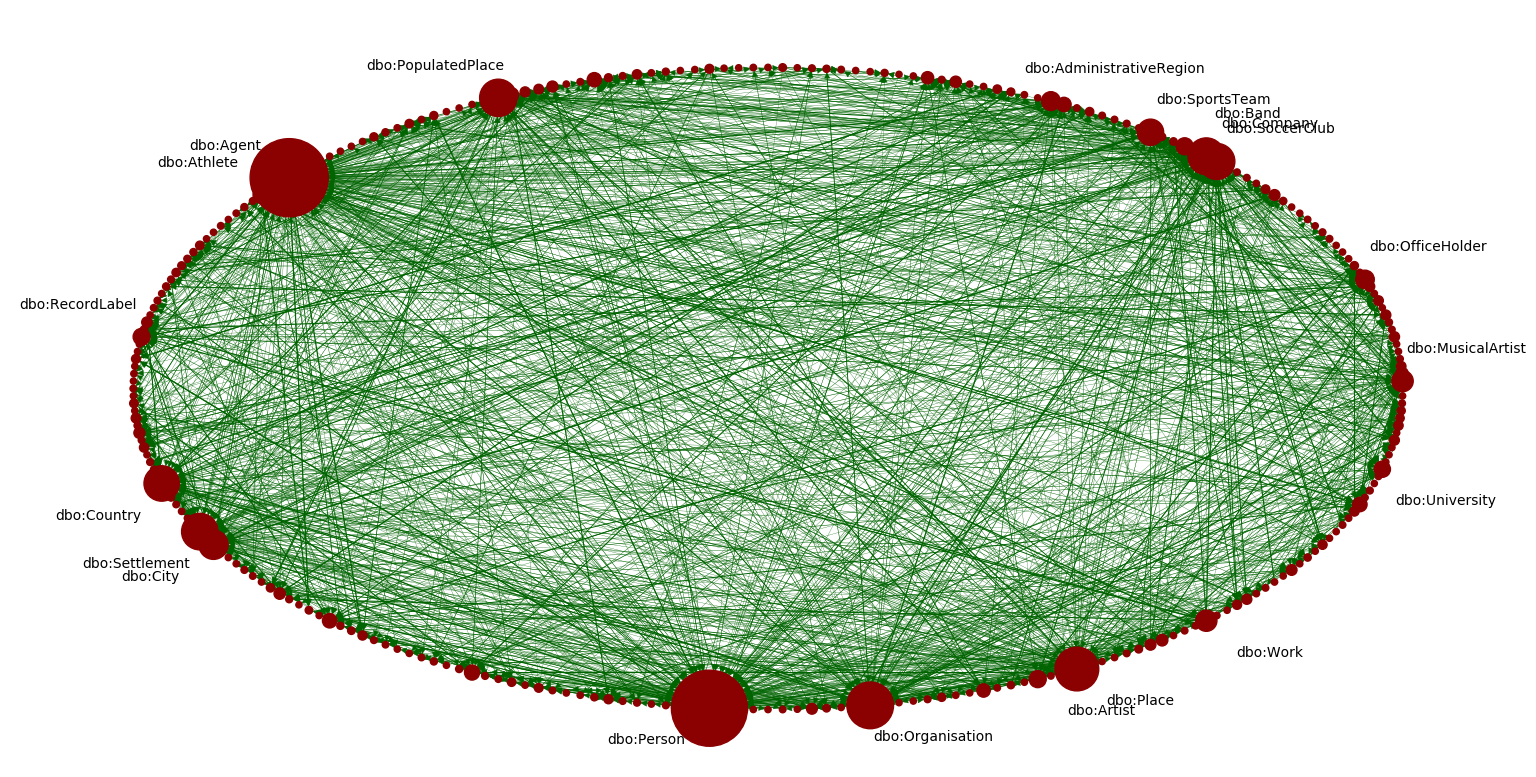}
    \caption{Type Evolution Graph. Nodes represent the classes involved and edges the properties whose objects changed from one class to another. Only the properties with lower number of changes are depicted.}
    \label{fig:static}
\end{figure}

Fig.~\ref{fig:static} shows the properties that produced the lesser amount of type changes - less than 10. Visual marks refer to the same variables as Fig.~\ref{fig:vol-graph}. We can see an increase in the number of properties, meaning that most of them are static in terms of type evolution. Some of the properties in this group are dbo:musicalArtist, dbo:leader, dbo:location, dbo:occupation, dbo:owner, and dbo:succesor. We see that it is unlikely for these properties to present a radical change of their range. In Fig.~\ref{fig:static} it can be appreciated that the classes that receive new entities are more general, as the likes of dbo:Person, dbo:Agent, dbo:Place, and dbo:Organisation.

From both of these analysis we might infer subclass relationships among types when they are not present in the schema. However, it is necessary to thread carefully: the relationship goes from superclass to subclass when the change is greater, but goes from subclass to superclass when the change is lower. We propose this analysis as an initial empirical evidence that we expect to generalize as future work. These analysis can also be generalized to define a metric of the \textit{volatility} of the involved properties.

\section{Discussion and Conclusions}
\label{sec:conclusions-mordor}
\noindent

Although the focus of this paper has been on DBpedia, it would be of interest to determine if the identified characteristics could be abstracted to other KGs. An item for future work could thus be to evaluate the characteristics through the lens of the phenomena identified by \cite{meusel2015web}, to determine if said phenomena extend to DBpedia.

Metrics to understand the volatility of properties and for the prediction of subproperties and subclasses are yet to be formalized and generalized using the initial measurements presented in this paper.

\chapter{Knowledge Graph vs Reality - How to Evaluate Knowledge Graph Evolution}
\label{sec:hufflepuff}
\chapterauthor{Ahmad Sakor, Amine Dadoun, Kholoud AlGhamdi, Laurine Huber, Sepideh Mesbah, Thomas Schleider, Vitor A. C. Horta, Harald Sack}

\section{Research Questions}
\label{sec:rq-hufflepuff}
\begin{enumerate}
    \item Given two versions of a knowledge graph, how to measure the syntactic, structural, and semantic differences between them?
    \item How to automatically detect and adapt KGs to changes in the ``real world''?
\end{enumerate}

\section{Knowledge Graphs Evolution and Preservation}
\label{sec:def-hufflepuff}
According to \citeauthor{kgdef} a knowledge graph acquires and integrates information into an ontology and applies a reasoner to derive new knowledge. Its main goal is to describe real world entities and their interrelations \cite{paulheim2017knowledge}. Many companies such as Google, Facebook, and Siemens are using knowledge graphs for different applications like risk management and process monitoring.

Since the relation between real world entities are constantly changing, a known challenge is to keep the knowledge graph up-to-date with its respective domain \cite{8334436}. In this sense, the \textbf{evolution of a knowledge graph} is represented by the multiple versions it acquires during this constant process of update. While this process occurs it is important to guarantee the preservation of the knowledge graph, which means that incoming changes should not harm its accessibility, consistency or real world representation.  


\section{Introduction}
\label{sec:intro-hufflepuff}

\noindent
Knowledge Graphs (KGs) offer us a way to model reality. However keeping KGs up to date while the entities in the real world continuously change is a challenging task. As the validity of most facts is time-constrained the implicit and explicit knowledge in KGs relies on them being up to date \cite{8334436}. When changes are proposed to a KG and a new version is available, the applications that consumes the KG must decide whether or not to use the most recent version. 

The problem is that predicting the impact of adopting a new KG version and measuring how different it is from the current version it not a trivial task. In this sense, some challenges are: (i) changes can affect large amounts of data which makes analysing them computationally expensive; (ii) small changes might have a high structural impact over the graph; (iii) the semantic difference between two graphs might not be proportional to the structural one. Besides this, in case there are significant differences between them, how can we make sure that changes and revisions on KGs are aligned with their respective real world entities, i.e. they are not false information? And how do we ensure that they stay consistent and hold up against reality at the same time?


In this paper we propose a new ensemble method to help with evaluating changes in different versions of KGs in two ways. First we define how to measure and compare the difference between two versions of a KG. Second we propose a way to check whether the changes in the new version are consistent with the reality. 
For this we will approach this task from different perspectives: syntax changes, graph structure, and semantics.  

For syntactic changes we analyse the RDF serialization in order to quantify changes in classes, properties, labels and relations in the KG. The outcome is to give a first perspective on how many changes were made to detect whether there is a need for deeper analyses (structural and semantics).


In case the number of changes is significant a structural analysis should be performed to understand the possible impact of adopting the new version. The main goal is to detect if the changes occurred in crucial parts of the KG and if they affected central nodes. For this step we consider only the graph structure of the KG and we focus on analysing whether two versions are topologically similar considering specially global measures such as graph diameter and centrality measures such as PageRank distribution.


Recent work on Knowledge Graph embeddings has been promising and the possibility to embed entities and relations into continuous vector spaces has many applications\cite{8047276}. Although some of the KG semantics becomes implicit when the graph is represented in a continuous vector space, these embeddings have shown to to have a consistent semantic representation \cite{rdf2vec}. We explore how to use graph embeddings to quantify and compare semantic changes or differences between KG versions. 


Although comparing graph embeddings is one way to track semantic changes, it might be considered that there are still competitive ways to directly evaluate ontological changes between two knowledge graphs without statistical modelings. We will  look at one state-of-the-art algorithm for ontology matching that can be used for KGs: LogMap. By doing this we aim to capture explicit changes in the ontology such as, e.g. functional property changes, which might not be captured by considering graph embeddings only.


Finally, we want to find out how well a KG represents reality. We will go through a survey of several techniques and their limitations and discuss how we can include this into our final measurement. We will further give some insights on how to tackle the limitations of the existing KG reality-checking technique by incorporating Unsupervised Open Relation techniques.

Therefore the main contributions of this work are: (i) an ensemble method to measure differences between knowledge graphs; (ii) a measurement to indicate which knowledge graph is more compatible with the reality.


The rest of the paper is structured as follows: Section~4 summarizes related work, while in Section~5 resource data for subsequent experiments are described. Section~6 outlines the workflow of the proposed approach and Section~7 summarizes achieved evaluation results. Section~8 concludes the paper with a short discussion and an outlook on future work.

\section{Related Work}
\label{sec:related-hufflepuff}
\textit{}

\noindent
In this section, we first look into techniques used for measuring changes in KGs. Next, we look into methods used for fact-checking in knowledge bases. 
\subsection{Changes in KGs}

In \cite{Nishioka18}, the authors proposed a method to analyze the evolution of KG and is therefore highly related to our approach. They tried to address several aspects of the problem of keeping track of human editing in the KG and the real-world changes that happen. The approach was to train a classifier that was not relying on the editors' history but on topological features.
In \cite{jimenez-cuenca2011}, the authors present LogMap, which still holds up to the Ontology Alignment Evaluation Initiative challenge for KGs, and is an example for a First Order Logic-based Ontology matching approach. In \cite{ribon} proposed a new semantic similarity measure framework that aimed to analyze the evolution of knowledge graphs. The framework is considering several resource characteristics encoded within KGs semantically.  They have identified and considered the relevant resource characteristics (such as, neighbors, class hierarchy or attributes) to determine similarity values among each characteristic accurately. The framework is applied to three different tasks which is therefore related to our approach in measuring the semantic similarities based on entity neighborhoods and their representation in graph embeddings.

\subsection{Fact-Checking}
Fact-checking is a research area in which algorithms are designed to determine the truth of a claim using human curators, external knowledge sources, common sense rules or etc. Fact-checking techniques have several applications such as Fake News Detection (FND) or checking the trustworthiness of RDF triples.
Recently researchers started to address the issues of identifying false news shared on the web. The work on FND mainly relies on humans to check the validity of the news \cite{qian2018neural} or linguistic aspects of the given text \cite{volkova2018misleading,jiang2018linguistic,rashkin2017truth, mihalcea2009lie}. PolitiFact.com \footnote{\url{https://www.politifact.com/}} is an example of fact checking website were the organizers targeted only one domain (e.g generally politics) and report on the accuracy of statements made by public figures or journalists.  Linguistic aspects deception detection technqiues on the other hand rely on manually annotated which is again hard to obtain for different domains.

Some recent work has started to tackle the KG error detection using automatic approaches which leverage knowledge graph interlinks \cite{liu2015towards,liu2017measuring} or web sources\cite{gerber2015defacto}. 
In \cite{liu2017measuring} the authors rely on finding consensus from other knowledge graphs to validate the  information in the KG. Gerber et al in \cite{gerber2015defacto} presented a framework named DeFacto (Deep Fact Validation)\footnote{\url{https://github.com/DeFacto/DeFacto}} where the authors validate the facts in DBpedia using the information on the web sources. While Defacto has shown to perform well for multi-lingual and temporal fact validation , their framework has been shown to be limited to only 10 relations which shows it is not scalable to different domains and for extending that it relies on manually annotated training data.

\subsection{Knowledge Graph Embeddings}

A Knowledge graph embedding is a representation of a knowledge graph's component into continuous vector space. The idea is to ease the manipulation of graph components (entities, relations) for prediction tasks such as entity classification, link prediction or recommender systems. A survey of approaches and applications for knowledge graph embeddings was done by Wang et al. \cite{Wang18}. Two main approaches exist in order to learn knowledge graph embeddings from a KG: translational distance models where the goal is to minimize the distance between neighbors entities in the graph; semantic matching models which are based on the semantics of the graph components compute a similarity score that measures the semantic similarity between each entity in the graph.
Represented in a vector space, the Knowledge graph Embeddings represent semantic closeness of the entities and the relations in that vector space. In this work, Knowledge Graph Embeddings are being used to capture the semantic evolution and change that occur in a Knowledge Graph. In \cite{Aditya16}, the authors used random walk strategies to sample sequences of nodes and relations, and then applied skip-gram model in order to create embeddings of the samples. However, random walk strategies, if applied different times on the same knowledge graph, will create shifted space, and thus two embeddings of the same entity can not be compared. For this reason, we used the following models (known as 'translational models'):
\begin{itemize}
    \item TransE \cite{bordes13} which learns representations of entities and relations so that the distance between the $e_{1} + r - e_{2}$ tends to zero.
    \item TransH \cite{Wang14} which in addition to TransE algorithm enables entities to have different representations when they have different relations by projecting entities on a hyperplane.
    \item TransR \cite{Lin15} which enables entities and relations to be embedded in a vector space in different dimensions through a projection matrix associated to a given relation.
\end{itemize}

\section{Resources}
\label{sec:resources-hufflepuff}
\noindent
DBpedia describes itself as a a crowd-sourced community effort to extract structured content from the information created in various Wikimedia projects \footnote{https://wiki.dbpedia.org/about}. A lot of of research has been publish with or based on DBpedia throughout the years \cite{Auer07dbpedia:a}\cite{dbpedia_jws}.


For our evaluation we considered the last two versions of DBpedia knowledge graph; 2016-4\cite{dbpedia_4_2016} and 2016.10\cite{dbpedia_10_2016}.  The data is provided as RDF-Triples. For our experiment, we focus on one entity (\url{http://dbpedia.org/resource/Cristiano_Ronaldo}) in order to explore and detect the changes between the two versions of DBpedia knowledge graph.




\section{Approach}
\label{sec:approach-hufflepuff}
\textit{}

\noindent

This section describes an approach on how to evaluate the evolution of a Knowledge Graph by building a work-flow based on three distinct characteristics of a KG: its syntax, its structure, and its semantics. 

In order to evaluate changes and their severity between two version of a knowledge graph we suggest an ensemble method that consists of several stages to measure changes between the syntax, the structure, the semantics and ends with comparing one or more facts that are reflected in the KG with the same fact as it can be found in the world wide web. To simplify our actual experiment we will use an excerpt of two DBpedia version dumps, as described in section 5.

\begin{figure*}[t]
	\centering
	\includegraphics[width=\textwidth]{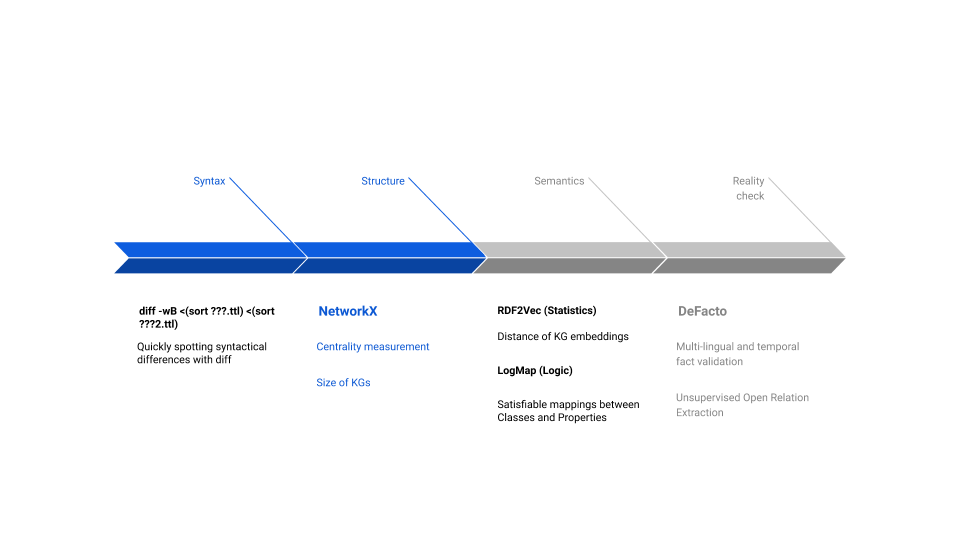}
	\caption{Workflow
	}
	\label{fig:archi}
\end{figure*}

The syntactic difference between two KGs can be captured by looking at their RDF representations. The intuition is that if two RDF representations are different, then the semantics of the KG has changed and capturing the structural changes (namely, how the connections in the KG have evolved) can give insights on it. 
The work-flow that is proposed is the following: the syntactic changes are captured by calculating the $disjunction$ of the $KG_{1}$ and $KG_{2}$ RDF triples.

\subsection{Preliminaries and Problem Definition}
In this section, we first provide definitions of some useful concepts. Then, we define out problem.

\begin{itemize}
\item{Knowledge Graph:}  A knowledge graph is defined as a set $K = (E, R, O)$ where $E$ is the set of entities, $R \subset{E \times \Gamma \times E}$ is a set of typed relations among entities, and $O$ is an ontology, which defines the set of relation types ('properties') $\Gamma$.

\item{Problem Formulation:} 
We define a Knowledge graph Change as follows:
\begin{itemize}
\item A change is represented by the tuple $ (e_{1}, r, e_{2}, e, s, t)$
where $ (e_{1}, r, e_{2}) $ is a triple added or removed, $e$ is an integer $\in {1,2}$ which represents (addition or deletion), $t$ represents the time where the change occurred, and $s$ is a similarity score between the two versions. 
\end{itemize}
\end{itemize}

\subsection{Measuring the Syntactic change}
In order to measure the syntactic changes, we use the GNU tool diff with parameters to ignore line changes, as we are only interested in string differences. The data format is in N-triples. We sort the triples first before extracting the differences to have the entities in both versions in the same order. The output of the extraction lights the syntactic changes.


\subsection{Measuring the Structural Change of a KG}
Knowledge Graphs are basically labeled graphs, i.e. changes in $KGs$ also might affect their underlying graph structure. There exist a large variety of (aggregated) measures to characterize the main properties of a graph, which might also be useful to characterize changes in $KGs$. Let $KG=(V,E)$ a Knowledge Graph with $V$ the set of vertices and $E\subseteq V \times V$ the set of edges between the vertices.

\begin{itemize}

\item \textbf{Subgraph inclusion} Let $KG_{1}=(V_{1},E_{1})$  and $KG_{2}=(V_{2},E_{2})$. It holds that $KG_1\subseteq KG_2$, iff $ V_{1} \subseteq V_{2} $ and $E_{1} \subseteq E_{2}$, that is to say a graph $KG_1$ is a subgraph of $KG_{2}$, if $KG_{1}$ is included in $KG_{2}$.

\item \textbf{KG Diameter} The diameter of a $KG=(V,E)$ is given by $max_{u,v}d(u,v)$ where $d(u,v)$ is the distance between any two vertices $u,v\in V$. In other words it is the largest number of edges that have to be traversed in order to travel from one node to another.

\item \textbf{Node degree} For a knowledge graph $KG=(V,E)$, the in-degree of a node $n_{in}(v)$ with $v\in V$ corresponds to the number of in-going edges in $v$. The out-degree of $n_{out}(v)$ with $v\in V$ is the number of out-going edges. Thus the ratio of in- and outgoing nodes is calculated by
$\frac{n_{in}(v)}{n_{out}(v)+1}$.
In general this enables to determine a centrality measure of a node but it does not take into account the impact of all the others nodes in the graph.

\item \textbf{Node Eccentricity} the eccentricity $e(v)$ of a node $v\in V$ is the maximum distance from $v$ to any other node $w \in V$, i.e. $e(v)=max_{w \in V} d(v,w)$.

\item \textbf{Graph Radius} the graph Radius $r(G)$ is the minimum eccentricity of any $v \in V$, i.e. $r(G)=min_{e(v) \in V}e(v)$.

\item \textbf{Pagerank} an alternative way of calculating the centrality of a node is the PageRank \cite{Pageetal98}. Instead of measuring the local impact of a node in a KG, \textit{PageRank} allows to consider its global impact on all the graph. This is done by the iterative propagation of the in-edges and out-edges impact.

\item \textbf{HITS} is interesting in the same way, but it is slightly different than PageRank as it calculates \textit{Hub} and \textit{Authority} scores instead of just one measure \cite{Kleinberg:1999:ASH:324133.324140}. This enables to distinguish the nodes that have a high influence on their neighbours (\textit{authority} scores considers the edges that are outgoing from a node) and the nodes that are highly influenced by the others (\textit{hub} considers the edges that are incoming to a node)

\end{itemize}

These measures being applied on two versions of a KG can be used to enable an overview of the structural evolution of a KG. After having identified already the syntactic changes, they allow to evaluate how relevant are these changes with respect to the importance of the nodes involved in it. The following sections explain what insight these measures can give for the comparison of two $KGs$.

\paragraph{Distribution of the nodes and edges} 
A graph based property to structurally describe the entire KG that is typically investigated is the distribution of the node degrees within the KG. We recall that the degree of a node is the number of neighbours of the node. This distribution is usually a long tail distribution, where the number of nodes per edges decreases following a power law.

\paragraph{Subgraph inclusion}
A first rough approach is to test whether a $KG_1$ (at $t$) is included in a $KG_2$ (at $t+1$). If a $KG_{1}$ is included in a $KG_{2}$, then $KG_1$ is a enhancement of $KG_2$ but the content of the previous KG is still present in the new one.

\paragraph{Node degrees}
One interesting point to consider is the characteristics of the nodes that are added and deleted in the KG. In general, $n_1$ in $KG1$ is similar to $n2$ in $KG2$ if their neighborhood (e.g the nodes that are surrounding it) are the same. Thus, studying the node degree allows to evaluate how similar are $KG_{1}$ and $KG_{2}$ with respect to the centrality of the node. Having a look on the degree of these nodes gives a measure of the importance of the nodes that have changed from one to another KG. Thus, the $n_{in}(v)$ and $n_{out}(v)$ are computed for the new nodes, and the centrality of a new node $v$ is given by $\frac{n_{in}(v)}{n_{out}(v)}$.

\paragraph{PageRank and HITS} These algorithms are also used to capture the centricity of a KG. Instead of just capturing the centrality of a node with respect to its neighbourhood, using these measures enables to have a more fine grain description of the impact of a node to the entire KG. 

\subsection{Measuring the Semantic Change in a KG}

When measuring the difference between two versions of a KG it should be considered that some changes might have high impact over the semantics represented within the KG even though these changes cause only small topological changes. In this case a topological-only based measurement might not be able to capture important differences between two knowledge graphs.

To illustrate this consider the addition of two triples $t_{1}=(s_{1},p_{1},o_{1})$ and $t_{2}=(s_{2},p_{2},o_{2})$ to a KG and consider that $s_{1}$ is semantically similar to $o_{1}$ while $s_{2}$ is semantically different than $o_{2}$. In this case it is expected that $t_{2}$ will produce a higher semantic change because the involved entities are less similar in this sense. A topological-only based measure might not be able to express that $t_{2}$ produces a higher impact than $t_{1}$, since there is no explicit distance difference between them.
Therefore, there is no guarantee that a topological-only based measure will reflect neither the expected differences between these changes nor the new versions produced by them.

To address this problem we propose a method to measure the semantic difference between two versions of a KG based on entity neighborhoods and their representation in graph embeddings. Inspired by graph theory, we consider that ``two nodes are similar if their neighborhoods are also similar"\cite{Koutra11algorithmsfor}. The main difference between our approach and existing works in graph theory is that we consider both the structural and semantic difference between two neighborhoods.

Given two versions $v_1$ and $v_{2}$ of a KG and assuming that $v_{1}$ is a current and stable version, the first step of our method is to create a graph embedding for $v_{1}$ . Then, for each node in $v_{1}$ extract its neighborhood $N_{n\_v1}$ and its respective neighborhood  in $v_{2}$  $N_{n\_v2}$, where the neighborhood $N_{n\_v1}$ is the set of all nodes directly connected to $n$ in version $v_{1}$. The next step is to extract from the graph embeddings the list of vectors that represents  $N_{n\_v1}$ and $N_{n\_v2}$. 
Finally we calculate the cosine similarity between the embeddings representing $N_{n\_v1}$ and $N_{n\_v2}$, and the obtained cosine similarity represents the semantic difference between the respective node in each version. Equation \ref{semantic_sim} shows how to calculate the semantic similarity of a node in two different versions if a knowledge graph.

\begin{equation}
\label{node_sim}
node\_sim(n)=cossim(GE(N_{n\_v1}), GE(N_{n\_v2})) 
\end{equation}

where $GE(N_{n\_v1})$ is the neighborhood representation of a node $n$ in the graph embeddings for the version $V_{1}$ and $cossim$ is the cosine similarity.

The semantic similarity between two version  can then be calculated by taking the average of the nodes' similarities, as shown in \ref{semantic_sim}. 

\begin{equation}
\label{semantic_sim}
semantic\_sim(V_{1},V_{2})= \frac{\sum_{i=1}^{|n|} node\_sim(n_{i})}{|n|}
\end{equation}

Measuring the relatedness among entities in $KGs$ is another way to measure the semantic differences/similarities between two versions of $KGs$ which we also built our method based upon it. We have followed the approach conducted by Morales et al. 2017 (in \cite{morales_collarana_vidal_auer_2017}) which is known as MateTee. The mentioned approach aims to compute values of similarity between entities in $KGs$. 

Assuming we have two versions of a KG, in each one of these versions, to measure the similarity between any pair of entities belonging to a Knowledge Graph, the next steps would be followed:

\begin{enumerate}

	\item  First, learning the embeddings of entities by analysing the connectivity patterns between entities in a KG.
	
	\item  Then, encoding these patterns into their vector representation, i.e., their embeddings. To conduct this encoding, we would use the same MateTee approach which is formally defined as follows:
	
	Given a knowledge graph G = (V, E) composed by a set T of RDF triples, where $V = \{s | (s, p, o) \in T\}\cup \{o | (s, p, o) \in T\}$ and $E = \{p | (s, p, o) \in T\}$, we would be able to find a set M of embeddings of each member of V , such that:

 \begin{multline}
arg min_{m_1,m_2\in M} Error(M) = \\ arg min_{m_1,m_2\in M} \sum_{m_1,m_2\in M} |S_{1}(m_{1}, m_{2}) - S_{2}(m_{1}, m_{2})| 
\end{multline}
    
	where $S_{1}$ is a similarity metric computed using Euclidean distance, and $S_{2}$ is a similarity value given by the Gold Standards. The Gold Standards are the values considered as ground truth. Ground truth is the values accepted by the scientific community because they were calculated manually with deep domain expertise, e.g, Sequence Similarity in the Gene Ontology domain.
	
	Note that: Learning the embeddings of entities means that similar entities in the KG should be also close in the embedding space and dissimilar entities in the KG should be also far in the embeddings space. 
	
	\item Finally, computing the similarity measure of both entities using the following formula:
		\begin{equation}	
            similarity(A,B)=\frac{1}{(1 + EuclideanDistance(A,B))}
        \end{equation}

\end{enumerate}

Thus, by achieving such calculation in each version of KG, we would be able to observe the changes of the entities within the KG semantically by analysing their relatedness.

\subsection{Measuring the Semantic Change in a KG based on First Order Logic}

As promising as Knowledge Graph Embeddings haven been in recent years to help with the Ontology Matching Problem, we still have to deal with problems related to deep learning, mostly the interpretability \cite{GOYAL201878}.
One alternative or rather complement in our case is focusing on ontology  matching methods that are based on first order logic, for example LogMap \cite{jimenez-cuenca2011}. It was still used for the most recent Ontology Alignment Evaluation Initiative (OAEI) in 2018 \footnote{\url{http://oaei.ontologymatching.org/2018/results/knowledgegraph/index.html}} and was one of only 6 other systems (2 of them are variations of LogMap) that could complete the task in the section Knowledge Graph. 

LogMap is indexing the input ontologies lexically and structurally to calculate initial anchor mappings and assigns confidence values to each of them. Then it starts an iterative process of detecting and repairing mappings with an ontology reasoner and a “greedy” diagnosis algorithm. The version 3 of LogMap, the most recent, is fully relying on OWL API 4.

As we want to quantify changes between knowledge graph versions and differences between knowledge graphs, we will take only the number of class and property mappings of the output of LogMap into account.

\subsection{Knowledge Graphs and the Reality}
The goal of this section is to present a way to measure how well a KG represents the reality.
An important problem in the life-cycle of a KG is to investigate ways to determine the trustworthiness of a claim in the KG and adapting the KG to the changes in the real world \cite{esteves2018toward,pasternack2013latent}.
The idea is given a fact represented as an RDF triple as input we measure a confidence value for this fact to check if the claim represents the reality.

The task of validating the facts in KG is often addressed by human curators where they are asked to use standard search engines and find relevant documents to find information about the fact. Human computation techniques such as Crowdsourcing \cite{acosta2013crowdsourcing} and Game-based approaches \cite{hees2011betterrelations} are other techniques used to involve user in the KG error detection. Although involving the humans in detecting false information increases the accuracy of the predicted items, its costly and time-consuming to collect the predictions for different domains \cite{gerber2015defacto,paulheim2017knowledge}.  

As discussed in the related-work section DeFacto is a well known open-source approach that supports simple RDF triple fact-checking. In Defacto the authors first transform the statements in DBpedia to natural language sentences and pass them to the Google search engine to find web pages containing those sentences.  Next they assigns low confidence score to the facts that appear in no or only very few web pages. One of the most important advantages of the DeFacto framework is its ability to tackle the problem of temporal information where a relationship is considered correct for just for a certain period of time (e.g., Barack Obama, president of, USA 2009-2017). However due to its reliance on several NLP tools such as BOotstrapping linked datA (BOA) \cite{gerber2012extracting}(i.e, a supervised machine-learning approach) it only supports 10 relations (i.e, award, birth, foundation, subsidiary, starring, death, nbateam, publication, leader and spouse) which leads to scalability problem.

To address this issue, we propose to extend the Defacto framework for the cases when the relation is other than the one supported by the library. We make us of the existing techniques for unsupervised Open Relation Extraction (ORE) \cite{elsahar2017unsupervised}. The idea in ORE is to discover arbitrary semantic connections between entities in unstructured texts \cite{culotta2006integrating}. 
Given an input sentence such as \textit{``Turing was born in England"} and two entities like \textit{$<$Turing, England$>$}, an ORE system should extract a sub-string which entails the semantic relation between the two entities (i.e. \textit{``was born in"}).

Given a RDF triple we follow the steps below:
\begin{itemize}
\item \textbf{RDF to Sentence}: we transform a RDF triple to a natural language text, by concatenating the entities and the relation in between. As an example we take the $<$Cristiano Ronaldo, clubs, Real Madrid C.F. $>$ triple and change it to \textit{"Cristiano Ronaldo club Real Madrid C.F"}.
\item \textbf{Retrieving Web Pages}: we give the aforementioned input sentence as an input to Google search to retrieve web pages which are relevant for the given sentence. 
\item  \textbf{Unsupervised Open Relation Extraction (UORE)}: we run the UORE technique proposed in \cite{elsahar2017unsupervised} to extract the relations from the unstructured text of the $n$ (e.g. 5 ) top retrieved Google pages. 
\item \textbf{Confidence Score}: inorder to check if any of the extracted relations matches the relatrion in the RDF triple we measure the semantic similarity between the extracted relations $EXRel$ and the input RDF relation $INRel$ and if the similarity value is higher than a given threshhold (e.g. 90\%) we consider it as a true fact. For measuring the semantic measures we use the Fast-ext word embeddings which are  trained  on  common  crawl  and  Wikipedia
Dataset and are available for multi-languages \footnote{\url{https://github.com/facebookresearch/fastText/blob/master/docs/pretrained-vectors.md}}

\end{itemize}

    


\section{Evaluation and Results: Use case Proof of concept - Experiments}
\label{sec:evaluation-hufflepuff}
\textit{}

\noindent
Due to time constraints we could not fully implement and execute the test setup. The following list is therefore only a plan and future work is needed to fully evaluate our approach. The first step could however be applied and we will therefore include these partial results.

\begin{itemize}
    \item Syntax: To measure the syntactical changes we used the GNU tool diff. We ran it locally on two different versions of N-Triples which contain all the triples about one specific entity (\url{http://dbpedia.org/resource/Cristiano_Ronaldo}). With this first overview, we can have a closer look at the structure of the concerned classes and properties.
    
     \item Structure: (NetworkX to check specifically the relevant classes and properties within the whole graph/ontology of dbpedia, two versions)
     
     \item Semantics (Statistics): We planned on using RDF2Vec to create embeddings in vector spaces of the concerned graph excerpt and then calculate the euclidean distance to measure a semantic difference.
     
     \item Semantics (Logic): With LogMap we planned to a look how much changes in the formal semantics can be measured at the specific section. As DBpedia does not rely on OWL description logic (DL) we would have had to modify the files accordingly
     
     \item Reality check: With the DeFacto framework there is an existing famework to automatically validate facts with the help of the world wide web.  As many facts are time-constraint we expect the newer version of DBpedia to perform better here. 

    \item As a concrete example we chose to analyse the part of DBpedia that deals with the football player Christiano Ronaldo to check if we can measure the severity of eventual changes on all aforementioned levels and if one of the two graphs (excerpts) and their facts is closer to reality.
\end{itemize}

\subsection{Results}
\noindent

We used the GNU command \textbf{grep} and the string "Christiano Ronaldo" to make a pre-selection and limited the dump to the dataset ``Infobox Properties Mapped'' of DBpedia 2016.10 and 2016.4 in the format TTL\footnote{\url{https://wiki.dbpedia.org/downloads-2016-10}} \footnote{\url{https://wiki.dbpedia.org/dbpedia-version-2016-04}}. Then we used diff as follows to sort the ttl-files, but to not show line differences snf only string differences: \texttt{``diff -wB $<$(sort file1.ttl) $<$ (sort file2.ttl)''.}
 What we could see was already a long list of difference, and we could only speculate about the reasoning. For example DBpedia 2016.10 included the following triple:
 \begin{verbatim}
 <http://dbpedia.org/resource/Cristiano_Ronaldo> 
   <http://dbpedia.org/property/birthDate> 
     ``1985-02-05''^^<http://www.w3.org/2001/XMLSchema#date>
 \end{verbatim}
  whereas this was not included in DBpedia 2016.04. All in all we could observe 161 triples that were different in total. As we only looked at the mapped infobox properties it could of course be that some of the data has been represented in different ways and at different nodes.

\section{Discussion and Conclusions}
\label{sec:conclusions-hufflepuff}

\noindent

In this article, we explain an eventual solution of existing problems concerning KG evolution and their validity in terms of modeling reality. Our approach shows that there are existing and maintained methods that have not been combined in this way before and we are convinced there is even more potential once a better alignment is reached between the different steps, which would require own software implementations based on the theory we have about KG evolution and changes. We were not able to implement our approach because of workspace and time limitations.


For future work, we strongly recommend adding a level which verifies how different versions or different KG are in accordance with the FAIR principle \cite{wilkinson2016fair}. This has been omitted in this project only due to the scope. We think it is a crucial point to include a quantification in this regard in order to keep the data in KGs reusable for the scientific community. 

\part{Empirical approaches for observing Knowledge Graphs evolution}
\label{part2}
\chapter{Measuring ontology evolution for supporting reuse}
\label{sec:gryffindor}
\chapterauthor{Martin Beno, Felix Bensmann, Harm Delva, Aneta Koleva, Martin Mansfield, Kader Pustu-Iren, Valentina Presutti}

\section{Introduction}
\label{sec:intro-gryffindor}
\noindent



Ontologies support knowledge sharing and reuse. They are widely used in both academia and industries. 
Because of the explicit semantic conceptualizations and reasoning capabilities that ontologies offer,  they constitute the backbone of the Semantic Web. Consequently, with the constant evolution and enrichment of the Semantic Web, the evolution of ontologies is inevitable. Since ontologies can be used independently as well as interdependent, observing the changes and identifying their cause can be of crucial importance for the design and integrity of the Semantic Web. Noy and Klay \cite{Noy&04} discuss that the following three types of changes can occur in an ontology: (i) changes in the domain; (ii) changes in the conceptualization; and (iii) changes in the explicit specification. The impact of these changes have severe consequences not only on tooling but foremost on datasets. It is important that versioning and updates of ontologies are executed in a controlled way such that they do not break datasets. Therefore, it is crucial to understand triggers and consequences of changes. Thus, we recognize the need of well defined metrics which would capture these changes. To analyze ontology changes, we identify patterns of changes in the number of sunglasses, concepts etc.  and measure their impact over time.

In this report, we present our contributions: (i) we provide an analysis of the evolution of two families of ontologies, (ii) we propose different metrics for observing the effects of evolution which should help in better understanding of the changes; and (iii) we present the results of the conducted experiments with the intention to demonstrate how ontologies evolve. 

The rest of this paper is structured as follows:
First we present the research questions which we try to address. Second, we describe the problem of knowledge graph evolution and ontology evolution. We then discuss several approaches as related work. In Section 5 we explain the used resources. In Section 6 we elaborate on our proposed approach and in Section 7 we discuss the evaluation of this approach. We conclude this report with Section 8 in which we provide discussion about the main findings of this work and some final conclusions.

\section{Research Questions}
\label{sec:rq-gryffindor}
The focus of this report are the following research questions:

\begin{description}
    \item[Q1] What changes can be observed in ontologies?
    \item[Q2] How can they be characterized?
    \item[Q3] Is there correlation between the changes that occur in two different ontologies that have a dependency?
\end{description}

\section{Knowledge Graphs Evolution and Preservation}
\label{sec:def-gryffindor}

In the field of knowledge graph evolution this report focuses specifically on the evolution of ontologies. We define Knowledge Graph Evolution as the process of KGs changing over time. Considered changes include nodes and edges being added or removed.
Ontology evolution is defined by Maedche et al. \cite{Definition} with the following definition :
\\
\textit{“The  timely  adaptation  of  an  ontology  and  consistent  propagation  of  changes to  dependent  artifacts.”}
\\
However, in this report we only regard ontology evolution as a process we aim to observe and understand. In particular we focus on the effect certain ontology changes have on interdependent ontologies.


\section{Related Work}
\label{sec:related-gryffindor}
\noindent

Ontologies are essential for the design and the integrity of linked datasets. Starting from the design phase of the ontology, multiple factors affect their value of use. In \cite{Onto-evol} the authors describe the conditions of the creation and the development of ontologies as a concept in the semantic web. 

As domains and conceptualisations change this adds further complexity to the operation and maintenance of ontologies. Particularly, Stojanovic et al. \cite{Stojanovic&02} argue for methods to provide ontology maintainers with means to control and customize evolution.

Several works are currently available that target research on ontology changes and their impact. The following works relate to this paper in the way that they also study change patterns and entailing consequences for specific tasks.
In an early paper on this topic Noy and Klein \cite{Noy&04} examine ontology evolution by analyzing the operations from two versions of one ontology. They propose a set of dimensions to consider when determining whether a new version of an ontology is compatible with the old one.
Another approach by Yang et al. \cite{2nd_paper_valentina} introduces a metrics suite for assessing the complexity of ontologies from the evolution point of view. The metrics examine concepts, relations and paths. They are verified on an exemplary gene ontology. They constitute an alternative approach leveraging alternative metrics to the ones discussed here.
More specific work on ontology evolution was documented in Dragoni et al. \cite{dragoni2012evaluating}, where the authors examine the impact of evolution on search tasks and tools, while taking a look at the introduction of versioning to support user. The approach of detecting change patterns and examining consequences for real-world tasks is similar to the one presented here but focuses specifically on the task of search.

Apart from the ontology evolution itself, relevant work is discussed by Klein et al. \cite{Klein&01} where building blocks for specific aspects of a versioning mechanism are explained.

We also examined the evolution of knowledge graphs, as opposed to ontologies, in order to determine if it is possible to reuse some approaches.
A relevant work is the Dynamic Linked Data Observatory which carries out and examines snapshots of (Linked Open Data) LOD cloud.  K{\"a}fer et al. \cite{dyldo} conduct similar studies on instance level whose metrics were partly adapted by us for reuse.
\cite{KG-evo} also focuses on evolution of KGs over time. The analysis is carried out with the intention to use it for predicting changes. Using 25 snapshots of Wikidata the authors observe change over time between two successive snapshots. The focus is on the topological features, in particular the number of nodes and edges.
\cite{paper3} on the other hand focuses on structural changes.
Further work is concentrated on the detection and assessment of changes.
\cite{know-evolve} describes work on learning and predicting changes in frequently changing graphs by creating an evolutionary knowledge network that learns non-linearly evolving entity representations over time using embeddings.

\section{Resources}
\label{sec:resources-gryffindor}
In order to analyze the effects of evolution on interdependency we focus on two families of ontologies in specific. The first one is the Data and Analytics Framework (DAF)\footnote{https://github.com/italia/daf} which was created to support the management of public administration data. It was developed for the Italian government to serve as a base for domain-specific ontologies. The second family of ontologies we use was created for the ArCo (Architecture of Knowledge)\footnote{http://wit.istc.cnr.it/arco/} project. This project's goal is to create a knowledge graph of Italian cultural data from the general catalogue of the Italian Ministry of Cultural Heritage and Activities (MiBAC). The ArCo project has lead to the creation of several ontologies that refer to concepts defined in the DAF ontologies, and it is this dependency that makes these ontologies particularly interesting for our use case. 

Both projects use a versioning system which allows us to look at their evolution over time. However, neither project makes historical versions easy to access. DAF's Github repository contains folders for each version, this is good enough for our use case but it is not ideal for usage in knowledge graphs. Older versions of ArCo can also be found on its Github repository, but only if you can find the specific commit that corresponds to a historical version. It is also interesting to note that according to DAF's documentation on Github, the ontologies are still unstable but they expect them to stabilize by March 2018 -- which is well over a year ago at the time of writing.

Apart from the ontologies that are part of these two projects, we also used ontologies from the BioPortal\footnote{http://bioportal.bioontology.org/ontologies} and the Linked Open Vocabularies (LOV)\footnote{https://lov.linkeddata.es/dataset/lov} projects. Both of these projects index existing ontologies, along with their historical versions. We used ontologies from these indexes to guide our intuition while developing our proposed approach.

Because of time constraints we were not able to fully develop our own ontology analysis tools. Instead, we rely on Bubastis\footnote{http://www.ebi.ac.uk/efo/bubastis/} to find logical differences between versions of the same ontology. Other similar tools such as Ecco\footnote{https://github.com/rsgoncalves/ecco} exist, and have been tested, but Bubastis was by far the easiest to set up. 


\section{Proposed approach}
\label{sec:approach-gryffindor}

\noindent

By consideration of possible use cases for a framework to monitor changes in linked data over time, \citeauthor*{Kafer&12}~\cite{Kafer&12} identify a set of empirical questions that such a framework should help to answer. We propose that an approach to monitoring evolutionary change of an ontology should address similar concerns, tailored to the requirements specific to ontological changes. We propose that a framework for monitoring evolutionary change in ontologies should include metrics to describe the observation of:

\begin{description}\setlength\itemsep{1em}

    \item [Frequency of change.] A metric for the rate at which an ontology evolves should describe how many distinct versions of an ontology are produced over a period of time.

    \item[Patterns of change.] A metric to distinguish types of ontology evolution should categorise changes by consideration of their impact on the ontology. A comprehensive collection of possible ontology changes and their impact is identified in \cite[Table 1]{Noy&04}.

    \item[Degree of change.] A metric to describe the amount by which an ontology has changed between versions should describe the extent of the impact of each change type.
    
\end{description}

\citeauthor*{2nd_paper_valentina}~\cite{2nd_paper_valentina} formalise metrics for measuring ontology evolution in terms of complexity, by consideration of the number of concepts and relations.

By reusing elements from related ontologies, an ontology becomes vulnerable to corruption by changes in another ontology. As a result, we propose that the evolution of related ontologies is not independent, but aspects of evolution might be induced by external changes. We propose a set of metrics for measuring such change, to aid the management of evolution of ontologies which reuse elements from related ontologies.

\begin{description}\setlength\itemsep{1em}

    \item[Evolutionary Synchronisation.] A measure of how closely aligned the evolution of distinct ontologies are, in terms of the temporal similarity of changes. Changes in an ontology might be considered induced rather than independent if they occur within some defined duration. This duration might be defined empirically, based on some knowledge of the ontologies and their evolution (e.g. observation of frequency of change). The Evolutionary Synchronisation (ES) of two ontologies is evaluated by consideration of the temporal similarity of corresponding changes to each ontology. A timestamp $t(C_{O})$ describes the point at which an update is made to ontology $O$, where synchronisation of updated to ontologies $O1$ and $O2$ is expressed as $ES = \abs{t(C_{O1}) - t(C_{O2}) } $. For a pair of updates to be considered synchronised, they should occur within some threshold $T$, such that $ ES \leq T $.

    \item[Change Alignment.] Where evolutionary synchronisation of related ontologies suggests that change is induced by a dependency, further confidence that change is externally induced can be based on evaluation of the alignment of changed elements. Alignment of elements might be explicitly modelled, or might be inferred by analysis.

    \item[Evolutionary Dependency.] A measure of how reliant the evolution of an ontology is on related resources. This might be expressed as a ratio of changes considered induced by an external change to changes determined to be independent to the needs of the ontology, i.e the Evolutionary Dependency (ED) of a given ontology is defined $ED = \frac{EC}{SC}$ for externally induced changes (EC) and ontology-specific changes (SC). 
\end{description}


\section{Evaluation and Results: Use case/Proof of concept - Experiments}
\label{sec:evaluation-gryffindor}
\begin{figure}
\centering
\includegraphics[width=0.7\linewidth]{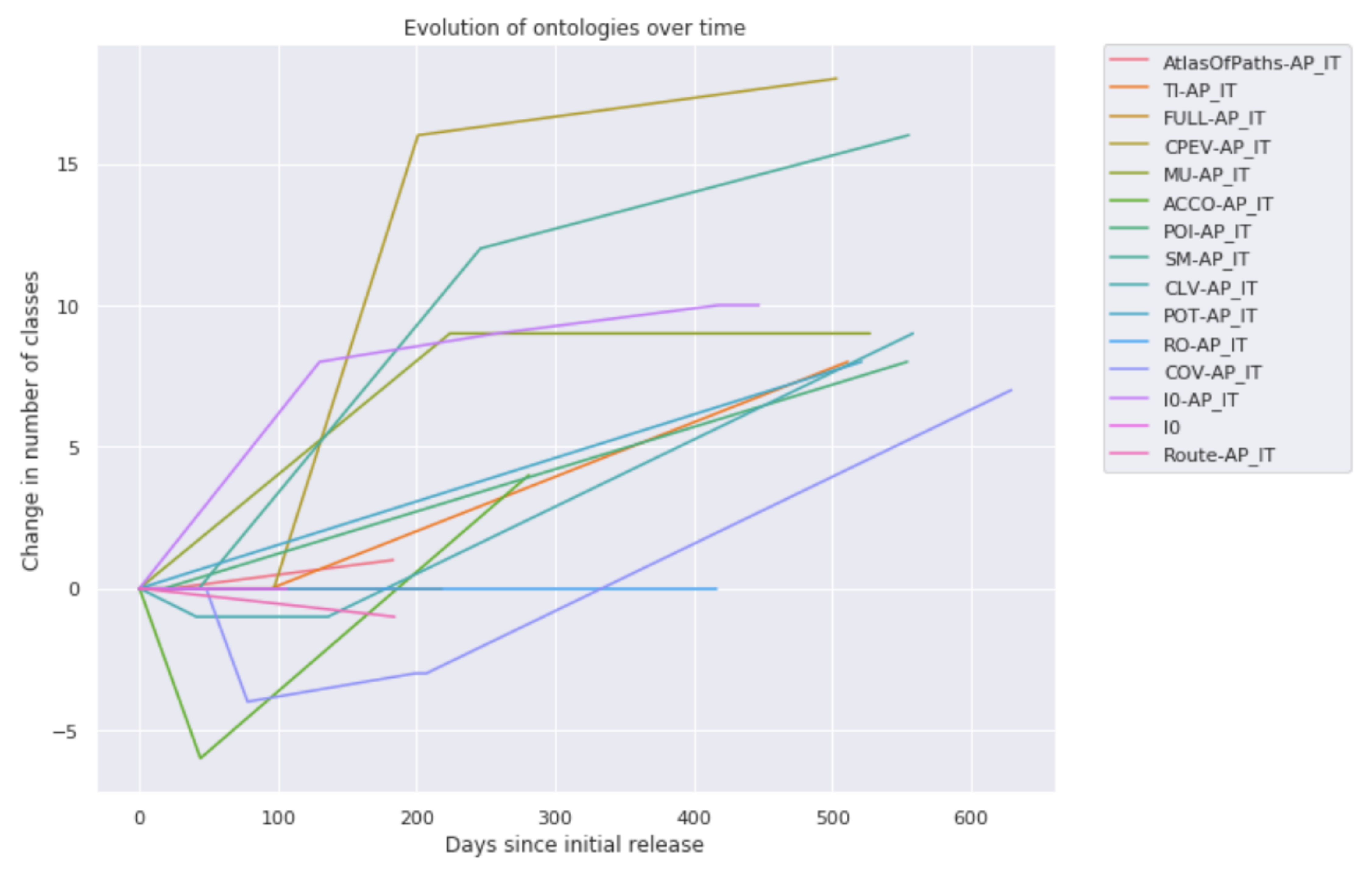}
\caption{Evolution of the amount classes over time}
\label{fig:class}
\end{figure}

\begin{figure}
\centering
\includegraphics[width=0.7\linewidth]{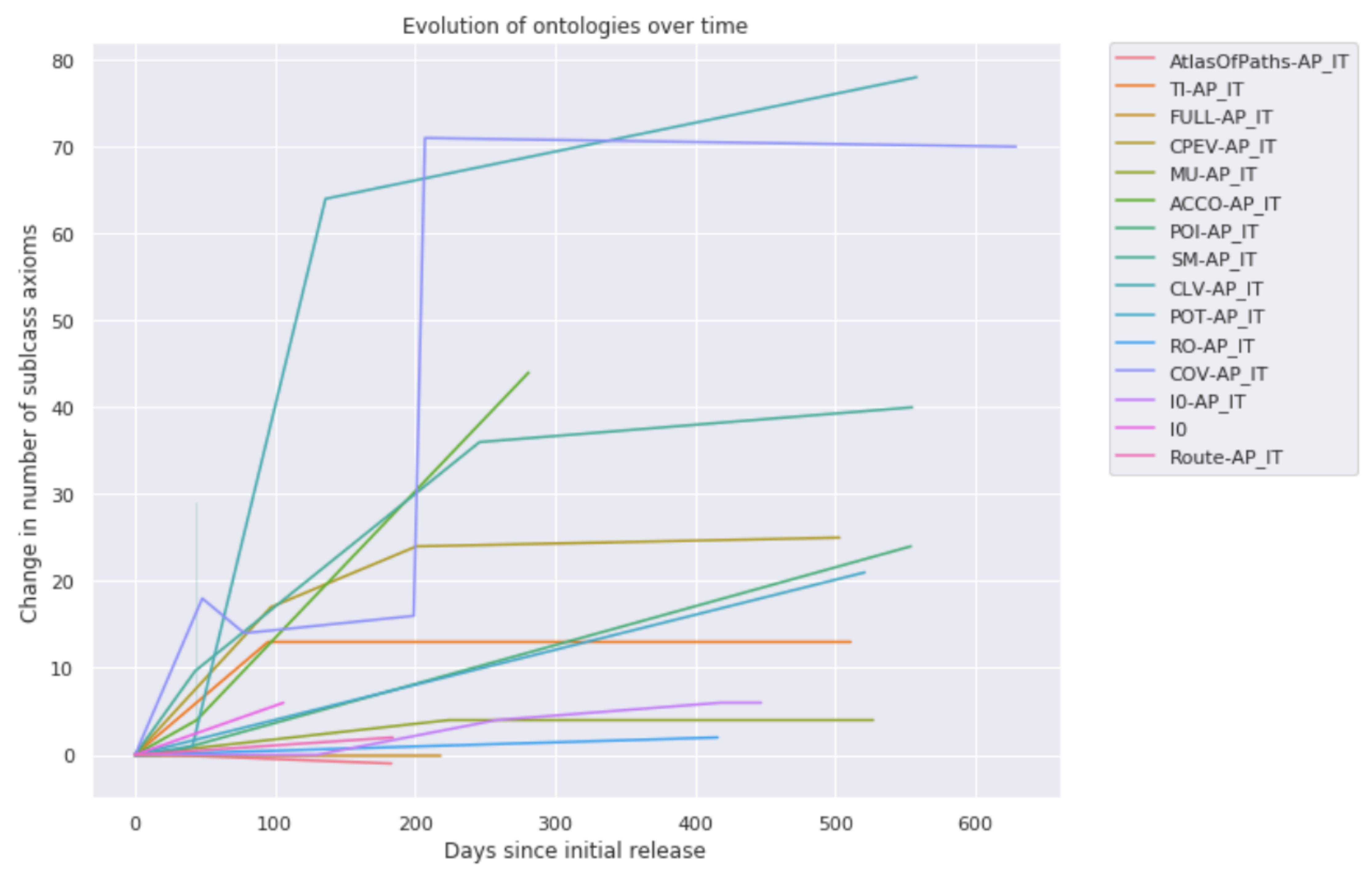}
\caption{Evolution of the amount subclass axioms over time}
\label{fig:subclass}
\end{figure}

We are interested in the evolution of the ontologies over time, so we compare every historical version of the ontologies to their first version and see how they have evolved over time. We have selected a few ontologies from the DAF and ArCo projects, limiting ourselves to the ontologies that have several versions and that contain modification timestamps. Figure \ref{fig:class} shows the evolution of the amount of classes over time. The x-axis shows the amount of days since the initial version of each ontology, while the y-axis shows the difference in amount of classes since the first version. We can see that it is not uncommon that class definitions are removed from an ontology, but the amount of classes does seem to increase over time. Figure \ref{fig:subclass} is similar, but the y-axis now shows the difference in the amount of subclass axioms over time. We notice that the amount of subclasses does not decrease nearly as drastically as the number of classes. 

\section{Discussion and Conclusions}
\label{sec:conclusions-gryffindor}
The present paper set out to investigate the evolution in ontologies using a set of predefined metrics. We intended to analyze a large set of ontologies, however it was surprisingly difficult to find historical versions. Furthermore, older versions rarely have a persistent URI. Nevertheless, we were able to analyze changes in patterns in the DAF ontologies, one of the very few ontologies with multiple older versions published in a public repository. 
\\
\\
One of the main weaknesses of the present study is therefore the small sample size. We cannot with confidence state, that our results could be extrapolated to other ontologies as well. Future work should involve running our evaluation framework over several different ontologies.

\begin{itemize}
    \item We need to take dct:modified seriously, and I know we are all guilty of often neglecting it.
    \item We found no evidence that updating an ontology causes updates in its dependencies -- although more data points are needed. 
    
\end{itemize}
\chapter{Pimp my Pie - Knowledge Graph Evolution from an NLP Perspective (and food)}
\label{sec:deloreans}
\chapterauthor{Mortaza Alinam, Wouter van den Berg, Lientje Maas, Fabio Mariani, Eleonora Marzi, Tabea Tietz, Marieke van Erp}

\section{Research Question}
\label{sec:rq-delorean}

How can the evolution of a concept from the real world as it is described in unstructured natural language text be represented in knowledge graphs?

\noindent In this work, the evolution of the concept \textit{apple pie} is discussed as a use case.


\section{Knowledge Graphs Evolution and Preservation}
\label{sec:def-delorean}

When defining the \textbf{evolution} of knowledge graphs (KGs) in the context of natural language processing (NLP), two perspectives are essential:
\begin{enumerate}
    \item The first perspective refers to natural language text determining the content of a KG through NLP. Here, evolution is defined by the text itself. Hence, historical text as the mirror to societies in a past reality thereby defines what is being modeled in a KG. In this perspective, NLP is an essential part of the process of KG evolution.
    \item The second perspective assumes that a KG is created or evolves independently of automated NLP processes. In this case, evolution means that classes, instances and values are created or altered by a source outside of the reality a text was authored or analyzed in. In this case, NLP is not part of the initial process, but applies whatever reality is defined in the KG to its source text. 
\end{enumerate}


\noindent Knowledge graph \textbf{preservation} in the context of NLP means that data about natural language text resources is preserved using knowledge graphs even if the original source ceases to exist. This refers especially to text resources that do not originate in digital form. In this work, we are focusing on knowledge graph evolution rather than preservation, even though it is acknowledged that in the relationship between KGs and NLP problems of preservation are presented.

\section{Introduction}
\label{sec:intro-delorean}


Knowledge graphs are graphs of data with the intent to compose knowledge. Thereby, \textit{composing} refers to a continual process of extracting and representing knowledge that enhances the interpretability of the resulting knowledge graph~\cite{Bonatti2019}. KGs represent what we consider true about (part of) the world. KGs are created at a certain point in time and can often be considered snapshots of the real world, i.e., they are essentially static~\cite{Mayesha2019}. However, \textit{``Knowledge lives. It is not static, nor does it stand alone"}~\cite{Bonatti2019}. We live in a world with infinite variation and variability. In other words, concepts continuously change over time and can vary between social contexts and locations. 

In Named Entity Linking, an attempt is made to map textual mentions to their representation in a knowledge graph. However, especially when analyzing historical text documents, the text often links to a knowledge graph that was not created in the same time period as from which the text originates (or even from the same cultural setting). Yet, what is true knowledge now might be untrue or misleading at another point in time. This leads to the problem that the concept to which the text is linked and the text itself can have disparate meanings, even though this is often not reflected in the annotation. Therefore, the challenge arises to keep KGs up-to-date, or at least make explicit in what time period the model was considered correct~\cite{Nishioka2018}. Adding a temporal dimension to KGs can thus be seen as a contribution to the truthfulness of KGs.




The goal of this paper is to investigate how the evolution of a real world concept as it is described  in unstructured natural language text can be represented in knowledge graphs. This study is performed on the use case of the evolution of apple pie recipes extracted from historical Dutch and American newspapers. The concept of apple pie is seemingly simple. An apple pie will always contain apple, a kind of flour, a sweetener and a fat -- so what is there to evolve?

In this paper, it will be shown that even the evolution of a recipe with four main ingredients is not trivial to be represented in a knowledge graph. The problems regarding this use case include:

\textbf{NLP:} extracting structured data from unstructured historical resources that did not originate in digital form. Especially text being digitized through OCR may cause problems for NLP approaches due to errors from the OCR process. 

\textbf{Units:} extracting and understanding ingredient units presented in the texts. This involves modern units and their conversion (kilogram, pound, liter, cups), historical units (ell, zentner), but also less tangible units like ``a load of butter" or ``two deep plates of apples".

\textbf{Spatio-temporal context:} For instance, an article published in an American newspaper that describes a typical Hungarian apple pie has several spatio-temporal contexts to be taken into account when modeling the KG. 

\textbf{Evolution:} Defining and structuring the fluidity of a concept in a KG.

\textbf{Preservation:} Historical recipes which appeared in newspaper articles may cease to exist at some point if they were not digitized. However, even digitized articles cannot be preserved in their ``original" digitized form if rightsholders keep the data in closed archives.


\subsection{Impact and Contributions}

In a broader perspective, the efforts in the presented work (if further advanced) may contribute to the following ongoing areas of research:

\textbf{Food Data Research}
One use case of this work contributes to the research of food data. Historians, food scientists and economists have been studying how food was prepared and consumed in history. Research questions regard (for instance) the evolution of the usage of sugar, fat or meat. Several dimensions can be taken into account to research the usage of specific ingredients. \textbf{Health:} can the evolution of the usage of sugar and fat reveal changing concepts of what is considered healthy food and healthy cooking? \textbf{Technical development:} the frequent usage of raw meat in recipes may deliver insights about when refrigeration was available and the introduction of more fine-grained weight instructions may show that scales were used in most households. \textbf{Economics and Trade:} A sudden appearance of sugar can give insights about the availability of sugar in e.g. Western Europe. Can changing amounts of sugar in recipes also relate to the wealth of a society?

\textbf{A Fluent Understanding of the Real World}
What we take for granted today as being true might not have always been the case. This counts for apple pie recipes as much as for an understanding of more significant political topics, relevant for societies. For instance, in Belgium it has been assumed that society has always consumed as much meat as today and therefore, many people cannot accept meat consumption as a major impact in climate change. However, this assumption has been proven false\footnote{\url{http://www.veldverkenners.be/uit-de-oude-doos-vlees-een-luxeproduct}}. That means, when a concept in the real world develops and is modeled in a KG, there are more dimensions to be taken into account, such as the time, location and cultural setting in which the concept is defined. If KGs capture these `fluent' concepts along with their changes, the validity of these assumptions can be backed or refuted with underlying data. 

\textbf{Modeling Fluent Concepts with KGs}
Modeling the changes of the concept of apple pie is one small use case to discuss challenges and find preliminary solutions. However, on that foundation, lessons learned about how time-based evolution of concepts can be achieved with KGs can contribute to most KG creation processes in the future. 

Apart from the hereby described broader fields in science this work may contribute to, the concrete contributions of this work are:

\begin{itemize}
\item Preliminary analysis of Dutch and American apple pie recipes
\item Definition of KG evolution and preservation in the context of NLP
    \item Apple pie ontology
    \item Mapping of apple pie ingredients to Wikidata
    \item Timeline visualization of Dutch and American ingredients
\end{itemize}

The remainder of the paper is structured as follows. Section~\ref{sec:related-delorean} presents related work which involved knowledge graph evolution and natural language processing. Section~\ref{sec:resources-delorean} presents the main resources used in this work, followed by a detailed approach description in Section~\ref{sec:approach-delorean}. In Section~\ref{sec:evaluation-delorean} the presented approach is discussed and Section~\ref{sec:conclusions-delorean} concludes the paper.


\section{Related Work}
\label{sec:related-delorean}

 \cite{Chang2017} provide an analytical tool to visually compare, combine and investigate chocolate chip cookie recipes collected from the Web for culture analytics. While it is possible to compare recipes, the authors do not focus on the change of food concepts in time sufficiently and the data is not modeled in a Linked (Open) Data KG. 
 
 \cite{WANG2011247} study concept drift over time based on the theories of concept identity and concept morphing. The authors define the meaning of a concept in terms of intension, extension and label. Intension changes when properties are added or disregarded, extension refers to the change of instances in the ontology, and a label changes when the name of a concept changes. They furthermore propose a framework to detect concept shift and apply it to several case-studies. Finally, they describe methods to identify concept changes within the given application contexts. This work proposes interesting methodological perspectives on concept drift. However, in contrast to the presented work, it does not take into account the difficulty of concept changes related taking place in (historical) natural language text.
 
 \cite{Nishioka2018} identify the central challenge of maintaining KGs with respect to the change of entities in the real world. This work mostly focuses on the verification of changes to ensure a high data quality. In their experimentation, the authors found that evolutionary patterns in KGs are similar to social networks. The authors' results contribute to an improved KG editing process towards better efficiency and reliability. This work takes KG evolution from a different angle than the presented paper. The authors describe that errors in KGs occur due to vandalism and carelessness. However, the issue that a definition about a concept may be true at one point in time, but not in another (maybe due to cultural or economic influences as discussed in Section~\ref{sec:intro-delorean}) is not tackled. 
 
 \cite{Vossen2016} present a system that reads news articles in English, Dutch, Spanish, and Italian language, extracts event-centric information and links textual mentions to existing KGs. The authors present their system that generated episodic knowledge from unstructured data which may contribute to the Semantic Web. Again, the authors focus on events while the approach presented in this paper is understood in a broader context on the use case of food related entities. Furthermore, in contrast to this presented paper, the evolution of concepts and entities in KGs is not a main focus of \cite{Vossen2016}.

\cite{tasnim2019summarizing} Knowledge graphs are dynamic and the facts related to an entity are added or removed over time. Therefore, the multiple versions of the knowledge graph represent a snapshot of the graph at some point in time. Entities undergo evolution when new facts are added or removed. The approaches to solve the problem of automatically generating a summary out of different versions of a knowledge graph are limited. The authors propose an approach to create a summary graph capturing temporal evolution of entities across different versions of a knowledge graph in order to use the entity summary graphs for documentation generation, profiling or visualization purposes.


\section{Resources}
\label{sec:resources-delorean}


\subsection{Apple pie recipes}

In order to study the evolution of apple pie recipes over time, we collected historical data from Dutch and American newspapers. As \cite{van2018constructing} remarked, recipes from newspapers reflect tastes and viewpoints in a certain time period and can offer understanding of food cultures. This makes newspapers an invaluable data source to study evolution compared with e.g., cookbooks, which provide a static collection of recipes. For our research purposes, we analyzed the ingredients of apple pie recipes and their corresponding quantities in different contexts (i.e., time, location).

Since recipes from historical newspapers are not easily accessible, a small selection of recipes from digitized newspapers is made to provide a proof of concept and illustration of our ideas. This selection includes recipes published in one of the four Dutch newspapers \textit{Trouw}, \textit{Het Parool}, \textit{Volkskrant} and \textit{NRC Handelsblad}, or one of the three American newspapers \textit{Evening Star}, \textit{Wilmington Morning Star} and \textit{Pacific Commercial Advertiser} in the period from 1857 until 1995. Recipes were searched using the search term \textit{apple pie}, after which false positives were filtered (e.g., recipes not concerning apple pie but containing the string `apple pie'), resulting in 347 apple pie recipes. In the next step, the recipes were transformed to a structured format, including the date, location and language of the publication (i.e., context information). We then selected 12 recipes with publication dates more or less evenly spread over the time period 1857-1995 to represent the full time span. More details about the data selection is provided in Section~\ref{sec:approach-delorean}.

\subsection{Ontologies}
To model the evolution of the concept \textit{apple pie} we built an ontology that allows the inclusion of provenance information, as will be described in Section~\ref{sec:approach-delorean}. For this, two reference ontologies are used: Dublin Core (DC)\footnote{\url{http://dublincore.org/}} and the Citation Typing Ontology (CiTO).\footnote{\url{https://sparontologies.github.io/cito/current/cito.html}} In addition, we aimed to use Food Ontology\footnote{\url{https://www.bbc.co.uk/ontologies/fo}} that provides a vocabulary for recipes and ingredients. Although we were initially able to access this ontology, it was offline later. More details about building the ontology are provided in Section~\ref{sec:approach-delorean}.



\section{Proposed approach}
\label{sec:approach-delorean}


The historical newspapers described in Section~\ref{sec:resources-delorean} are digitally available both as image and as text obtained through optical character recognition~\cite{mori1999optical}. From these resources, we manually extracted the ingredients and corresponding quantities for each of the selected recipes. The ingredients were categorized according to the General Standards for Food Additives from the FAO (Food and Agriculture Organization of the United Nations).\footnote{\url{http://www.fao.org/gsfaonline/index.html}} In addition, if the free-text recipe explicitly stated the recipe originated from a different country, this information was extracted (coded following the ISO 3166 standard). For the current study, all information extraction was done manually. However, future large-scale analyses should consider an automated approach based on NLP techniques such as POS-tagging, named entity recognition, and regular expressions (see for example~\cite{van2018constructing}).

\begin{figure}
  \centering
  \includegraphics[width=0.7\textwidth]{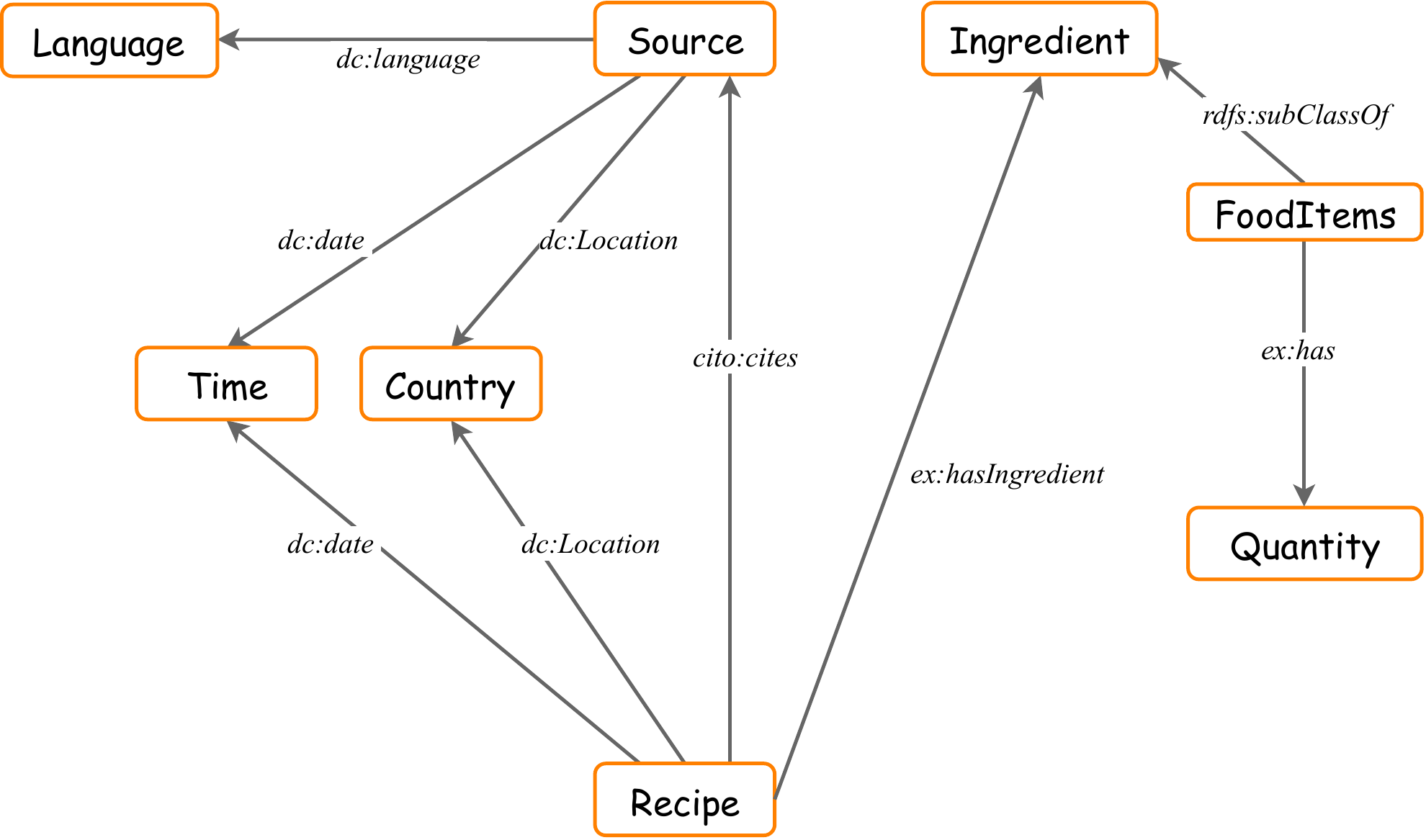}
  \caption{Data model}
  \label{fig:ontology}
\end{figure}

To model the ontology, we started with an existing ontology about food, which unfortunately went offline during our study. Existing patterns did not yield positive results, therefore we chose pre-existing general vocabularies for our model. To trace the evolution of the apple pie concept in time and space, it is essential to distinguish the spatio-temporal metadata of the recipe from the space-time metadata of its source. If the spatial-temporal metadata of the recipe are not present we can assume that they are the same as the source. Each recipe also has a superclass ingredient which contains all the classes about nutritional categories(i.e. fresh fruit or Herbs and Spices) deduced from the General Standards for Food Additives from the FAO and crossed with the Wikidata categories. In this way we can observe the variations of ingredients in space and time. Finally, each ingredient has a quantity class: considering the dataset we can detect the presence of different units of measurement that we will distinguish between imperial and decimal. This last data is implicitly linked to the language and/or reference country of the recipe. The final model is represented in Figure~\ref{fig:ontology}.

\section{Evaluation and Results: Use case/Proof of concept - Experiments}
\label{sec:evaluation-delorean}

\begin{figure}[ht]
  \centering
  \includegraphics[width=1.0\textwidth]{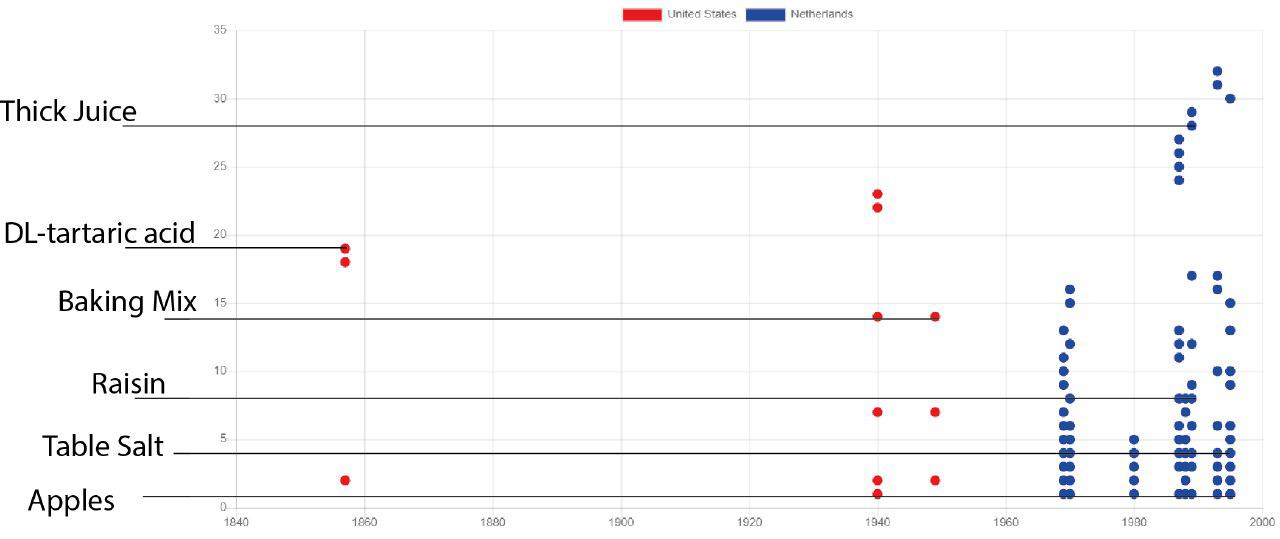}
  \caption{Timeline visualization}
  \label{fig:timeline}
\end{figure}

Apple pie as a use case for knowledge graph evolution is practical, because this small and contained concept already provides a number of challenges for ongoing research in the Semantic Web community. However, the example concept also provides a few limitations. For instance, the data revealed that instances (specific ingredients) and values (ingredient quantities) change, but on class-level (fat, sweetener, fruit), not many changes could be detected which limits the possibilities to evaluate knowledge graph evolution from all perspectives. Figure~\ref{fig:timeline} shows a preliminary visualization of the ingredients used in American and Dutch apple pie recipes. 

A full evaluation was out of the scope of this project, but in this section, we describe how several aspects of the created model could be evaluated.

When evaluating knowledge graph evolution, one important aspect is to measure how well the model reacts to changes in the data. One way to investigate this is to analyze whether all apple pie recipes in a dataset are actually covered by the knowledge graph. A suitable example for testing is a recipe extracted from an American apple pie recipe as an instant meal that only requires the `cook' to add water.\footnote{\url{https://bit.ly/2YAtYOb}} If this recipe can be covered by the model, it is an indication that the model covers the domain. 

On the other hand, it can be analyzed whether data which are not apple pie recipes are covered by the model. In an Austrian recipe dataset, the following recipe was found: ``\textit{Take 1000 kilos of bombs, a few hundred hand grenades, as many boxes of cartridges, go to Vienna with them, make a coup there and wait until you get arrested. Then the apple strudel will be ready so that one, even neatly sliced into cuts and carefully cleaned by Remdk\"orpern, will be served with the wish to `get well'}" \footnote{\url{http://anno.onb.ac.at/cgi-content/anno?aid=kik&datum=19190907&query=\%22Apfelstrudel+Rezept\%22~10&ref=anno-search&seite=7}}. This recipe is a war metaphor and not an actual recipe, and therefore it should not be possible to express this via the created model. 

A further evaluation approach may involve crowdsourcing. For instance, users may be provided with recipes and requested to judge whether or whether not the provided resource describes an apple pie recipe. If the selection by the users reflects the created knowledge graph, it is another indication for a suitable model.  

However, an essential takeaway message here is that evaluating whether or not something is to be regarded as an apple pie and should be represented in a knowledge graph depends for instance on cultural background as well as the spatio-temporal setting. Therefore, we believe that no concise ``ground truth" can be created to determine what is an apple pie and what is not, only tendencies can be given.

\section{Discussion and Conclusions}
\label{sec:conclusions-delorean}

The real world is constantly changing and knowledge that was considered true at one point in time in a specific cultural and spatial setting may not be true in another context. That means contexts evolve. On the other hand, there are knowledge graphs, which are created and maintained to continuously compose knowledge. However, often KGs are static and only reflect one snippet of reality. This static representation of the real world is a problem when attempting to understand historical descriptions of concepts (e.g., in newspapers), because linking historical concepts to today's understanding of the same concept may distort its meaning.

This paper tackled this problem on the use case of the evolution of the concept \textit{apple pie}. While the concept itself is seemingly simple, it clearly illustrates challenges concerning knowledge graph evolution in the context of NLP. The contributions of this paper include definitions of evolution and preservation of knowledge graphs in the context of NLP. A preliminary analysis of Dutch and American apple pie recipes was performed which resulted in insights about the variability of what was once understood as an apple pie. Furthermore, an ontology was developed to capture not only the recipe ingredients on their own but also specify that a resource of a recipe (e.g. a newspaper) was published in a temporal and spatial setting than the recipe itself deals with. A visualization shows the differentiating approaches to apple pie in Dutch and American recipes. 

Our take home message is that apple pie, as simple as the concept seems, can never be fully defined using a KG. Instead, the right level of broadness has to be investigated to best describe a real world understanding of apple pie as a concept as it relates to a broad context regarding time, space and culture.

\part{Decentralized technologies for Knowledge Graphs preservation and evolution}
\label{part3}
\chapter{Privacy-Protection within the Evolution and Preservation of Knowledge Graphs: 
The \vader\ approach towards Medical Citizen Science}
\label{sec:jedis}
\chapterauthor{Chang Sun, Federico Igne, Gianmarco Spinaci, Glenda Amaral, Kabul Kurniawan, Marc Gallofr\'e Oca\~na, John Domingue}

\section{Introduction} \label{sec:intro-jedi}

Knowledge graphs are dynamic in nature, new facts about an entity are added or  removed over time \cite{tasnim2019summarizing}. Certain uses of knowledge graphs require strong guarantees of data integrity, such as medical and financial data, which need to be verified by the end user automatically and reliably \cite{third2017linkchains}.  For some types of information, it is also important to ensure the privacy of users' data, which is related to the right of individuals to control how information about them is used by others. For example, the General Data Protection Regulation (GDPR) is an European legal framework that requires businesses to protect the personal data and privacy of European citizens. In citizen science - the practice of public participation and collaboration in scientific research to increase scientific knowledge - people share and contribute to data monitoring and collection programs. In health citizen science, new computational and sensing innovations, coupled with increasingly affordable access to consumer health technologies have encouraged individuals to generate personal health information to submit to a shared archive or repository, in order to allow the investigation of their own bodies, behaviors, and conditions \cite{bietz2019data}. This movement, also known as data donation, is another example of sharing sensitive and personal data.

In this paper we propose \vader\ (Volunteer Anonymous Decentralised Data to Empower Research), an architecture for the preservation, evolution and sharing of knowledge graphs containing sensitive and private information, based on decentralised technologies, such as blockchain and Solid \cite{mansour2016demonstration}.

Decentralised technologies based on blockchains are able to both address data privacy requirements and provide a trustworthy guarantee that records have maintained their integrity since publication \cite{third2017linkchains}. Considering a scenario where users have complete control over their own data, stored as knowledge graphs, blockchain technology allows for a network of knowledge graphs to be safely and securely interconnected. Through this network, a wide array of participants, such as knowledge graph owners, data providers, knowledge experts, and data scientists will be able to privately and safely share information and transact with each other.

The remainder of this paper is organized as follows. First, in Section \ref{sec:preliminaries}, we introduce the reader to the main notions on the 
context in which this work has been developed and to some related work. Then, in Section \ref{sec:architecture} we present our approach for privacy-preserving evolution and preservation of knowledge graphs using decentralized technologies.  We then finalize with some conclusions and a discussion on future work in Section \ref{sec:example-application}.

\section{Background and Related Work} \label{sec:preliminaries}

This section presents some background information and related work, thus giving the context in which this work has been developed.

\vspace{8pt}
\smallskip\noindent\textbf{Data Privacy.} Privacy has become one of the most important human rights issues of the modern age. The increasing sophistication of information technology with its capacity to collect, analyze and disseminate information on individuals introduced a sense of urgency to the demand for data privacy preservation. Recently, repeatedly reported incidents of surveillance and security breaches compromising users' privacy call into question the dominant model, in which third-parties collect and control massive amounts of personal data. Therefore, several attempts to address privacy issues have been taken, both from a legislative perspective (e.g. GDPR), as well as from a technological standpoint. In this paper, we describe a decentralized personal data management model that ensures users own and control their data.

\begin{figure}
	\centering
	\includegraphics[scale=0.20]{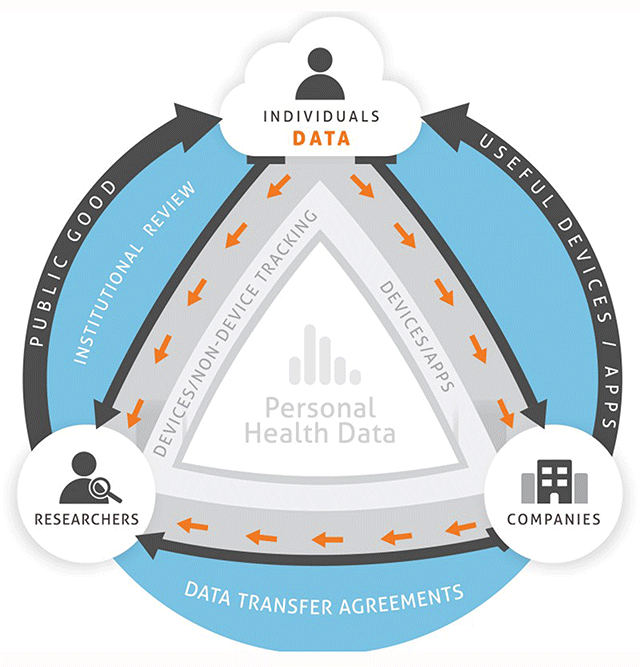} 
	\caption{ A model example on data donation in citizen science \cite{bietz2019data}}
	\label{fig:data-donation}
\end{figure}

\vspace{8pt}
\smallskip\noindent\textbf{Data Donation in Citizen Science.} Bietz et al. \cite{bietz2019data} define data donation research as ``research in which people voluntarily contribute their own personal data that was generated for a different purpose to a collective dataset''. In the context of citizen science and health research, these data may directly or indirectly contribute to an understanding of humans, and they are contributed by the individuals to whom the data refer \cite{bietz2019data}. By ``voluntarily contribute'' we mean those cases in which individuals make a clear choice to allow data about themselves to be used in a research study. In other words, participants ``opt in'' to the research. Another important aspect is that the data that get donated are often originally generated for purposes other than the research study itself. Even when the data collection is closely related to the research itself, there is often an individual benefit from the data collection that is separate from the study purposes. The development of a collective dataset can allow individuals to compare themselves to others and can yield population-level generalizations. Figure \ref{fig:data-donation} illustrates an example of  a health data exploration network, which brings together researchers in personal health data to catalyze the use of personal data for the public good \cite{bietz2019data}. In this model, individuals (data donators) share their data, such as activity levels and vital signs (heart rate, blood pressure, galvanic skin response) to be used in health research.

\vspace{8pt}
\smallskip\noindent\textbf{Knowledge Graphs.} Knowledge graphs are usually assumed to be large and arbitrary entities may be interrelated, thus covering various topical domains \cite{paulheim2017knowledge}. In his survey of knowledge graph refinement, Paulheim \cite{paulheim2017knowledge} listed the minimum set of characteristics that must be present to distinguish knowledge graphs from other knowledge collections. According to the author,  a knowledge graph  ``(i) mainly describes real world entities and their interrelations, organized in a graph, (ii) defines possible classes and relations of entities in a schema, (iii) allows for potentially interrelating arbitrary entities with each other and (iv) covers various topical domains''.

\vspace{8pt}
\smallskip\noindent\textbf{Preservation and Evolution of Knowledge Graphs.} Knowledge is not static, but constantly evolving. Not only does the knowledge itself change, but also our perception of and beliefs about it, such as its trustworthiness or accuracy. Therefore, ``if knowledge graphs are to capture at least a significant portion of the world's knowledge, they also need to be able to evolve and capture the changes made to the knowledge it contains'' \cite{bonatti2019knowledge}. For example, in the medical domain patient records contain information about the states of a patient. Something that is believed at one point might be proven false in the next time instant. This needs to be captured, tracked and reasoned with, when analysing patient data.

\vspace{8pt}
\smallskip\noindent\textbf{Blockchain.} A blockchain is essentially a distributed database of records, or public ledger of all transactions or digital events that have been executed and shared among participating parties. Each transaction in the public ledger is verified by consensus of a majority of the participants in the system. Once entered, information can never be erased. The blockchain contains a certain and verifiable record of every single transaction ever made \cite{crosby2016blockchain}.

\vspace{8pt}
\smallskip\noindent\textbf{Smart Contract.} A smart contract is a program that runs on the blockchain and has its correct execution enforced by the consensus protocol \cite{szabo1997idea}. Since they reside on the chain, they have a unique address. One can trigger a smart contract by addressing a transaction to it. It then executes independently and automatically in a prescribed manner on every node in the network, according to the data that was included in the triggering transaction. Smart contracts allow us to have general purpose computations occur on the chain \cite{christidis2016blockchains}.

\vspace{8pt}
\smallskip\noindent\textbf{Solid and Solid PODs.} Solid (Social Linked Data) is a decentralized platform, in which each users store their data in a Web-accessible personal on-line datastore (or POD). Applications run as client-side Web applications in a browser or as mobile applications. These applications use an authentication protocol to discover the user’s identity and profile data, as well as relevant links that point to the user’s POD, which contains application data. Solid supports decentralized authentication and access control, and it also supports standardized data access mechanisms \cite{mansour2016demonstration}

\section{Conceptual Architecture} \label{sec:architecture}

We are now going to give a description of the \vader\ framework.
We start with an overview of the actors involved, giving, for each of them a description of their point of view.

It is important to remark that the system aims to provide a decentralized, community-driven environment to help conducting a certain kind of research studies;
in this sense we want to avoid any central authority or any use of third party ``trusted'' services.

\begin{figure}
	\centering
	\includegraphics[scale=0.08]{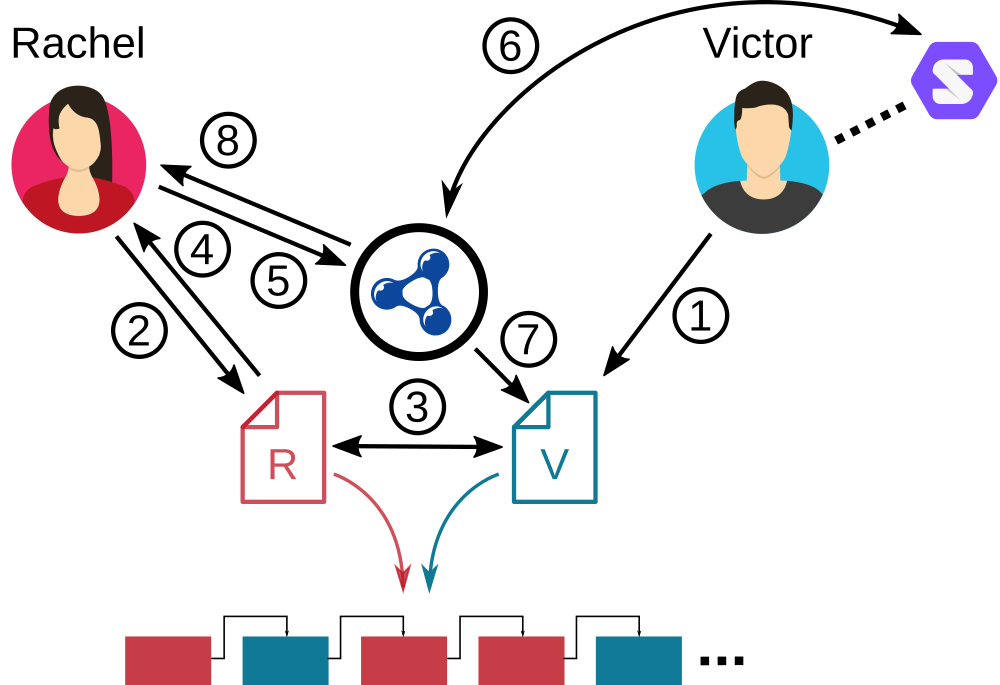} 
	\caption{Overview of the \vader\ conceptual architecture, showing a full interaction cycle between a researcher Rachel and a volunteer Victor. (1) Victor applies to be a volunteer. (2) Sometime in the future Rachel registers a research project. (3) Research and volunteers' data access constraints are checked. (4) An access point and a token identifier are generated and returned to Rachel. (5) Rachel accesses the RDF store. (6) The RDF store collects the necessary data and executes the relevant SPARQL queries. (7) The results are checked against the volunteers' data access constraints. (8) The research results are returned to Rachel.}
	\label{fig:arch}
\end{figure}

\subsection{The Actors}

We designed the system to allow two kinds of actor: researchers and volunteers.

Volunteers can provide their (personal) data through a secure private channel and validate any transaction through the underlying blockchain.
A fine-grain permission system enables the volunteer to specify which data to share, and when/how to allow the use the data.

Researchers have the ability to publish their research proposal through the blockchain, asking for data and providing a complete outline for the research.
We assume this is done providing a descriptive version of the knowledge graph (data) involved and a set of SPARQL queries used in the research.
This helps a transparent communication between the researcher and the users involved in the community.
Once enough data is collected, the system handles the data collection, query execution and results validation.
The results are then provided to the researcher, while the temporary data collected for it is destroyed.

Overall, the blockchain ecosystem we designed provides two main smart contract, each of which handles the request of the two different actors.
While the volunteers smart contract handles the join and leave request of the user and the recording of data access permissions, the researchers smart contract deals with the publication and validation of the conditions according to which the research project can be executed.

\subsection{Joining the System as a Volunteer}

User data resides \emph{with the user} in their Solid POD.
This gives the user controll over their data and provides a fine-grained privacy layer between the volunteer and the researcher.

When users want to join the ecosystem they simply invoke the related smart contract in order to be included in the set of volunteers.
Ideally the invocation contains a detailed \emph{description} of the provided data, along with any condition to fulfill to be able to access them.
The description is provided as a set of triples resembling the structure of the data in the POD.
At this stage no data is provided, and the users can revoke their permissions at any time.

\subsection{Joining the System as a Researcher}

Whenever researchers want to conduct a research study on a selection of data, they need to publish a description of the research.
This includes also the set of conditions to satisfy during the data collection process.
The researchers invoke the relevant smart contract and register their request on the blockchain.
An automatic routine can be set to collect volunteers during a certain period of time.
Eventually, the system will connect enough volunteers to take part in the research study.

The research request might also contain information about its life span. If the data collection is not possible before this period of time, the blockchain can report the failure to the user.

When successful, the researcher smart contract returns a token to the researcher along with an access point (e.g., using IPFS, a peer-to-peer network protocol for sharing digital content in a distributed file system).

The token can be used as authentication to the access point.

\subsection{Research Execution and the Temporary Knowledge Graph}

Once the right amount of data is collected the system can proceed with the data gathering and execution of the SPARQL queries over the knowledge graph.

The token generated by the smart contract contains all the details needed to execute the data gathering process and the SPARQL queries for the research.
Morover the system provides:

\begin{itemize}
    \item an RDF store to store the temporary Knowledge Graph, i.e., the result of the composition of all data collected from the volunteers. this can be virtualized, distributing the query over the set of PODs involved;
    \item a module capable of extracting data from the Solid PODs (via the provided API) and to execute the relevand SPARQL queries over the data.
    The module is also thightly bound to the underlying blockchain, anche validate the data access permissions both from the point of view of the research specification and the user data access.
\end{itemize}

Community members can agree on the specifics of the service used to carry out these tasks in a fully decentralised way.

The extraction of the triples from the PODs is registered as a transaction in the blockchain.
This allows to check if a given triple was indeed present in the study.

Once the answer is produced it can be validated again against the volunteers permissions, it is registered to the block chain and then returned to the researcher.
As a result the researcher was able to execute a set of queries over a collection of (possibly sensible) information, without having to deal with the information.

It is also worth noting that it is always possible to hide information stored in the blockchain via encryption, e.g., if the answer to some query is considered sensible information we can store the hash of its encryption.

\section{Conclusions and Future Work} \label{sec:example-application}

We initially designed the system with a specific user case in mind which addresses the problem of data collection and fair data use in the medical domain.

In this scenario users collect their personal medical data in their PODs and decide what to share to the system.

The privacy aspects highlighted before perfectly match the needs of the medical field, in which (most of the) data is sensible data and cannot be easily disclosed.
With our system, medical researchers would be able to reason over user data without the need to handle it.

In order for the system to be user-friendly and easy to use, the whole process of exchanging and updating data needs to be hidden, yet transparent, to the user.
In principle the only interactions the patients need to have with the system involve:

\begin{itemize}
    \item keeping data in the personal POD up-to-date;
    \item manage permissions according to which the data is shared with the researchers.
\end{itemize}

Both can be done through a simplified UI and in particular the open nature of the Solid allows to have seamless integration with other system.
In principle, the user will be able to connect their apps and (medical) tools to their PODs (e.g., fitness trackers, blood pressure monitor).
Finally, this approach provides \emph{preservation} and \emph{evolution} properties to the user's data.
The entire medical record of a patient can be encoded in RDF triple in its POD and use of this data can be checked, verified and filtered through the system.

In the future we plan to create incentive mechanisms for motivating volunteers  to both participate and remain active in the project. Our approach will be based on a credit system that recognizes volunteers' contributions to research projects. It is well known that reputation systems play a crucial role in building trust, promoting quality, improving collaboration and instilling loyalty \cite{dellarocas2010online}. And we aim to harness this to improve the sustainability over our framework.

With our approach researchers will have access to a wide variety of personal and sensitive data from volunteers whilst from a volunteer's perspective privacy is maintained. We believe that this provides a promising start for truly volunteer empowered trusted data donation.

\part{Methods and tools for supporting Knowledge Graphs evolution and sharing}
\label{part4}
\chapter{Tracking the Evolution of Public Datasets and Their Governance Bodies by Linking Open Data}
\label{sec:vulcanian}
\chapterauthor{Jan Portisch, Vincent Emonet, Mohamad Yaser Jaradeh,
Omaima Falattah, Bilal Koteich, Paola Espinoza-Arias, Axel Polleres}

\section{Introduction}
\label{sec:intro}

\noindent
In recent years, Linked Open Data (LOD) has been attracting many researchers in the Semantic Web community. LOD refers to the structured data available on the web which also can be processed by machines. It correlates data published in web resources by applying the Linked Data principles \cite{W3LinkedData}\cite{yu2011developer}. Further, Linked Data is also the base technology behind publicly available Open KGs such as Wikidata \cite{vrandevcic14wikidata}  and DBpedia~\cite{auer07dbpedia}, that are available as Linked Data and through SPARQL interfaces.


For instance, the \textit{European Union Data Portal} (EU ODP\footnote{\url{https://data.europa.eu/euodp/en/about}}) and the \textit{European Data Portal}\footnote{\url{https://www.europeandataportal.eu/en/homepage}} are two popular open governmental data collections i.e. large catalogs of metadata descriptions and links to public datasets. They combine data harvested from national OGD portals and other public bodies. In this work we will attempt to align these Open Data catalogs and their metadata to openly available, large knowledge graphs such as DBpedia and WikiData. 

Many works have investigated the interlinking and labeling Open Data portals and the respective datasets therein. The importance of such aligning lies in enabling users to query and search Open Data catalogs \cite{neumaier2018enabling}. However, to the best of our knowledge, there are no working applications that link the provenance metadata, namely, the public institutions who published the datasets in a systematic manner to publicly available large knowledge graphs. For instance, in DBpedia, the class ‘Organisation’ is one of the top 5 classes with 241,000 instances \cite{auer07dbpedia}, which seems to indicate that there is potential in enriching and interlinking Open Data catalogs by additional information from these knowledge graphs about the publishing organisations: we argue that mapping the description of Open Data sources with large KGs will provide an insight of the data published by organisations to capture how they evolve over time. Thus, it can help with improving the quality of Open Data in genera  and allow for further innovations in terms of data findability and scaleability.
 
In this work, specifically we aim at examining mapping techniques to align publishing organisations in existing metadata description of the \textit{European Data Portal} with organisations mentioned in the WikiData and DBpedia KGs. To this end, for instance we propose to link metadata about open data sets in catalogs to these open domain public knowledge graphs. We assume that such linkage can be achieved by applying linking methods from the entity linking domain, for instance, exploiting string similarity, or structural similarity in terms of shared attributes. One particularly important metadata element in Open Data catalogs, that can be linked, is the attribute ``publishing organisation'', but it is not but the only relevant attribute for linkage.

Our focus on linking the organisation shall allow for various analyses such as:
\begin{compactitem}
    \item \textit{Influence of Organisational Changes:}
    with linked metadata, the influence of organisational changes to open data could be analyzed, allowing us to answer questions such as: \textit{Did the frequency of updates of datasets decrease since Donald Trump was elected into office in datasets maintained by the US administration?} or \textit{Does the data of environmental offices change significantly between presidencies?}
    \item \textit{Findability of data:}
    if datasets are linked to publishing organisations, it gets easier to find other datasets by the same organisation. Linking Open Data publishing organisations to the Linked Open Data cloud~\cite{Bizer2007} will contribute to aligning a vast amount of valuable data with the FAIR (findable, accessible, interoperable, reusable) principles \cite{fair2016}. In addition, linking allows for faceted search. 
    \item \textit{Timeline analyses:}
    If open datasets are linked to organisations, aggregated timeline analyses are easier to perform. It would be, for example, possible to count the total number of published datasets by a public administration body over a certain time frame.
    \item \textit{Trust:}
    knowing for sure which organisation published which dataset and being able to find out more about the organisation through interlinked knowledge graphs will help to rely on data or datasets. When a researcher sees, for instance, that a dataset was published by \url{dbr:Federal_Statistical_Office_of_Germany}\footnote{All URL prefixes used herein, such as \texttt{dbr:} can be looked up at \url{http://prefix.cc}} and that this entitiy is a \texttt{dbc:National\_statistical\_ser\allowbreak{}vices}, it might help her to judge in how far the data is reliable if she trusts national statistical offices. If the researcher does not know the publishing body, she can use the URI to explore further associations and to learn more about an organisation.
    \end{compactitem}

\medskip\noindent
From the above challenges we derive the following research questions:
    \begin{compactitem}
    \item \textbf{RQ1:} \textit{How can publishing organisations of open data knowledge sources be linked to existing knowledge graphs? In how far are the open data publishing organisations represented in knowledge graphs?} 
    \item \textbf{RQ2:} \textit{How should mappings look like in order to account for knowledge graph evolution as well as evolution of the dataset (including the corresponding metadata)?}
    \item \textbf{RQ3:} \textit{How can organisational evolution be represented in a knowledge graph? Or, in other words: are resepctive attributes to model organisational change represented and used in existing KGs?}
\end{compactitem}


\section{Knowledge Graphs Evolution and Preservation}
\label{sec:def-vulcanian}

Knowledge Graphs have evolved as flexible and powerful means of representing general world knowledge, and have been in the focus of research since 2012 resulting in a wide variety of published descriptions and definitions. According to Paulheim \cite{Paulheim15}, a Knowledge Graph mainly describes real world entities and their interrelations, defines possible classes and relations of entities in a schema, allows for potentially interrelating arbitrary entities with each other and covers various topical domains. 

Knowledge Graphs (KGs) represent relevant aspects of a domain of interest or general knowledge of common interests as in the case of WikiData and DBpedia. Both Knowledge Graphs describe relevant entities and their important relationships. It is important to point that these aspects and thereby the KGs themselves and their underlying data (metadata) are evolving over time. Thus, reflecting the changes in Knowledge Graphs usually requires constant update of the graph or even replacing it entirely. In doing so, metadata describing the Knowledge Graph is changing and evolving as well. Open Governmental Data (OGD) is a domain where this issue can be touched. OGD support structured data of common interest to citizens, usually with some specific data (e.g. detailed statistical or census data), not being modelled in that degree of detail in existing Open KGs. Therefore, we propose a method that interlink OGD datasets to large KGs to observe their evolution over time. From our prospective, in order to interlink structured data from Open Data catalogs with KGs we need to keep track of three aspects of the KGs evolution: (i) the evolution of the underlying datasets we want to link, (ii) the metadata describing those datasets, and (iii) the history of this evolution.

\section{Related Work}
\label{sec:related-vulcanian}

\noindent

The \textit{Open Data Portal Watch (ODPW)}\footnote{\url{https://data.wu.ac.at/}} \cite{neumaier2016li} project is an attempt to collect weekly snapshots of several monitored data portals. These snapshots are mapped and published by means of homogenized metadata from standard vocabularies such as DCAT\footnote{\url{https://www.w3.org/TR/vocab-dcat/}} and Schema.org. 
However, these metadata are not linked to other Knowledge Graphs in order to provide information about publisher organisations and to be able to track the evolution of the mentioned organisations. 
The \textit{Google Dataset Search} is a dataset discovery tool that provides search capabilities over potentially all datasets published on the Web \cite {noy2019google}. In order to make it possible, this tool use  metadata information provided by publishers following Schema.org or DCAT vocabularies. Then, these metadata is reconciled  with the items of the Google Knowledge Graph. Despite this tool hosts several datasets published on the Web, the provenance of metadata is usually missing and the publisher metadata is not linked to other public Knowledge Graphs.
Recently, a new framework\footnote{\url{https://data.wu.ac.at/odgraphsearch}} has been implemented in order to allow semantic search between Open Data portals \cite{neumaier2018}. It focuses on spatio-temporal as the most relevant metadata that data portals should include as pointed in \cite{kacprzak2019characterising}. However this framework does not address how to interlink the datasets of organisations from Open Data to public Knowledge Graphs and how to track the evolution of the involved datasets. It is worth mentioning that this work is useful in order to understand the entity linking techniques applied in this approach. 
The \textit{ADEQUATE} \cite{neumaier2016automated}\cite{neumaier2018search} is a data-and community-driven quality improvements project aimed to improve the quality of metadata in Open Data. The main features of this system include: (i) assessing the quality of datasets and its metadata with quality metrics, (ii) applying different heuristic algorithms to eliminate the quality issues, and (iii) implementing different Semantic Web Methods to Open Data to transform CSV formats into a Linked Data. This work has resulted in interlinking many of the existed Open Data publishing systems. However, this project has not included a mapping solution between the polishers metadata and other open resources.
In relation to the Knowledge Graph evolution, there are some techniques that have been proposed earlier \cite{umbrich2010dataset} e.g. to detect changes during their evolution, to use vocabularies in order to represent the change information, or to propagate changes to replicas. Recently, a framework for detecting and analysing changes and the evolution history of LOD datasets has been proposed \cite{tasnim2019summarizing}. This framework consist of generating a summary of the evolution of entities as molecules containing a compact representation of the objects and properties including its temporal validity. Our work rather aims at exploiting the metadata used to annotate the datasets as a way to handle the representation of the organisational evolution.

\section{Resources}
\label{sec:resources-vulcanian}
\noindent
For our experiments, we used the metadata provided by the \textit{Open Data Portal Watch} (ODPW) project. In particular, for a preliminary analysis, we focused on two portals listed on ODPW, namely: (i) The \textit{European Data Portal} due to the good quality of the metadata available there, and, as a counter-example, (ii) the \textit{Hawaiian Data Portal} as the provenance metadata is of less sufficient quality there. This decision was taken in order to judge the quality of the linker in different realistic situations. The VulcanLinker itself uses pre-crawled label data from DBpedia and Wikidata in order to maximize performance.

\section{Proposed Approach}
\label{sec:approach-vulcanian}
The approach that we are considering addresses linking certain properties of the metadata describing a dataset.

\subsection{Linking}
This section addresses \textit{research question 1}. Most open datasets reside on individual Web sites and are available in various formats such as CSV, Excel, or RDF, for example. Data portals such as the \textit{European Data Portal} or \textit{Open Data Portal Watch} compile various datasets and provide metadata about the dataset such as the creation date or the publisher. However, the metadata is not formalized and exists only in a pure textual representation as string.
In order to perform the task at hand, we present a system called \textit{``VulcanLinker''}. VulcanLinker works by matching certain values to its corresponding entities in linked Knowledge Graphs (e.g. Wikidata and Dbpedia). Our proposed system queries some of the available Knowledge Graphs and computes the coverage of linked entities from those sources. To do so, the system computes the closure of all subclasses $C$ of the entity in question (e.g. \texttt{dbr:Organization}), and then performs a similarity measure $sim$ on the labels of the instances ($i$) of the fetched set of subclasses. Equation (\ref{eq:similarity}) shows the formalisation of the method.

The approach is a straight forward process of fetching resources from the Knowledge Graphs and then matching them to the available metadata from our Open Data dataset descriptions. 

\begin{equation}
    \varrho = sim(e,Att_e,d,Att_d)
    \label{eq:similarity}
\end{equation}
where $sim(.)$ is a similarity measure for matching a KG entity $e$ with a dataset $d$, parameterised by comparing the associated KG attributes in the set $Att_e=\{att_1, ... att_n\}$ with a corresponding set of metadata attributes $Att_d$ in the dataset description $d$. In the simplest case, in our current prototype implementation we just compare \texttt{rdfs:label}s with the \texttt{dcat:publisher} field of the dataset description.

The skeleton of the SPARQL query that can fetch the instances of a particular entity type (in our case, organisations) along with attributes relevant for matching from the KGs is shown in Listing \ref{lst:fetch_candidates}. However, in practice we tried to run the query on online instances of Wikidata and DBpedia, and they failed due to the fact that this kind of query requires the materialisation of all instances belonging to the type class of the entity (to compare) and all it's subsequent classes.


\lstset{frame=tb,
  language=SPARQL,
  aboveskip=3mm,
  belowskip=3mm,
  showstringspaces=false,
  columns=flexible,
  basicstyle={\small\ttfamily},
  numbers=none,
  numberstyle=\tiny\color{gray},
  keywordstyle=\color{blue},
  commentstyle=\color{dkgreen},
  stringstyle=\color{mauve},
  breaklines=true,
  breakatwhitespace=true,
  tabsize=3
}
\begin{lstlisting}[caption={Skeleton SPARQL query to fetch data for matching}, label={lst:fetch_candidates} ]
SELECT ?entity
WHERE {
    ?entity <isA>/<subClassOf>* <entityType>;
            <att1>              ?att1;
            ... 
            <attN>              ?attN ; .
}
\end{lstlisting}


To circumvent this, we used look-up services for both example use-cases Knowledge Graphs (e.g. Wikidata and DBpedia) which fetches matching entities from an index based on the label (\texttt{rdfs:label}) of those entities.
Afterwards, the next step is see which entity from the fetched set of results matches the surface from of the string we VulcanLinker is linking. At the moment, we use simple exact string matching.

Last but not least, VulcanLinker computes the coverage $\Gamma$ of a certain KG on a property using the formula (\ref{eq:coverage}). We note that this does not yet evaluate the accuracy of the matching, as we simply test the share of successful matches (independent of false positives and negatives). A more in-depth evaluation in terms of computing other metrics on the dataset matching (e.g. precision, recall) is on our agenda, but would involve manual construction of a ground truth sample.

\begin{equation}
    \Gamma = \frac{\# matched}{\# overall} \%
    \label{eq:coverage}
\end{equation}


\noindent

\subsection{Linking Formats}
This section addresses \textit{research question 2}. In order to discuss mappings, the concepts \textit{correspondence} and \textit{alignment} are defined in the following: A correspondence is defined here as a semantic link between two elements from two datasets. The set of correspondences of two datasets is also known as alignment. This follows the definitions established in the ontology matching community \cite{euzenat_ontology_2013}.
One challenge that is to be addressed is the \textit{moving target problem}: Over time, datasets and Knowledge Graphs are updated and changed. A publishing institution, for instance, might be renamed and accordingly be removed from a Knowledge Graph. Correspondences might end up linking to non-existing elements. In order to address these, we suggest to provide at least the following metadata for each alignment:

\begin{itemize}
    \item \texttt{CreationDate}: The date the alignment was created. This ensures the validity of the alignment as of a certain date.
    \item \texttt{CreationMethod}: The creation method defines which process mapped the metadata. It might be an automated method but a \textit{manual process} is also a valid option.
\end{itemize}
Furthermore, we suggest to add the following metadata for each correspondence:
\begin{itemize}
    \item \texttt{Confidence}: The confidence of the human annotator of the automatic linker.
    \item \texttt{Explanation}: An explanatory, human understandable, explanation for the individual correspondence by the linker.
\end{itemize}
The metadata given for each correspondence/alignment shall be extensible so that additional metadata can be added. In addition, we propose that the metadata can be queried using a SPARQL \cite{standard/w3c/sparql} endpoint. The latter could, for instance, be provided by \textit{Open Data Portal Watch} or any other central data portal. 
Moreover, the evolution of metadata needs to be queryable. This means that an alignment as of a certain year or date shall be retrievable. Note that one particular snapshot of a dataset might have multiple alignment versions depending on the method applied and the validity (\texttt{CreationDate}). 

Alignments shall be represented in RDF in order to be queryable. Listing \ref{lst_alignment_format} shows an example of a possible RDF alingment representation inspired by the Alignment API format \cite{alignmentapi2011} that is used by the ontology matching community. In the listing \texttt{a\_uri} refers to the URI of the alignment. \texttt{c\_uri} refers to the unique URI of an individual correspondence. 

\begin{lstlisting}[caption={Exemplary structure of an alignment format.}, label={lst_alignment_format} ]
<c_uri> rdf:subject <dataSetURI> <a_uri> .
<c_uri> rdf:object <mappedOrganisation> <a_uri> .
<c_uri> rdf:predicate <publisher> <a_uri> .
<c_uri> <confidence> "0.8" <a_uri> .
<c_uri> <explanation> "explanation" <a_uri> .
<a_uri> <dc:createdOn> "13-11-1991" <rootGraph> .
<a_uri> <dc:creator> <orcid> <rootGraph> .
<a_uri> <methods> <similarity/distance method> <rootGraph> .
\end{lstlisting}

\subsection{Do Current Open Knowledge Graphs Cover Organisational Evolution?}
\label{ssec:assessment}

This section addresses \textit{research question 3}. We assess whether 
\emph{models} used by current Open Knowledge Graphs, enable 
to capture the evolution and context of entities (in our case, organisational changes affecting data publishers, date of those changes, etc). Secondly, we want to assess to which extent those models are actually used, e.g., whether and to which degree the \emph{data} in the Knowledge Graphs actually captures evolution of organisations?
To this end, two dimensions are suggested to be explored: 
\begin{enumerate}
    \item Does the ontology (T-Box) allow to define metadata related to the evolution of an organisation? 
    \item Does the organisational data (A-Box) fully describe the evolution of organisations (T-Box usage)?
\end{enumerate}

Such assessment can be performed using SPARQL queries on the KG directly counting the usage of respective properties, combined with a qualitative assessment by manual inspection of sample relevant organisations.
For a prototypical evaluation along these lines on Wikidata, we refer to Section~\ref{sec:onto_eval} below.

\section{Evaluation and Results: Use case/Proof of concept - Experiments}
\label{sec:evaluation-vulcanian}
We open the horizon of evaluating the system, in term of what has been implemented and discuss possibilities and approaches for subsequent work.
\subsection{Interlinking evaluation}

\noindent

For running proof of concept experiments, the experimental setup comprised a Linux virtual machine running Ubuntu 18.04 LTS, with 48G of memory, and 8 CPUs 2.7GHz. We extracted meta-data from the two portals (\url{open-data.europa.eu}, \url{data.hawaii.gov}), as mentioned above, to attempt automated linkage. Our prototypival organisation linkger, VulcanLinker, is written in Python 3.6, and at the moment only implements simple exact string matching on organisation labels as mentioned above, but we envision this matching to be extensible and to be improved in the future. All code that was written in the scope of this prototype has been made publicly available on GitHub.\footnote{\url{https://github.com/isws-2019-vulcanian/Prototype}}

We ran a baseline experiment on DBpedia and Wikidata to compute overall coverage for the datasets on these portals. Here, coverage is defined as the number of links that were found (independently of whether those links are correct).

\begin{table}
\centering
\caption{Coverage percentage of VulcanLinker system on two different data portals, two Knowledge Graphs, matching the ``organisation'' property of the datasets}
\begin{tabular}{l|ll}
                            & \textbf{DBpedia} & \textbf{Wikidata} \\ \hline
\textbf{Open Europa portal} & 64.1\%           & 88.3\%            \\
\textbf{Hawaii data portal} & 6.32\%          & 55\%             
\end{tabular}
\label{tab:baseline}
\end{table}

Table (\ref{tab:baseline}) shows that Wikidata has a better coverage than DBpedia; also,  the quality of the dataset itself plays a role in the scoring, as shown by the Hawaiian portal dataset.

As shown in table \ref{tab:baseline}, the coverage differs significantly between the linking targets. In our experiment, Wikidata performed better, in terms of the number of available relevant organisations in it. Likely reasons are more overall concepts or better label annotations. 

In addition, the coverage differs between the two data portals. This is due to the varying quality of the provenance metadata: 
The quality of the alignments highly depends on the metadata that is provided. In many cases, the metadata given is insufficient, on the \textit{Hawaiian Data Portal}, for instance, organisations are often given in the form of first names such as \textit{Kevin, Kayla}, or \textit{Kaisa}. Such entries are hard to link. 
A good matcher has to be capable of ignoring such entries. Another approach in linking such harder cases is to include more metadata information in the matching process. 
From the data set URL, for instance, an organisation might be inferred in order to improve the similarity calculation. This is a future research direction that can be explored.

\subsection{Wikidata and DBpedia Ontologies and Coverage Assessment}
\label{sec:onto_eval}
In order to evaluate RQ3, we evaluated and discussed the dimensions presented in Section \ref{ssec:assessment} on both Wikidata and DBpedia, in terms of analysing the availability of suitable terms in the KGs' shemata and their usage to describe organisational evolution. 

\subsubsection{Wikidata assessment}

On Wikidata, various properties and classes are available to define the evolution of an organisation:

\begin{compactitem}
\small
    \item \texttt{inception} (\texttt{wdt:P571}\footnote{\url{https://www.wikidata.org/wiki/Property:P571}})
    \item \texttt{dissolved, abolished or demolished} (\texttt{wdt:P576})
    \item \texttt{parent\_organization} (\texttt{wdt:P749})
    \item \texttt{replaces} (\texttt{wdt:P1365}\footnote{The Wikidata ontology also proposes an inverse property for \texttt{replaces} which is \texttt{replacedBy} (\texttt{wdt:P1366})})
    \item \texttt{merger} (\url{https://www.wikidata.org/wiki/Q452440})
\end{compactitem}

The data model defined by Wikidata offers an efficient way to represent the evolution of an organisation by providing references, such as the point in time property (\texttt{wdt:P585}) which provides the date from when a fact started to be true.

As for the usage, one relevant organisation is described in the following: the \textit{Scottish National Party} (\url{https://www.wikidata.org/wiki/Q10658}). This organisation replaces the \textit{National Party of Scotland} (\url{https://www.wikidata.org/wiki/Q6974819}), created in 1928 and dissolved in 1934, as well as the \textit{Scottish Party} (\url{https://www.wikidata.org/wiki/Q7437896}), created in 1930 and dissolved in 1934. 
In this case, the \texttt{replaces} property does not hold any references providing the date (which 1934) at which the statement started to be true. This information can theoretically be inferred by the date of dissolution of the the previous party. 
When looking closer, one can infer that there was merge of the \textit{National Party of Scotland} and the \textit{Scottish Party}. However, such an event should have been recorded by the \texttt{merger} property. So, the underlying ontology is not fully exploited on Wikidata. Multiple such instances can be found.

In order to give an indication of the usage, the \texttt{replaced} property has been closer examined: The SPARQL query in listing \ref{lst:countother}, counts the use of the \texttt{replaces} property for subclasses of organisation.

\lstset{frame=tb,
  language=SPARQL,
  aboveskip=3mm,
  belowskip=3mm,
  showstringspaces=false,
  columns=flexible,
  basicstyle={\small\ttfamily},
  numbers=none,
  numberstyle=\tiny\color{gray},
  keywordstyle=\color{blue},
  commentstyle=\color{dkgreen},
  stringstyle=\color{mauve},
  breaklines=true,
  breakatwhitespace=true,
  tabsize=3
}
\begin{lstlisting}[label={lst:countother},caption=Count the use of the \texttt{replaces} property for subclasses of organisation]
SELECT (COUNT(*) AS ?C)
WHERE {
    ?item wdt:P1365 ?formerItem .
    ?item wdt:P31/wdt:P279* wd:Q43229.
}
\end{lstlisting}

The results returned for the \texttt{replaces} property is a count of 15,225 statements. The same query has been used for the \texttt{replacedBy} property, returning 19,102 statements. 
Those results can be compared to the number of organisations described in Wikidata computed using the SPARQL query presented in listing \ref{lst:countReplaces}.

\lstset{frame=tb,
  language=SPARQL,
  aboveskip=3mm,
  belowskip=3mm,
  showstringspaces=false,
  columns=flexible,
  basicstyle={\small\ttfamily},
  numbers=none,
  numberstyle=\tiny\color{gray},
  keywordstyle=\color{blue},
  commentstyle=\color{dkgreen},
  stringstyle=\color{mauve},
  breaklines=true,
  breakatwhitespace=true,
  tabsize=3
}
\begin{lstlisting}[label={lst:countReplaces},caption=Count the number of organisation subclasses]
SELECT (COUNT(*) AS ?C)
WHERE {
    ?item wdt:P31/wdt:P279* wd:Q43229.
}
\end{lstlisting}

The result is 1,786,652 organisations. It seems then that information about evolution concern no more than around 1.93\% (1,786,652/(19,102+15,225)) of organisations.

\subsubsection{DBpedia assessment}

DBpedia proposes a class to describe organisations: dbo:Organisation\footnote{\url{http://dbpedia.org/ontology/Organisation}}, which defines such properties relevant to describe the organisation evolution (i.e. \texttt{dbo:p\newline arentOrganisation}, \texttt{dbo:formationDate}, \texttt{dbo:extinctionDate})

As for the usage, no properties have been found to define the evolution of an organisation besides  \texttt{dbp:merger} \footnote{\url{http://dbpedia.org/property/merger}} and \texttt{dbo:mergerDate} \footnote{\url{http://dbpedia.org/ontology/mergerDate}}, the former of which we found to be used, for instance in the example organisation \texttt{http://dbpedia.org/p\newline age/Scottish\_National\_Party} (who could also be an Open Data provider), yet the \texttt{dbo:mergerDate} being missing. Also, note that the \texttt{dbo:mergerDate} property's \texttt{rdfs:domain} is \texttt{dbo:Place}: in fact, though \texttt{dbo:Place} is defined as disjoint with \texttt{dbo:Agent}, which is the parent class of \texttt{dbo:Organisation}, so its usage to define a date for the merge of multiple organisations would make the instance inconsistent. We infer that the consistency constraints of the DBpedia ontology make the joint usage of these properties invalid.

As for quantifying the usage of the said properties, we could deploy similar count queries as for Wikidata, but expect similarly low coverage.

\section{Discussion and Conclusions}
\label{sec:conclusions-vulcanian}
In this chapter, we suggested to link structured, open datasets from Open Government Data Portals to knowledge graphs (KGs). A particularly interesting linking dimension we focused on was the publishing organisations, under the assumption that public institutions are well covered and represented in open KGs.  A good starting point is to link the metadata of a dataset - in particular publishing organisation data. We argue that a holistic approach to address this problem should address both organisation linking and analysing organisational change. In this preliminary work we prototypically address a baseline linking approach which can be extended to cover different aspects of evolution.

The preliminary experiments we conducted included various interesting analyses which we hope to extend in the future to more  detailed results e.g. about the evolution of these organisations, their governance and the data they publish.

Our approach considers the evolution of data sets and the evolution of a knowledge graph. Firstly, we implemented a prototypical organisation linker, the VulcanLinker, which allows to match plain strings (such as organisation strings) to Dbpedia and Wikidata KGs. We ran this simple matcher on publishing organisations given by two data portals. Our experiment revealed that the provenance data quality differs among different portals and that the linking outcome is highly dependent on the latter quality. Our experiment further hinted at Wikidata likely being a better target for matching organisations.
Secondly, we proposed a method to version metadata in a way so that point and delta queries are possible.
In addition, we suggest that alignment data itself should be versioned, and contain a minimal set of metadata information, e.g., allowing for multiple matching mechanisms to annotated. Furthermore, we should consider the fact that metadata changes over time.
Lastly, we analysed the schema of the existing KGs in terms of coverage of properties to model organisational change and evolution.

Future work should focus on versioning approaches of metadata - in particular considering memory consumption and query performance. In addition, the presented \textit{VulcanLinker} returns reasonable results but focuses only on simple matching of labels. However, there are additional attribues than could be taken into consideration (regarding attributes mentioned in equation (\ref{eq:similarity})) such as URLs of organisations, country of origin (for disambiguation of same named institutions in different countries), publisher \texttt{dcat:contactPoint} info, and much more. Furthermore, the evaluation of different version of the KGs can be tested to discover how the organisational information has evolved (changed) overtime.

Another research direction would be to calculate the coverage over different knowledge graph versions to show how the knowledge graphs evolved and changed over time.

\chapter{Designing a Platform for Versioned Knowledge Graph Data Access and Provision on the Web}
\label{sec:slytherin}
\chapterauthor{Viet Bach Nguyen, Michalis Georgiou, Ivan Heibi, Mengya Liu, Maheshkumar Mistry, Anna Nguyen, Michel Dumontier}

\section{Research Questions}
\label{sec:rq-slytherin}
The FAIR Principles ask that unique and persistent identifiers are specified for all digital resources. What we want to achieve is to answer the question: How to identify and retrieve resources from knowledge graphs that are continuously changing? For example: How can we provide users with persistent identifiers to exactly identify and retrieve the same data they used when they made the query on a knowledge graph? The evolving graphs can offer a great opportunity to build and validate the accuracy of predictive algorithms using temporal subsets of data. For example, a predictive data model can be generated from known drug indications between 2000-2018 and validated with data after this period. 


\section{Introduction}
\label{sec:intro-slytherin}
The increasing requirement of data in many fields such as data analytics, data mining and machine learning makes us question if the data we use has the Fair Principles. In 2016, Wilkinson et al., comment on the first formal publication of the FAIR Principles \cite{fair1}. There are many reasons why we should have to use FAIR Principles. One of them is that we can make it easier for us to use our data for a new purpose, or make it easier for other people to find, use or cite our data. This can also ensure that the data are available in the future for future work on it. In addition, it can satisfy the expectation around the data management from different institutions, funding agencies or journals. To tackle these problems, we intend to build an API service which uses the Trusty URI specification\footnote{http://trustyuri.net/}. Giving the users the opportunity to use this service, it can ensure that the data being used meet the requirements of the FAIR Principles.

\section{Use cases}
For our use cases (UC), imagine being a researcher who wants to work on a particular knowledge graph of 2015 which follows the FAIR Principles:
\begin{itemize}
    \item \textbf{UC1 Knowledge Graph Search:}  The user may search for a knowledge graph and then it will provide a Trusty URI for the knowledge graph and a second Trusty URI for the metadata;
    \item \textbf{UC2 Knowledge Graph Retrieval:} The user can retrieve the knowledge graph and their metadata using a Trusty URI;
    \item \textbf{UC3 Queries on specific versions of KGs that evolve over time:} The user can define a query that will be executed and return a dataset result with a Trusty URI that can be published and used later in the future.
\end{itemize}

The purpose of making a service that meets all these use case requirements is to provide a way for reproducing the same datasets in the future, therefore conserving all provenance details, and can be used for further analysis. One of the main added values is this will allow users to publish these datasets in a persistent and reliable way. For example, in data analysis, after the user has successfully retrieved his dataset by making a query to our service, then he might want to run a prediction algorithm on the data you have from 2015 to 2018 and compare it to the data of 2019. If the results are correct, then he might want to use his prediction algorithm to predict the data of 2020. In this way, he will have another opportunity to split test and training dataset to improve his algorithm.

\section{Related Work}
\label{sec:related-slytherin}
In this section, we focus on finding out resources and related work that can help us in the process of designing the system.

\subsection{FAIR Principles}

In order to ensure transparency reproducibility and reusability, the authors of \cite{fair1} provide us with the FAIR Principles. The FAIR Principles consist of findability, accessibility, interoperability, and reusability of the digital data.

To be \textit{Findable} means that (meta)data is assigned to a globally unique and persistent identifier,  that it is described with rich metadata, metadata clearly and explicitly includes the identifier of the data it describes, and (meta)data are registered or indexed in a searchable resource.

To be \textit{Accessible}, the (meta)data are retrievable by their identifier using a standardized communications protocol, the protocol is open, free, and universally implementable, the protocol allows for an authentication and authorization procedure, where necessary. The metadata is accessible, even when the data are no longer available.

To be \textit{Interoperable}, the (meta)data use a formal, accessible, shared, and broadly applicable language for knowledge representation, (meta)data use vocabularies that follow FAIR principles and (meta)data include qualified references to other (meta)data.

To be \textit{Reusable}, the meta(data) are richly described with a plurality of accurate and relevant attributes such as, (meta)data are released with a clear and accessible data usage license, (meta)data are associated with detailed provenance, (meta)data meet domain-relevant community standards.

\subsection{Trusty URI Specification}
Trusty URIs is a framework that provides Verifiable, Immutable, and Permanent Digital Artifacts for Linked Data. The Trusty URI intents to make URIs verifiable, immutable and permanent. This is a modular approach, where dissimilar modules handle dissimilar kinds of content on dissimilar conceptual levels of abstraction, from byte-level to high-level formalisms. The Trusty URI includes a cryptographic hash value, that can be used to verify the particular source. Next is an example of a Trusty URI:
\begin{example}
http://example.org/r1.RAcbjcRIQozo2wBMq4WcCYkFAjRz0AX\newline-Ux3PquZZrC68s
\end{example}

As we can see, every character that comes after the \textbf{r1.} is the artifact code of the Trusty URI. In this example, we can see that the first 2 characters which are \textbf{RA} define the type and the version of the module. The hash value is represented in the remaining 43 characters \cite{trusty_uri}. 

\subsection{Nanopublication}
Nanopublication's main involvement is to share scientific data in a computer-interpretable way in a formal of semantic notation such as RDF\footnote{\url{http://nanopub.org/wordpress/}, last accessed 2019-07-05}. Nanopublications have the ability to cite other nanopublications through their URIs, by creating their own version of complex citation networks \cite{trusty_uri}. Nanopublications are immutable which means that once they are shared with their version of the resource, no one can modify them. They use Trusty URIs with cryptographic hash values designed on the RDF content. Thus, if you use a nanopublication with an identifier (A1) it will cite you to other nanopublication with the identifier (A2). With this, you retrieve nanopublications in a verifiable way. Additionally, with the way nanopublications use their Trusty URIs they can be self-referenced which means that they contain their provenance information and meta-data.  In the end, following this nanopublications are verifiable, immutable and permanent. 

\subsection{Bio2vec}
To enhance the development of data analytics and machine learning methods in the fields of biology and biomedicine the Bio2Vec platform\footnote{\url{https://bio2vec.cbrc.kaust.edu.sa/}, last accessed 2019-07-05} was created. The aim was to discover molecular mechanisms underlying complex disease and drugs’ mode of action. The platform covers embeddings from text and knowledge graphs like GO terms, proteins, drugs, and diseases. It also offers FAIR data, helping the users by providing them already prepared data.

\subsection{Ostrich}
Ostrich is an RDF triple store that allows multiple versions of a dataset to be stored and queried at the same time \cite{ostrich}. Ostrich combines three RDF storage strategies:
\begin{enumerate}
    \item The Independent Copies (IC) approach creates separate instantiations of datasets for each change or set of changes;
    \item The Change-Based (CB) approach instead only stores change-sets between versions;
    \item The Timestamp-Based (TB) approach stores the temporal validity of facts.
\end{enumerate}

Ostrich will store fully materialized snapshots followed by delta chains. The materialization of a knowledge graph is the process of building and storing all the entire KG. A delta chain is a set of consecutive changes that have been made into the KG. Each delta chain is stored in six tree-based indexes, where values are dictionary-encoded and timestamped to reduce storage requirements and lookup times. These six indexes correspond to the combinations for storing three triple component orders separately for additions and deletions. Having these indexes ensures a quicker resolving time for the requested queries. 

Each delta chain will start with a fully materialized snapshot. A delta chain is represented through two dictionaries, one for the snapshot and one for the deltas. The snapshot dictionary consists of triple components that already existed in the snapshot. All other triple components are stored in the delta dictionary. This dictionary is shared between the additions and deletions, as the dictionary ignores whether or not the triple is an addition or deletion. The snapshot dictionary can be optimized and sorted, as it will not change over time. The delta dictionary is volatile, as each new version can introduce new mappings.

To sum up, Ostrich is a novel trade-off approach in terms of ingestion time, storage size and lookup times. It fulfills the requirements for a backend RDF archive storage solution for supporting versioning queries in the TPF framework.

\begin{figure}
    \centering
    \includegraphics[scale=0.4]{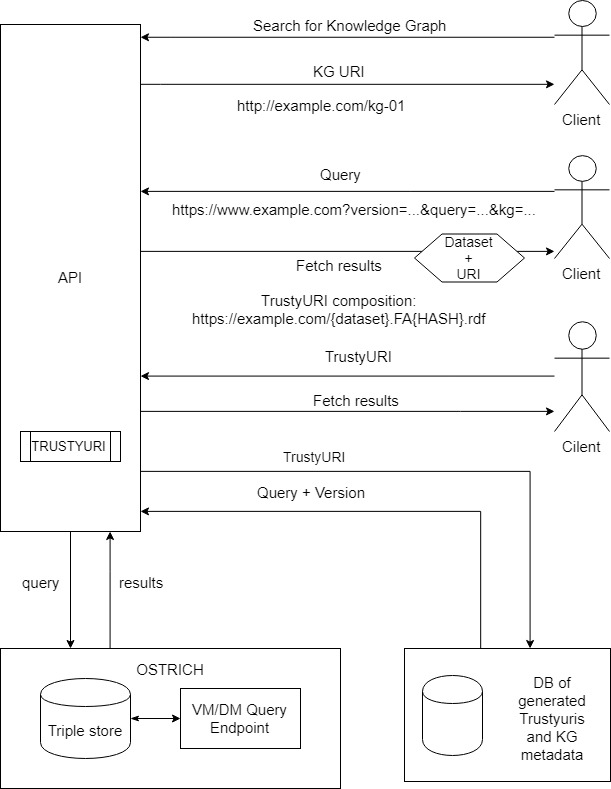}
    \caption{The general architecture of our proposed approach. Each different client on the right side of the diagram represents a different use case, which will send/call the API so it will handle its request. The API is a connection point between the different technologies used }
    \label{fig:architecture}
\end{figure}




\section{Proposed approach}
\label{sec:approach-slytherin}
Storing and querying RDF datasets that evolve and change over time has been addressed in multiple ways. Our effort is devoted to proposing a system design which combines existing partial solutions to address the problem at hand. The mission is to search and analyze potential technologies, tools, and frameworks to come up with a systematic solution to directly solve the problem. The goal is to design an RDF data provision service software that supports versioned querying. We aim to offer our solution to KG providers who want to meet the mentioned requirements. In our system the query will be sent to Ostrich, which will: 
\begin{enumerate}
    \item Detect the right version of the KG (the dataset snapshot).
    \item Apply the delta chain of changes to such version and materialize it.
    \item Apply the query on the materialized KG.
    \item Return the results of the query.
\end{enumerate}

\section{Preliminary Results}
We propose a web application design that incorporates the Ostrich triple store and Trusty URI and takes care of providing a way to efficiently store, access and query different versions of RDF data and a way to persistently identify the exposed resources, respectively.

\medskip
\noindent\textbf{System Overview}. Our system design consists of several components: a web user interface, an API layer, a database of Trusty URIs and metadata, and a triple store. The API layer can be called from any web client and to provide easy access to the functionalities, users will be able to call the APIs through a friendly web user interface. The API is represented as a connection point between the database and the triplestore. 
This overview description of the system can be seen in the following diagram. Behind the scenes, we also provide a comprehensive overview of all activities and interactions inside our system in the following sequence diagram.

\begin{figure}
    \centering
    \includegraphics[scale=0.4]{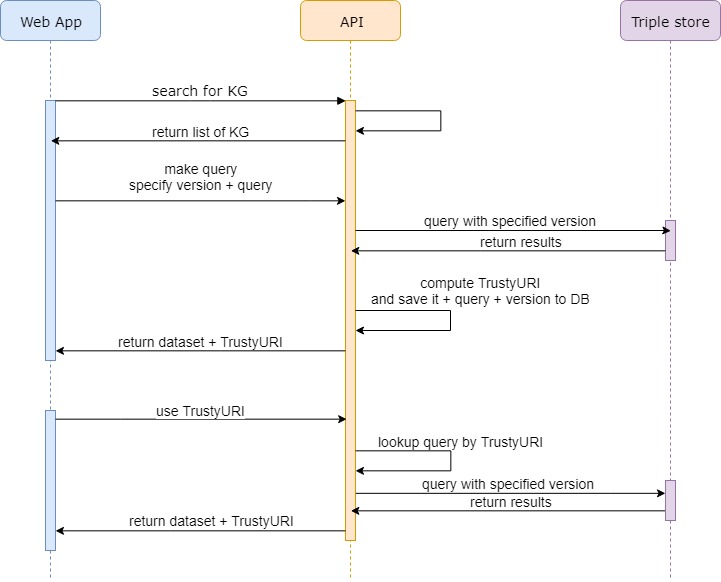}
    \caption{A sequence diagram to describe the workflow of the proposed design through the three main components: web app, API, and the triple store}
    \label{fig:SD}
\end{figure}

\medskip
\noindent\textbf{Features}
This application has several features, including the ability to search for KGs using their metadata (since we are providing access to multiple KGs), filter graph versions using date ranges and query for data from the selected version(s). Users will be able to download the desired data and the TrustedURI for that resource. This URI can be published and used later to let the user or anyone else access that same resource. Here we list each of these features and how we are planning to elaborate on them.

\medskip
\noindent\textbf{Searching for Datasets/KGs}. In this case, users need to specify which datasets they want to query from, this operation is done by providing the dataset metadata, such as the author name, and the language. The application will then show a list of matches and the user will select one of them. To keep track of the datasets, we provide the API endpoint with a dictionary database of metadata that takes care of finding matched datasets.

\medskip
\noindent\textbf{Version Selection}. Metadata about KGs include their versions which are retrieved from the Ostrich triple store. KG versions and their creation date are also stored in our metadata database to provide version filtering in the application.

\medskip
\noindent\textbf{Querying \& Creating Datasets}. After the user has selected a target version, he can start developing his query to create his desired dataset. After the query is done processing, the result is displayed and is available for download. Behind the scenes, the query, including the target version of the selected KG, is sent to the API, where it is passed on to the triple store for evaluation. The triple store will then return query results to the client.

\medskip
\noindent\textbf{Obtaining A Persistent Identifier}. The user has the option to create a unique and persistent identifier for each query that is made. This identifier is a Trusty URI and is generated using all information about the query including the extracted data and it serves as the unique pointer to the generated resource. This way, the user can also verify the result content that he received. This resource identifier can then be published and used to reproduce the same data results.

\begin{figure}
    \centering
    \includegraphics[scale=0.5]{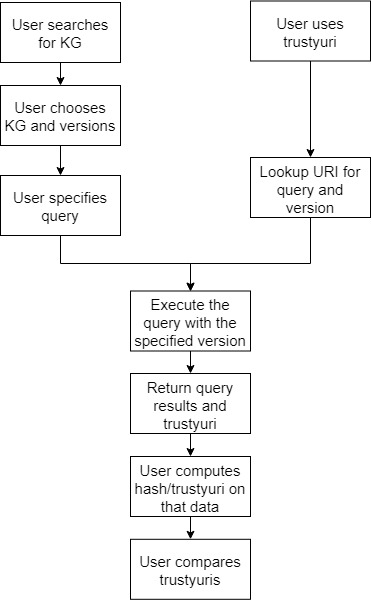}
    \caption{System flowchart}
    \label{fig:I1_flowchart}
\end{figure}

\begin{figure}
    \centering
    \includegraphics[scale=0.3]{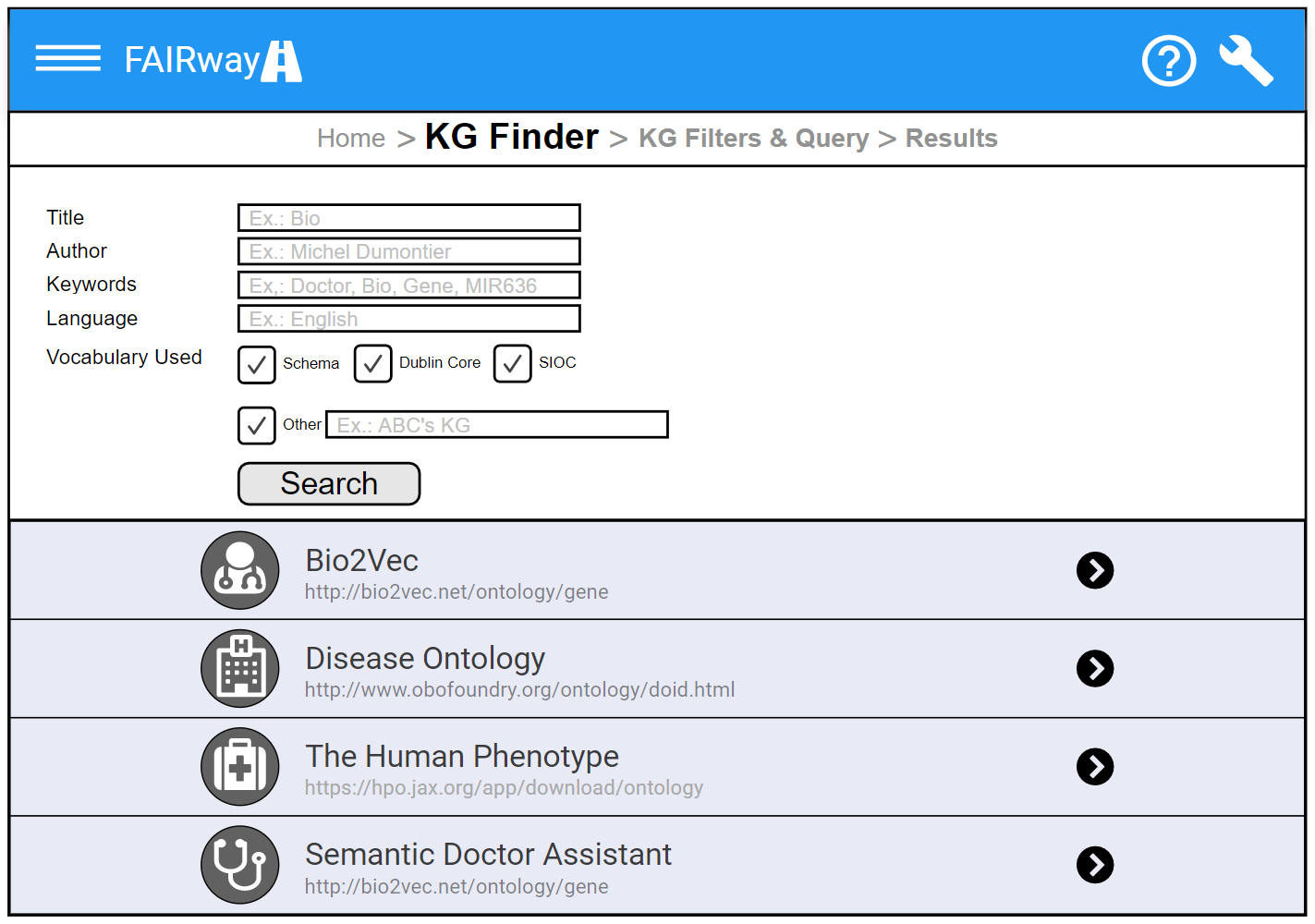}
    \caption{KG Finder Screen}
    \label{fig:I2_finder}
\end{figure}

\begin{figure}
    \centering
    \includegraphics[scale=0.3]{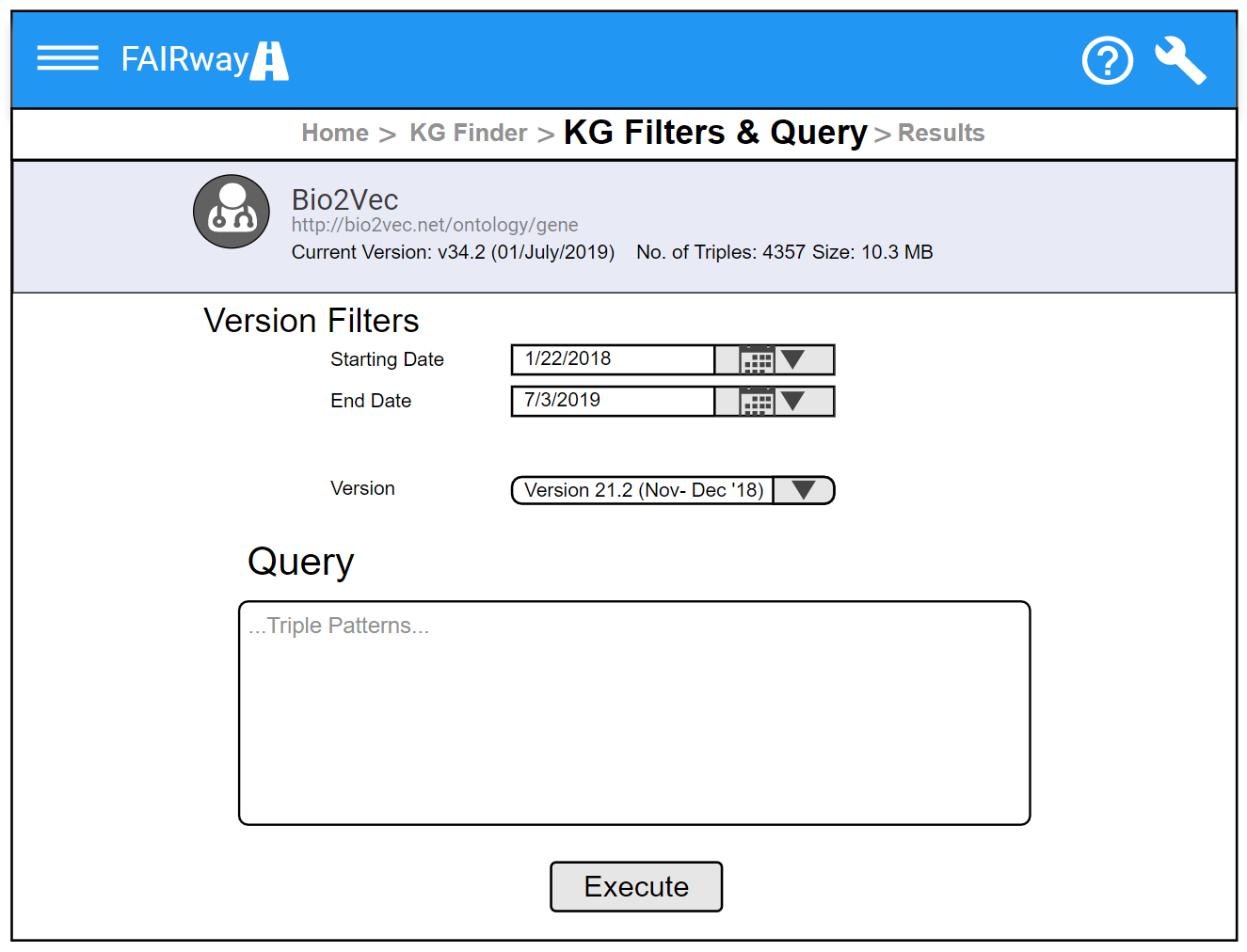}
    \caption{KG Filters and Query Screen}
    \label{fig:I3_filter}
\end{figure}

\begin{figure}
    \centering
    \includegraphics[scale=0.3]{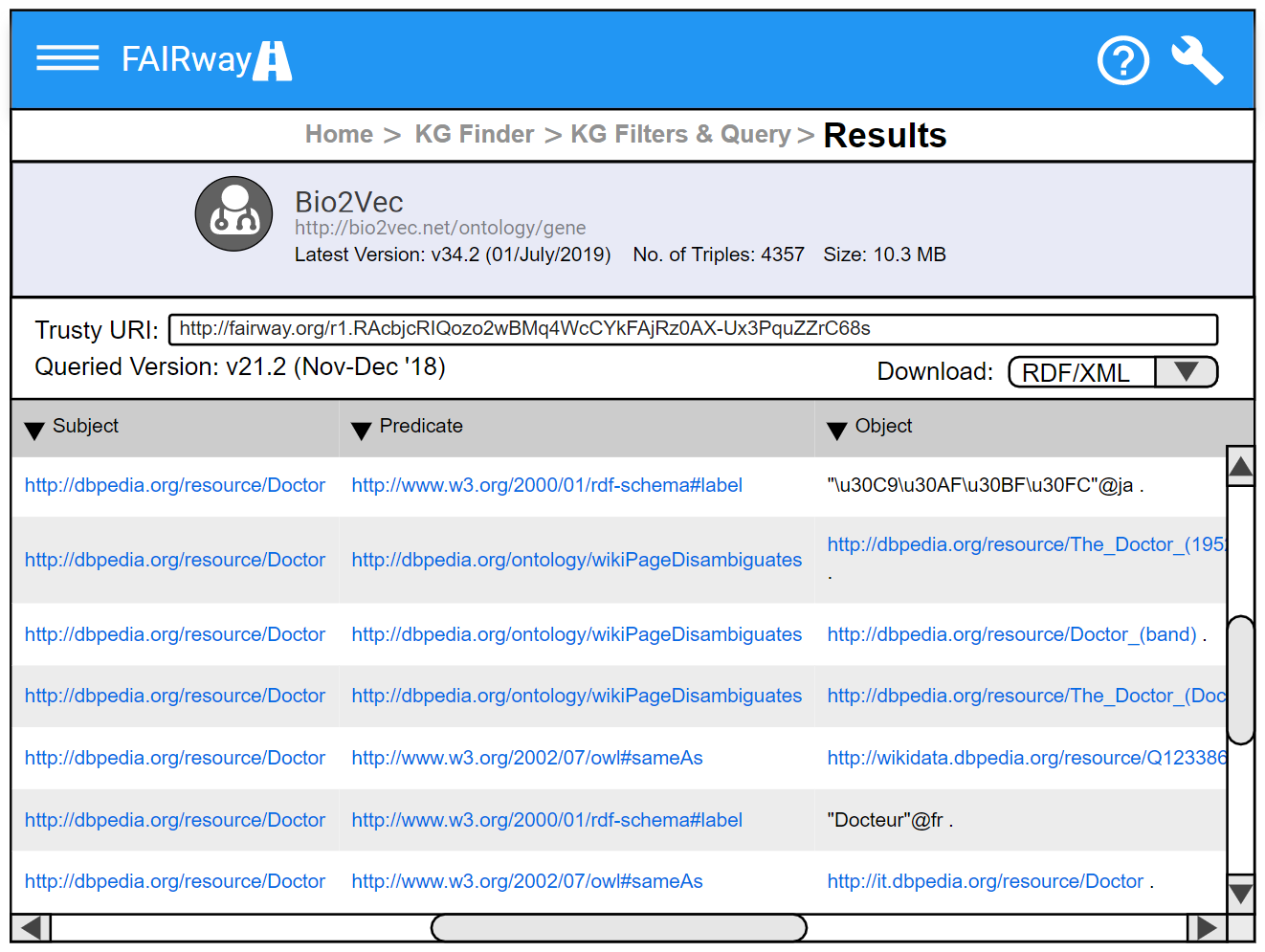}
    \caption{Results Screen}
    \label{fig:I4_results}
\end{figure}

\medskip
\noindent\textbf{User Interface}
Figure \ref{fig:I1_flowchart} represent the process to get a Trusty URI and Triples, starting from the desired version of the chosen KG and a specific query. The first screen in Figure \ref{fig:I2_finder} provides the user an interface with multiple field options to fill such as: Title, Author, Keywords, etc. to get input from the user and give back a list of KGs to select from. After the selection of the desired Knowledge Graph, the next screen is shown in Figure \ref{fig:I3_filter}, where users can submit a date range to get a list of KG versions. In the same interface, screen user can also provide a query and execute it to get results. The next screen in Figure \ref{fig:I4_results} is the Results window, where users can view the triples derived from given KG. The user has the ability to generate a Trusty URI for these results which is based on the actual results and metadata about the target KG. The user also has the option to download the triples in any of the provided formats (RDF/XML, Turtle, N-Triples, CSV). The download package contains 2 files, one with the triples of the results and another one with the metadata that identifies the original resource (KG data, version and the query used to get the result).



\section{Discussion}
\label{sec:discussion-slytherin}
For the main discussion, we want to assess our design against the use cases regarding versioned querying and persistent and unique identifiers of resources.

\medskip
\noindent\textbf{UC1 + UC2: Knowledge graph search and retrieval}. These use cases are fully covered by our API where we keep track of knowledge graphs in a separate database which contains all metadata about knowledge graphs including their Trusty URIs and graph versions. This allows us to effectively look for the desired datasets and select a version to query from.

\medskip
\noindent\textbf{UC3: Queries on specific versions of KGs that evolve over time}. This use case is mainly covered by the Ostrich triple store engine and also is made easier for users to query through the proposed user interface design. The persistent identifier is generated using the produced data and metadata of the original KG source, therefore, the user can also verify the retrieved content by computing the Trusty URI.

\medskip
\noindent\textbf{FAIR Assessment}
The FAIR principle asks that data resources need to meet certain requirements. In Table \ref{fair-table}, we assess our proposed service with respect to these principles and also provide the reasons for why and how does our service meets these requirements.


\section{Conclusions}
\label{sec:conclusion-slytherin}

In this paper, we have provided an answer to the research problem of identification and provision of volatile RDF resources by analyzing existing potential partial solutions and technologies and combining them to create a software architectural design that covers all use-cases that are implied from the research questions. We also provide the descriptions of these technologies and the reason why they are suitable to incorporate in our design. Then we delivered a FAIR assessment table where we include most of the FAIR principles requirements in our work. This means that our work is FAIR applicable. We are aware of several limitations of our work, including not incorporating a full SPARQL solution but only a triple pattern query, since the Ostrich does not support that feature. For future work we intend to build the system and make an evaluation of it on real data. We also expect to extend the functionalities of the system to support full SPARQL queries. In the end, we would also create an interface for the data providers for better data, knowledge graph management.

\newpage

\section*{Appendix}

\begin{table}[ht]
\begin{tabular}{|p{5.5cm}|l|p{5.5cm}|}
\hline
\multicolumn{3}{|l|}{\textbf{Findable}}\\ \hline
F1. (Meta)data are assigned a globally unique and persistent identifier                                        & Yes & We use TrustyURI to do that. \\ \hline
F2. Data are described with rich metadata (defined by R1 below)                                                & Yes & We store a local database containing the metadata of the datasets                                                         \\ \hline
F3. Metadata clearly and explicitly include the identifier of the data they describe.                         & Yes & TrustyURI associate a unique identifier for each datasets                                                                 \\ \hline
F4. (Meta)data are registered or indexed in a searchable resource                                              & Yes & We store them into a searchable database, which could be iterated to search in it.                                        \\ \hline
\multicolumn{3}{|l|}{\textbf{Accessible}}                                                                                                                                                                                                        \\ \hline
A1. (Meta)data are retrievable by their identifier using a standardised communications protocol                & Yes & We recommend the data providers to use a REST API implementation over a HTTPS protocol                                    \\ \hline
A1.1 The protocol is open, free, and universally implementable                                                 & Yes & It's in the REST API protocol specifications                                                                              \\ \hline
A1.2 The protocol allows for an authentication and authorisation procedure, where necessary                    & Yes & It's in the REST API protocol specifications                                                                              \\ \hline
A2. Metadata are accessible, even when the data are no longer available                                        & Yes & Metadata for each queries and targeted KG including its version are permanently stored in the database with Trusted URIs. \\ \hline
\multicolumn{3}{|l|}{\textbf{Interoperable}}                                                                                                                                                                                                     \\ \hline
I1. (Meta)data use a formal, accessible, shared, and broadly applicable language for knowledge representation. & Yes & We provide multiple formats for download, e.g. RDF/XML, Turtle, N-triples                                                                       \\ \hline
I2. (Meta)data use vocabularies that follow FAIR principles                                                    & -   & We recommend data providers to reuse these vocabularies, e.g. schema.org, dublin core.                                                                   \\ \hline
I3. (Meta)data include qualified references to other (meta)data                                                & Yes & We will use common code lists, e.g. for countries, currencies.                                                                          \\ \hline
\multicolumn{3}{|l|}{\textbf{Reusable}}                                                                                                                                                                                                          \\ \hline
R1. Meta(data) are richly described with a plurality of accurate and relevant attributes                       & Yes & We recommend our data providers to include as much description as possible                                                \\ \hline
R1.1. (Meta)data are released with a clear and accessible data usage license                                   & Yes & For our metadata about KGs, it will be a free license like CC0\footnote{\url{https://creativecommons.org/choose/zero/}}. For the actual data, we will use the same licenses of each dataset that were provided.                                                               \\ \hline
R1.2. (Meta)data are associated with detailed provenance                                                       & Yes & We recommend our data providers to include as much as possible provenance data for their datasets. Also all origins of versions of each KG will be preserved.                          \\ \hline
R1.3. (Meta)data meet domain-relevant community standards                                                      & Yes & We recommend our data providers to use the domain-relevant community standards. \\ \hline
\end{tabular}
\caption{Whether our proposed platform design handles each of the FAIR principles, and how it does that}
\label{fair-table}
\end{table}

\chapter{Supporting Interactive Updates of Knowledge Graphs}
\label{sec:ravenclaw}
\chapterauthor{Nacira Abbas, Francesca Giovannetti, Chuangtao Ma, Margherita Porena, Ariam Rivas, Soheil Roshankish, Sebastian Rudolph}

\section{Research Questions}
\label{sec:rq-ravenclaw}
\begin{itemize}
\item How to support the evolution and preservation of knowledge graphs in terms of consistency and integrity?
\item How to make the updating of knowledge graphs a supervised and easier  operation, so to facilitate knowledge workers to modify data?
\item How to prevent incomplete change requests?
\end{itemize}

\section{Evolution and Preservation of Knowledge Graphs}
\label{sec:def-ravenclaw}
The ongoing evolution and preservation of knowledge graphs (KGs) has become one of the most challenging tasks, thanks to the rapid and constant change of beliefs and knowledge linked to all kinds of domains.  Our perspective on this problem is to grant the preservation of the "well-formedness" of KGs, by making sure that their consistency and integrity is not compromised when changes are applied. An ontology is commonly called consistent if it is free of logical contradictions and called inconsistent if it violates the syntactic constraints of the language or knowledge modelling guidelines \cite{jaziri2019preventive}. More generally, the integrity of an ontology (or knowledge graph) refers to the fact that the information recorded at any given point in time reflects a meaningful state of the described domain.

Our main purpose is to assist "knowledge workers" -- who are not experts in logical modelling -- in the KG updating process, in an interactive way,  while ensuring the consistency and the integrity of the knowledge graph in the course of its evolution.

\section{Introduction}
\label{sec:intro-ravenclaw}
\noindent
Regular updates to knowledge graphs should be encouraged, in order to reflect the evolution of a domain, which is certainly desirable. Nevertheless, adding new statements or modifying existing ones is an error-prone operation which can lead to data inconsistency or violation of the integrity of the data, also because the change request could imply or require other changes that a non-expert may not conceive. The availability of user-friendly tools capable of supporting humans in the process of updating large knowledge graphs is important, especially when updates are performed by a knowledge worker (usually not specialized in ontology design) other than the knowledge engineer who originally developed the conceptual model informing the knowledge base. The Ontology Update Language (OUL) \cite{losch2009tempus} addresses this specific task, focusing on denying or allowing modifications (with respect of the ontology model and of its update specifications) and making clear to the knowledge worker every  consequences of the change request he made. 
What this language does not take into account is the possibility that --  depending on the cause or context of the update -- some change requests may require different change patterns,, between which the knowledge worker should be able to choose. Our proposal is to implement a language extension to make possible a different kind of interaction between the system and the knowledge worker updating the graph, when different change patterns are to be selected from  in the modification.

A possible scenario where this approach could be applied is the following: a library decides to dispose of a collection that is considered not important for the institution, and donates books to other libraries. This action may have several consequences on books inside the collection: the library could decide to dispose of all the books contained, but it also could decide to keep some books with particular topics that may be interesting for library users. The choice of the change pattern can’t be decided {\textit{a priori}}, but is something that must be considered in every concrete case, at the time the actual change request is issued.

\section{Related Work}
\label{sec:related-ravenclaw}
\begin{figure}[ht]
	\begin{center}
		\includegraphics[height=2.73in]{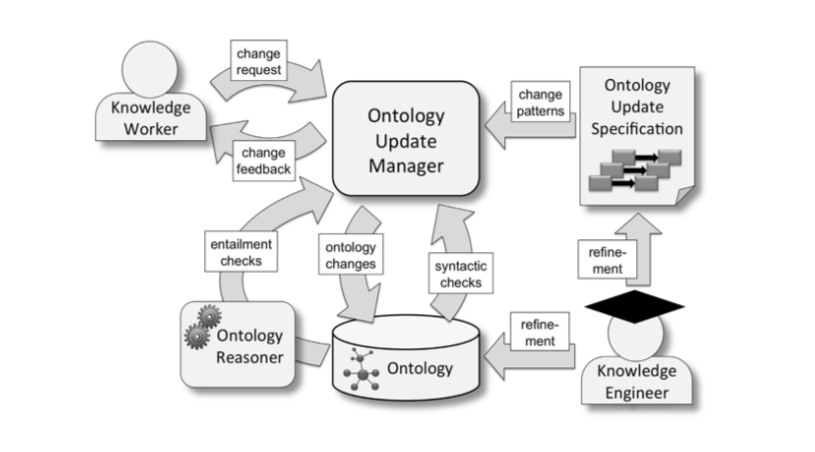}
		\caption{Ontology Update Architecture (L\"{o}sch et al. 2009).}
	\end{center}
\end{figure}

\noindent
There is a common scenario of requesting change and updating ontology due to continuous change in dynamic environments. Especially in a dynamic knowledge graph, the corresponding semantics also need to be changed when ontology need to be updated. The semantic change relates to the ontology evolution, change, management, and so forth. Ontology evolution is the process of an ontology changing in terms of size, content and management \cite{stavropoulos2016framework}. In addition, there are some available tools and systems have been developed for supporting ontology evolution in the KAON ontology engineering framework \cite{stojanovic2004methods}.

The Ontology Update Language\cite{losch2009tempus} was proposed with the aim to accommodate the dynamic changes and maintained ontology for keeping the ontology up-to-date. The main functions of OUL are capture frequent domain changes and facilitate the regular update. However, the manual ontology updating is a time-consuming work, hence the event-driven ontology update approach was proposed based on the ontology update language for improving the efficiency and supporting the maintaining of underlying ontology \cite{sangers2012event}.

On the basis of the aforementioned ontology update approaches and frameworks, there is no doubt that the efficiency of the ontology updating was greatly improved \cite{de2015event}. However, an increasing number of knowledge graphs are constructed \cite{paulheim2017knowledge}, which provide crucial knowledge support for decision-making. The semantic web of knowledge graph is provided by linked data \cite{data2015linked}, and the majority of knowledge graphs are created based on semi-structured knowledge, which create the huge challenge to preserve semantic consistency. To tackle above problems, the knowledge graph refinement \cite{pujara2013knowledge} is proposed. The assume of knowledge graph refinement is that there are knowledge graphs have been given, and some missing knowledge will be added and updated to remove the errors and inconsistency in knowledge graph \cite{akoglu2015graph}. The knowledge graph refinement attempt to identify erroneous pieces of information based on the inference of existing knowledge graph and adding missing knowledge. 

However, the problems of how to preserve the semantic consistency when it relates to user interactions and how to prevent the incomplete change request should be studied. Hence, this work focus on the above problems, the approach of extended ontology update language was proposed for supporting the interactive knowledge graph modification.


\section{Proposed approach}
\label{sec:approach-ravenclaw}
We propose an approach based on a language supporting interactive knowledge graph modification.
\noindent

In this section we propose an extension of the syntax of the Ontology Update Language, together with the precise description how ontology change requests are to be handled by the ontology management
component, including the interactions to be carried out (Alg.~\ref{alg:change_request}).

\begin{figure}
\small
\begin{verbatim}
CREATE CHANGEHANDLER <name> 
FOR <changerequest>
AS
  	[ IF <precondition>
 	 THEN ] <actions>
<changerequest> ::== add [unique] (<SPARQL>)
     | delete [unique] (<SPARQL>)
<precondition> ::== contains(<SPARQL>)	
     | entails(<SPARQL>)
     | entailsChanged(<SPARQL>)
     | (<precondition>)
     | <precondition> and <precondition> 
     | <precondition> or <precondition>
<actions> ::== [<action>]|<action><actions>
<action> ::== <SPARQL update>
     | for( <precondition> ) <actions> end; 
     | feedback(<text>)
     | applyRequest
     | <interaction>
<interaction>::== approval (<text>) <actions> end;
     | fixedSingleChoice (<text>) <selection> end;
     | fixedMultipleChoice (<text>) <selection> end;
     | unboundedSingleChoice (<text>) choose(<precondition>) 
        item(<text>) <actions> end;
     | unboundedMultipleChoice (<text>) choose(<precondition>) 
        item(<text>) <actions> end;
<selection>::== <action> or <actions>|<action> or <selection>

<SPARQL> ::== where clause of a SPARQL query
<SPARQL update> ::== a modify action (in SPARQL Update)
<text> ::== string (may contain SPARQL variables)
\end{verbatim}
\normalsize
\caption{Ontology Update Language syntax specification.}\label{fig:syntax}
\end{figure}

The implementation adds to OUL another type of action, allowing the interaction between the system and the knowledge worker. As mentioned in the previous section the knowledge worker did not have any options while he wanted to modified the knowledge graph. Therefore, in the current suggested structure different options will be offered by the system to the user. 

There are three different types of options. The first one is the {\textit{approval}} interaction, allowing user to approve or not changing in graph that are not mandatory. The second one is {\textit{fixedChoice}} group of intercaction, divided into {\textit{fixedSingleChoice}} and {\textit{fixedMultipleChoice}}. Here the interaction consists in choosing between options that are fixed, because their number is already decided by the ontology update specification.
The last one is the {\textit{unboundedChoice}}, (divided into {\textit{unboundedSingleChoice}} and {\textit{unboundedMultipleChoice)}}, used in case the modification implied a selection between variables that can't be numbered {\textit{a priori}}, but only thanks to a SPARQL.

\subsection{Algorithm}
\label{sec:algorithm-ravenclaw}
The algorithm for Change Request (Alg.~\ref{alg:change_request}) receive as input a change request specification 'US' and which operator to apply. The first step is to check the syntax and if exists in the ontology the given input.
The second step is find all relations that exist for that given input. To do that, we take into account a set of constraints that should not be removed for each predicate in the ontology. It facilitates the ontology management to the knowledge worker. The method {\textit{getChoose()}} is the user interaction where he has to select a list of candidates. With the list of candidates done, the action is executed and the ontology is updated.

\begin{algorithm}
	\caption{Processing of Change Request}
	\label{alg:change_request}
	\hspace*{\algorithmicindent} \textbf{Input:} \hspace*{\algorithmicindent}ontology O consisting of axioms\\
	\hspace*{\algorithmicindent}ontology update specification US treated as list of changehandlers\\
	\hspace*{\algorithmicindent}change request op(Ax) where op $\in$ \{add,del\} and $A_{x}$ is a set of axioms
resp. triples  \\
	\hspace*{\algorithmicindent} \textbf{Output:} Updated ontology O
	\begin{algorithmic}[1]
		\Procedure{Update ontology}{}
		\State $checkInputOntology(US,op)$
		\If{$existsInput(US)$}:
		\State $listCandidate$$\leftarrow$$findAllChange(US,op)$
		\State $toChange$$\leftarrow$$getChoose(listCandidate)$
		\State $executeAction(toChange,op)$
		\EndIf
		\EndProcedure
	\end{algorithmic}
\end{algorithm}

\section{Evaluation and Results: Use case/Proof of concept - Experiments}
\label{sec:evaluation-ravenclaw}
\begin{figure}[ht]
    \centering
    \includegraphics[scale=0.7]{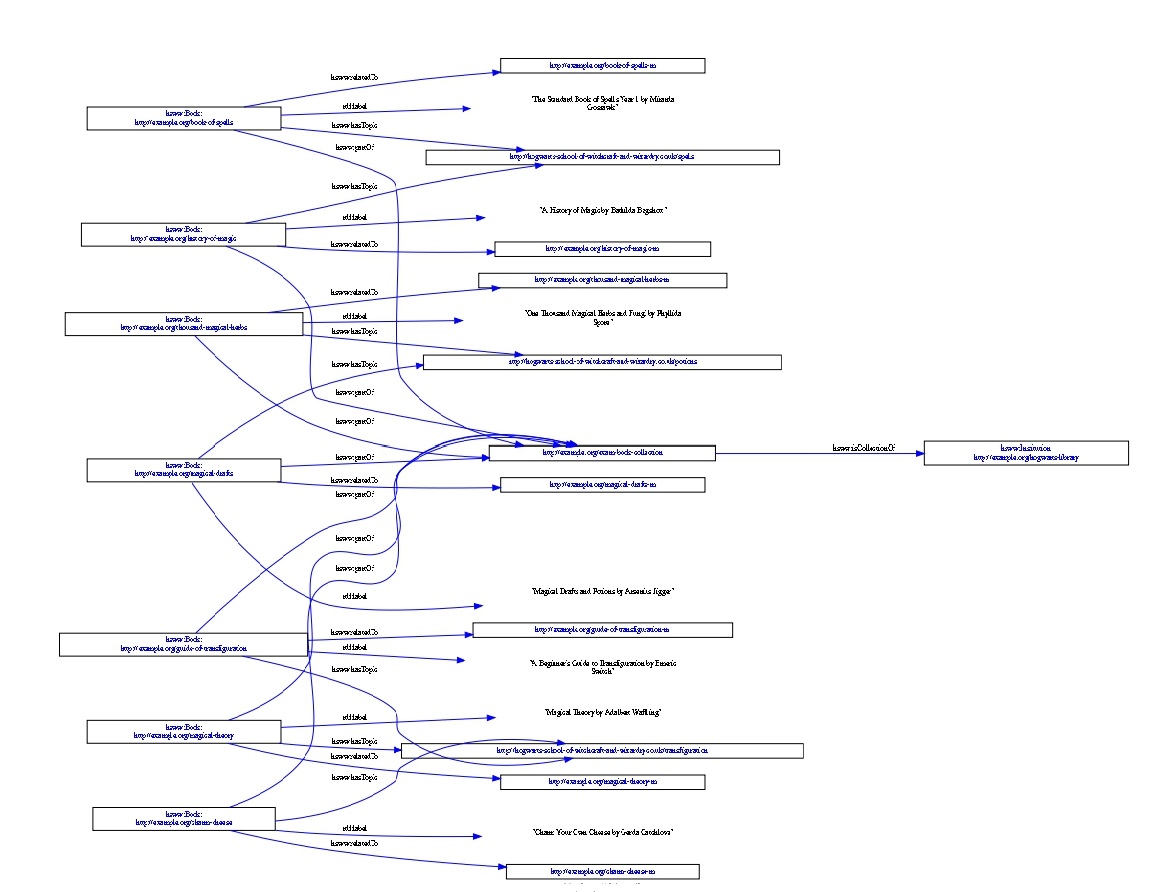}
    \caption{Hogwarts Library's knowledge graph}\label{fig:knowledgeGraph}
\end{figure}

\textit{}
In the current section we provide an example to show the concrete work flow. We start with the knowledge base from Fig. \ref{fig:ontology-ravenclaw}., represented in the knowledge graph in Fig. \ref{fig:knowledgeGraph}. 

\begin{figure}[H]
\begin{lstlisting}[basicstyle=\footnotesize\ttfamily]
@prefix ex: <http://example.org/> .
@prefix hsww: <http://hogwarts-school-of-witchcraft-and-wizardry.co.uk/> . 
@prefix rdf: <http://www.w3.org/1999/02/22-rdf-syntax-ns#> .
ex:exam-book-collection hsww:isCollectionOf ex:hogwarts-library .

ex:hogwarts-library rdf:type hsww:Institution .

ex:book-of-spells a hsww:Book ;
    rdf:label "The Standard Book of Spells Year 1 by Miranda Goshawk" ;
    hsww:partOf ex:exam-book-collection ;
    hsww:relatedTo ex:book-of-spells-m ;
    hsww:hasTopic hsww:spells .

ex:guide-of-transfiguration a hsww:Book ;
    rdf:label "A Beginner's Guide to Transfiguration by Emeric Switch" ;
    hsww:partOf ex:exam-book-collection ;
    hsww:relatedTo ex:guide-of-transfiguration-m ;
    hsww:hasTopic hsww:transfiguration .

ex:history-of-magic a hsww:Book ;
    rdf:label "A History of Magic by Bathilda Bagshott" ;
    hsww:partOf ex:exam-book-collection ;
    hsww:relatedTo ex:history-of-magic-m ; 
    hsww:hasTopic hsww:spells .

ex:magical-theory a hsww:Book ;
    rdf:label "Magical Theory by Adalbert Waffling" ;
    hsww:partOf ex:exam-book-collection ;
    hsww:relatedTo ex:magical-theory-m ;
    hsww:hasTopic hsww:transfiguration .

ex:thousand-magical-herbs a hsww:Book ;
    rdf:label "One Thousand Magical Herbs and Fungi by Phyllida Spore" ;
    hsww:partOf ex:exam-book-collection ;
    hsww:relatedTo ex:thousand-magical-herbs-m ; 
    hsww:hasTopic hsww:potions .

ex:magical-drafts a hsww:Book ;
    rdf:label "Magical Drafts and Potions by Arsenius Jigger" ;
    hsww:partOf ex:exam-book-collection ;
    hsww:relatedTo ex:magical-drafts-m ; 
    hsww:hasTopic hsww:potions .

ex:charm-cheese a hsww:Book ;
    rdf:label "Charm Your Own Cheese by Gerda Catchlove" ;
    hsww:partOf ex:exam-book-collection ;
    hsww:relatedTo ex:charm-cheese-m ; 
    hsww:hasTopic hsww:transfiguration .

\end{lstlisting}
\normalsize
\caption{Example knowledge base in Turtle.}\label{fig:ontology-ravenclaw}
\end{figure}

This knowledge graph contains the knowledge of the current status of our domain.


\begin{figure}[h!]
    \centering
    \begin{lstlisting}[captionpos=b,basicstyle=\ttfamily,frame=single]
CREATE CHANGEHANDLER deleteCollection
FOR del { ?x isACollectionIn ?y }
    AS applyRequest;
    unboundedMultipleChoice ("Choose obsolete topics in collection ?x !") 
       choose({ _:1 hsww:partOf   ?x . 
                 _:1 hsww:hasTopic ?topic .})
       item("?topic")
       for(contains({ ?book ?p            ?o . 
                        ?book hsww:partOf   ?x . 
                        ?book hsww:hasTopic ?topic .}))
          delete data { ?s  ?p  ?o . };  
          feedback("Deleted the following triple: ?s ?p ?o .");
       end;
    end;
    \end{lstlisting}
    \caption{Example ontology update specification}
    \label{fig:example}
    \end{figure}

In figure \ref{fig:example} we show an example of how to
ontology is updated. For the specific predicate (isACollectionIn), first  look for all triples with predicate partOf and show the different Topic
to the user. After that, there is an interaction with the user where he chooses the topic to remove. Based on example knowledge base, figure \ref{fig:ontology-ravenclaw} the knowledge worker want to remove this input: \textbf{\textit{ex:exam-book-collection isACollectionIn HogwartsLibrary}}. The output is to show all topics related to \textbf{\textit{ex:exam-book-collection}} to the user, one by one, so he chooses which topic to delete.
As a result of user interaction are removed all book with the topic selected by the user.

\noindent



\section{Conclusions and Future Work}
\label{sec:conclusions-ravenclaw}
\noindent
In this work, the ontology update language was extended for describing the interactions in dynamic knowledge graph. The algorithm of proposed approach was depicted, after that, the specified example was given base in turtle. In addition, the Hogwarts Library's knowledge graph was constructed, and a use case are analysed to examine the concrete workflow.
However, there are some possible research directions that could be investigated based on current work. 

\textbf{Develop the tool to support extended ontology update language.} It’s necessary to develop the user-friendly GUI (graphical user interface) tool supporting the swift respond of the dynamic ontology update based on current extended ontology update language.

\textbf{Extend the language for responding and supporting temporal change.} It’s unavoidable to receive and respond some temporal changes in dynamic knowledge graph. Hence, how to further extend the ontology update language for responding the temporal change should be explored.  

\textbf{Explore the framework and approach of autonomous ontology updating framework.} There is no doubt that the change of ontology will be more and more frequent, hence how to capture or predict the changes, how to autonomously update the ontology, how to identify and repair the errors and inconsistencies should be investigated.
\chapter{A human-in-the-loop framework to handle implicit bias in crowdsourced KGs}
\label{sec:klingon}
\chapterauthor{Alba Morales Tirado, Allard Oelen, Valentina Pasqual, Meilin Shi, Alessandro Umbrico, Weiqin Xu, Irene Celino}

\section{Research Questions}
\label{sec:rq-klingon}
Crowd-sourced Knowledge Graphs (KGs) may be biased: some biases can originate from factual errors, while others reflect different points of view. How to identify and measure biases in crowd-sourced KGs? And then, how to tell apart factual errors from different point of views? And how to put together all these steps contextualized in a human-in-the-loop framework?

\section{Knowledge Graphs Evolution and Preservation}
\label{sec:def-klingon}
Knowledge Graphs are built with the help of technological tools and the intervention of people. Few KGs are constructed without the involvement of humans. Generally speaking, the process of building a KG comprises the interaction of humans in two ways \cite{dagstuhl}: to acquire knowledge and verify its accuracy as the type of knowledge differs from human to human; to seek information with the help of applications built on top of KGs. As \cite{Demartini-2019} methodology, our focus is to investigate about how crowdsourcing can be used to understand contributor bias for controversial facts included into crowdsourced KGs. Moreover, we want to trace the provenance of crowdsourced fact checking and some additional information (preservation step) in order to update KG (evolution step). This process enables bias transparency rather than aiming at eliminating bias from KGs \cite{Demartini-2019}.

\section{Introduction and Problem Statement}
\label{sec:intro-klingon}


%
%
%
%

Knowledge Graphs have bias that must be properly managed to guarantee a correct evolution and preservation of the knowledge inside. KGs that pursue a collaborative approach ({\em crowdsourced}) to the creation of knowledge like e.g., Wikidata \cite{wikidata} enable the creation of high quality structured information and let semantics emerge in a bottom-up fashion \cite{dbpedia-2007,maedche-2001}.
However, reality is ambiguous and there can exist different opinions and perspectives due to cultural differences and heritages. This means that contributors may introduce bias into the resulting knowledge. Human contributors to crowdsourced KGs can be biased by their personal point of view and can therefore enter {\em implicitly} biased information to the KGs. The implicit nature of the bias means that the contributions are unconsciously and unintentionally adding a bias to the KGs. A meaningful example of such situations is Catalonia being part of Spain or being an independent country, as shown in \cite{Demartini-2019,Bolukbasi-2016}, where contributors may easily bring their own bias answering this question.

{\em Implicit bias} can affect the correctness of knowledge and they must be properly managed within the evolution of KGs \cite{dumitrache-2019,janowicz-2018,Dumitrache-2018}. In some cases however, biases reflect {\em knowledge diversity} that must be preserved. 
There is a number of issues to deal with in order to properly manage bias with respect to knowledge graph evolution and preservation. 

\begin{itemize}
    \item It is necessary to let {\em implicit bias} emerge and make them {\em recognizable};
    \item Identified bias must be evaluated so that to decide whether they represent incorrect information to {\em fix} ({\em critical bias}) or knowledge diversity to preserve;
    \item Decide how to handle critical bias and how to {\em refine} a KGs to remove the inconsistency
\end{itemize}

%
Knowledge graphs are created in socio-technical systems and in most of the cases correlated aspects like e.g., knowledge acquisition and knowledge ``maintenance" cannot be fully automated. As pointed out in research of \cite{dagstuhl}, a proper exploitation of the human-factor within the knowledge graph management life cycle, has several benefits. On the one hand, a formal characterization of user profiles with respect to a number of relevant features like provenance or educational background, could be useful to support the identification and implicit bias. On the other hand, the ``enrollment" of users in the evaluation of identified implicit bias could be useful to determine which ones are critical and how to solve them.

In this context, our aim is to propose an iterative framework to address the three points cited above and therefore  formalize and capture bias, communicate them and act on the KG to solve inconsistencies caused by ``critical bias". The envisaged approach strongly rely on the contribution of humans. Specifically, we propose to distinguish between two types of end-users: (i) {\em collectors} or {\em contributors} that like in any other crowdsourced KGs are common people taking part in knowledge acquisition; (ii) {\em reviewers} that are selected users that are in charge of evaluating implicit bias identified by the framework, and proposing a solution for them. So our contribution aims at pursuing a twofold human-in-the-loop iterative verification and refinement loop to deal with consistent evolution of KGs. Automated steps of the process flow support users by ranking bias to identify critical ones and then aggregate reviewers' feedback to refine the KG.


\section{Related Work}
\label{sec:related-klingon}


\noindent From the state of the art we analysed literature related to approaches to human-in-the-loop, KG refinement, different kinds of human bias detection, analysis and measurement in KG and other technologies, preservation of data provenance in KG.

\begin{itemize}

    \item Many are the approaches to Human-in-the-loop, as human computation \cite{law-2011}, citizen science \cite{irwin-2002} and crowdsourcing \cite{Howe-2008}. Those are the techniques that we will use in our solution whenever the human contribution is needed.
    
    \item \cite{Paulheim-2016} provides the definition of KG refinement e.g., by adding missing knowledge or identifying and removing errors. Our work focuses on the latter case. Removing biases can be considered a case of KG refinement, we intend to address it through both automatic method and human involvement.  
    
    \item \cite{janowicz-2018} emphasizes debiasing knowledge graphs will become a pressing issue as their usage is widening. This aspect is actually a real common ground with our research. Additionally, \cite{janowicz-2018} focuses on different kinds of biases (i.e schema level bias), while our perspective is to handle facts level biases inside a structured work flow aimed at preservation and evolution of KG.  
    
    \item In \cite{Eickhoff-2018} cognitive bias are considered a source for noisy label submissions, mentioning that people can create subjective social realities. Three tests are set up on three large corpus in order to show people's interpretation ambiguity effect. As argued in \cite{Eickhoff-2018}, biased label collection can have significant influence on label quality, retrieval system evaluation and ranker training, underlying the need to carefully detect implicit bias in order to avoid label degradation. Our research follows this path, but aims to be applied to KG.
    
    \item As already mentioned, \cite{Demartini-2019} is the actual starting point of our research, from the perspective of modelling a debiasing methodology in KG context. 

    \item CrowdTruth metrics  \cite{Dumitrache-2018} capture and interpret inter-annotator disagreement in crowdsourcing. The CrowdTruth metrics model assumes the concept of inter-dependency between crowdsourcing system's three main components (worker, input data, and annotation). Metrics' goal is to capture the degree of ambiguity in each of these. The shared idea between our proposal and Crowdtruth is about the attempt to rank information ambiguities. Some similarities emerge on the level of shared methodology with respect to our "preservation" step. Additionally their research relies upon unstructured textual data and images, so managing KGs evolution is not a CrowdTruth's concerned topic. We aim to build on CrowdTruth metrics to have a quantitative estimation of bias and then, to make this information explicit in the KG.

    \item Our attention is also pointed to user contribution profiling. As data provenance becomes a central issue on the social web, \cite{Janowicz-2009} points up that absence of provenance may render trust ratings useless. Despite the different application scenarios (social bookmarking systems and KGs), the provenance matter is central point in both cases. Part of the knowledge graph preservation is in recording and then displaying the composition of the contributing crowd.

\end{itemize} 


\section{Methodology}
\label{sec:approach-klingon}
%
%
%

\begin{figure}[ht]
    \centering
    \includegraphics[width=0.5\textwidth]{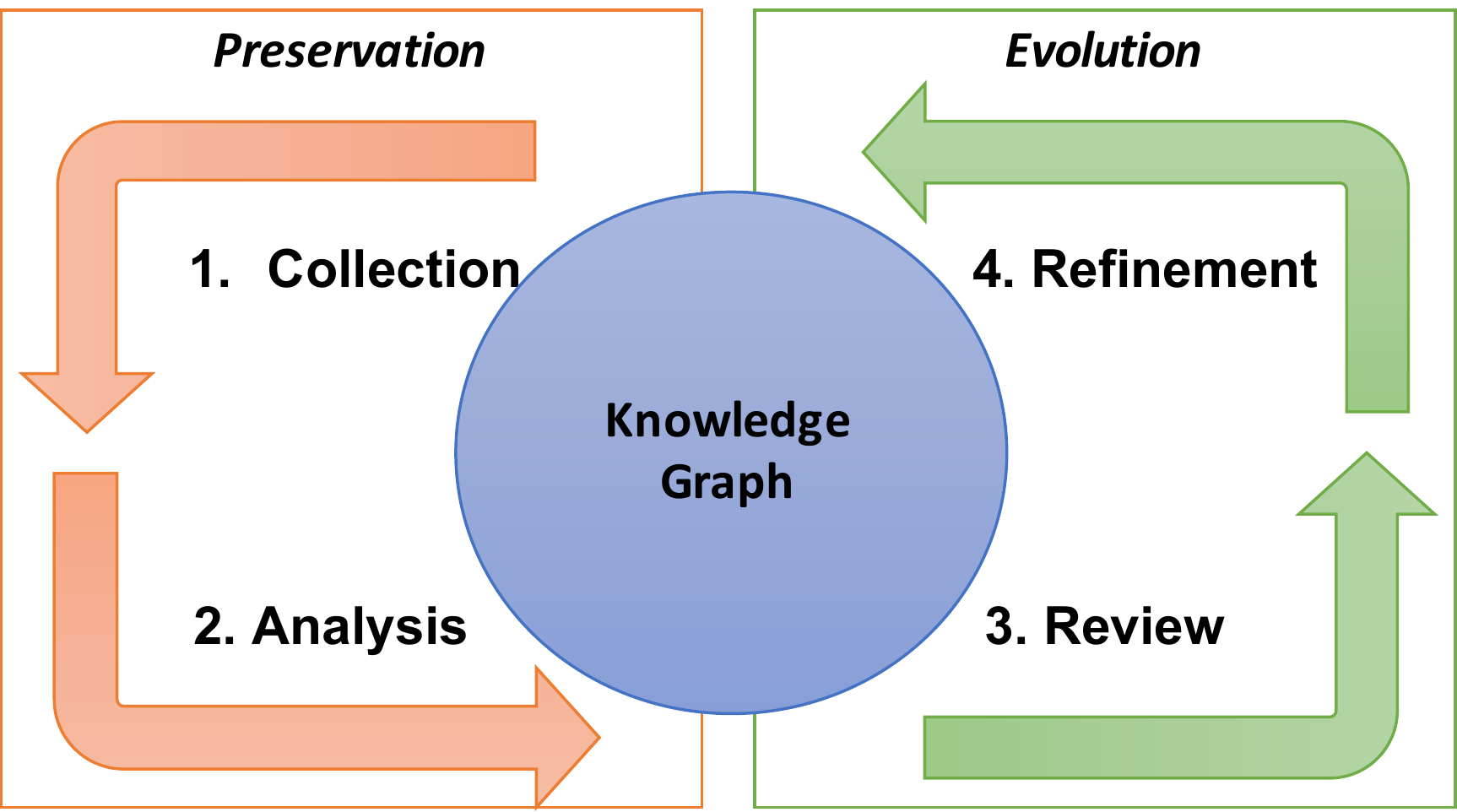}
    \caption{Process overview. Orange side of the process concerns with the preservation of KGs while the green side with the evolution. Both processes could involve either not-automated or automated steps that do or do not require human intervention.}
    \label{fig:process}
\end{figure}

The core idea is to involve humans within the evolution and preservation of KGs while managing bias that appear in cultural or geographical context. Similarly to other research fields like e.g., cognitive sciences and Theory of Mind \cite{thagard2005mind,varela2017embodied}, our aim is to lay the foundation for understanding and predicting {\em mental models} characterizing opinions that (classes) of people may have on different topics, and identify which opinions to preserve and which ones to correct. Our contribution is the formalisation of a general workflow for identifying, classifying and solving bias in KGs, thus iteratively refine KGs preserving and allowing knowledge to evolve correctly. 
Figure \ref{fig:process} shows the structure of the approach we envisage. The objective is to manage bias accordingly by involving the contribution of humans within the process. Our proposed workflow consists of four steps that can be either made by humans or automated.

\subsection{Collecting biases}
An existing set of triples is used as a starting point for identifying bias. In theory, any statement inside the knowledge base can be biased. However, some subjects are more controversial than others, therefore the statements that represent these topics are more likely to contain bias. In order to make a start on bias identification, existing lists of controversial topics can be used, e.g.\footnote{\url{https://en.wikipedia.org/wiki/Wikipedia:List_of_controversial_issues}}. After selecting a set of controversial topics, the related triples for these topics should be fetched. The next step in the process is to capture the bias for these triples. The bias is collected in a crowd-sourced manner. In this crowd-sourced setup, users are able to either agree or disagree with a certain statement. For each user part of the crowd-sourcing effort, additional data is recorded. This can include the user's date-of-birth, cultural background or educational background. As soon as a statement is rated, the user data is stored as provenance data for the statement. 

\begin{figure}[hb]
    \centering
    \includegraphics[width=0.60\textwidth]{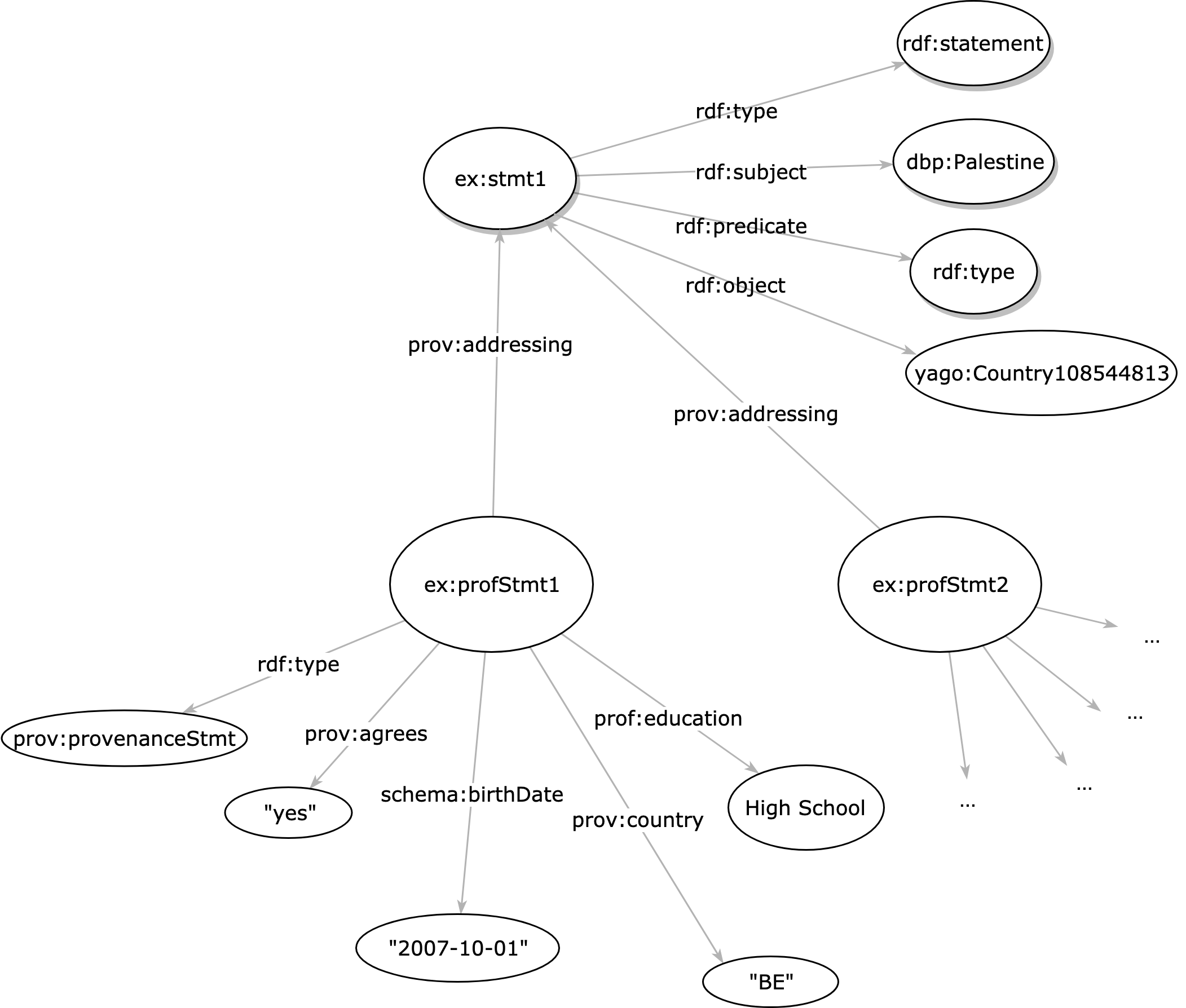}
    \caption{Data structure for storing user provenance data on statement level}
    \label{fig:data-structure}
\end{figure}

In order to store provenance data, the concept of reification in RDF is used. Although there is a variety of different provenance methods in RDF \cite{hernandez2015reifying}, the implementation of standard reification is straightforward and therefore fits the purpose of our methodology. Each individual rating is stored as provenance data on triple level. This results in a knowledge base that provides not only statement, but also information on the validity of those statements. While some statements might be either right or wrong, for other statements this might not be the case. In Figure \ref{fig:data-structure} the data structure is visualized. In this example, a user has rated a statement, this rating together with some user data is stored on statement level. When other users also rate the same statement, a new provenance statement is added.

\subsection{Analysis}
Once the crowd-sourced biases are identified, the next step is to rank which bias triples we would like to treat first and to transfer these triples to the next step; this ranking includes several factors which will be introduced in the following. This second step will be an automatic procedure rather than a manual work. In order to rank each biased statement, a set of metrics needs to be defined that measures the level of controversy.

\subsection{Review}
After capturing and processing the crowd-sourced biases, the goal of this step is to review the biased statements and decide how to present it to the end-users. Based on the ranking information from the previous step, 
we include human-in-the-loop and select reviewers based on topics. We would ask people to review and evaluate the statement whether they think the identified bias should be corrected or not. Given the different background information of users contributing to the statement, it is important to have human at this stage to evaluate the statement and further improve the crowd-sourced KGs in the next step.

\subsection{Refinement}
The last step is the refinement of KGs. Once the results from the collection and processing show that the crowd-sourced bias is solvable, we propose to refine and update the existing KGs with provenance data. In  case  the bias is not solvable, we propose to store additional information in the knowledge base. Maintaining the agreement and disagreement of the controversial statements at the same time in KGs provides the end-users with the diverse information, which better suits the end-user's background.




\section{Experimental Protocol Evaluation}
\label{sec:evaluation-klingon}
In this section we present the experimental protocol to evaluate the envisaged methodology.
According to Figure \ref{fig:process}, the experiment is structured into two parts that follow the steps concerning knowledge preservation and the steps concerning knowledge evolution.
%

%

%
%
%
\subsection{Collection} 
To carry out the evaluation, we checked the list of controversial issues from Wikipedia\footnote{\url{https://en.wikipedia.org/wiki/Wikipedia:List_of_controversial_issues}}. 
The subject ``Israeli-Palestinian conflict and all related issues" is selected, corresponding to the triple\footnote{http://dbpedia.org/page/Palestine} \textit{dbp:Palestine} \textit{rdf:type}\textit{yago:Country108544813}. which  is a potential biased triple. Given a KG containing this triple, we design a ``crowd-sourcing" campaign to discover if this is a biased statement.

\begin{figure}[ht]
    \centering
    \includegraphics[width=0.8\textwidth]{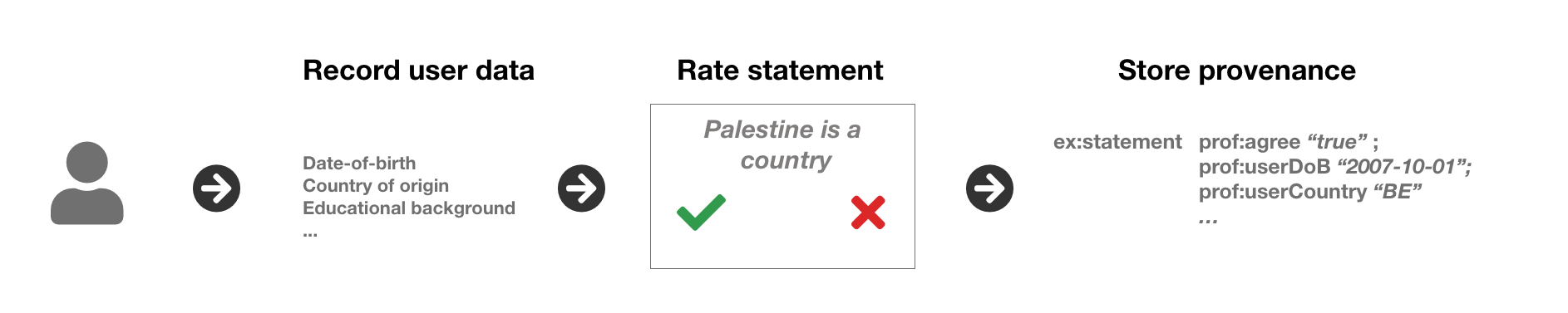}
    \caption{Profiling end-users through provenance metadata and collecting opinions}
    \label{fig:collecting-bias}
\end{figure}

We plan to do this by showing the triple with different visualizations and asking if the statement is true or not.
An example is showing the question ``is Palestine an independent country?" and then gather "Yes/No" response. We will select collectors according to their nationality. For example a possible configuration of collectors could be half  from Arabic countries and half from ``other countries". Metadata about collectors' provenance and their responses are stored to enrich the KG.
The pipeline in Figure \ref{fig:collecting-bias} describes the procedure followed to store collectors' metadata about provenance and gather opinions about the selected topic.

\subsection{Analysis}
We plan to analyze individual statements collected to identify potential bias. To do so, we will use a number of metrics to evaluate the level of bias.
The {\em Worker-Media Unit Agreement} (WUA) metric used in CrowdTruth \cite{Dumitrache-2018} is useful for this purpose; we will investigate for example if other metrics estimating the {\em entropy} of the knowledge included in the contributions collected in the previous step.
In general, if the resulting metric value is high, it is a sign of potential bias.
In the case of the Palestine statement for example we expect this evaluation metric to be high.
The result of this analysis will be integrated into the KG as additional metadata for the statement.

\subsection{Review}
\begin{figure}[ht]
    \centering
    \includegraphics[width=.6\textwidth]{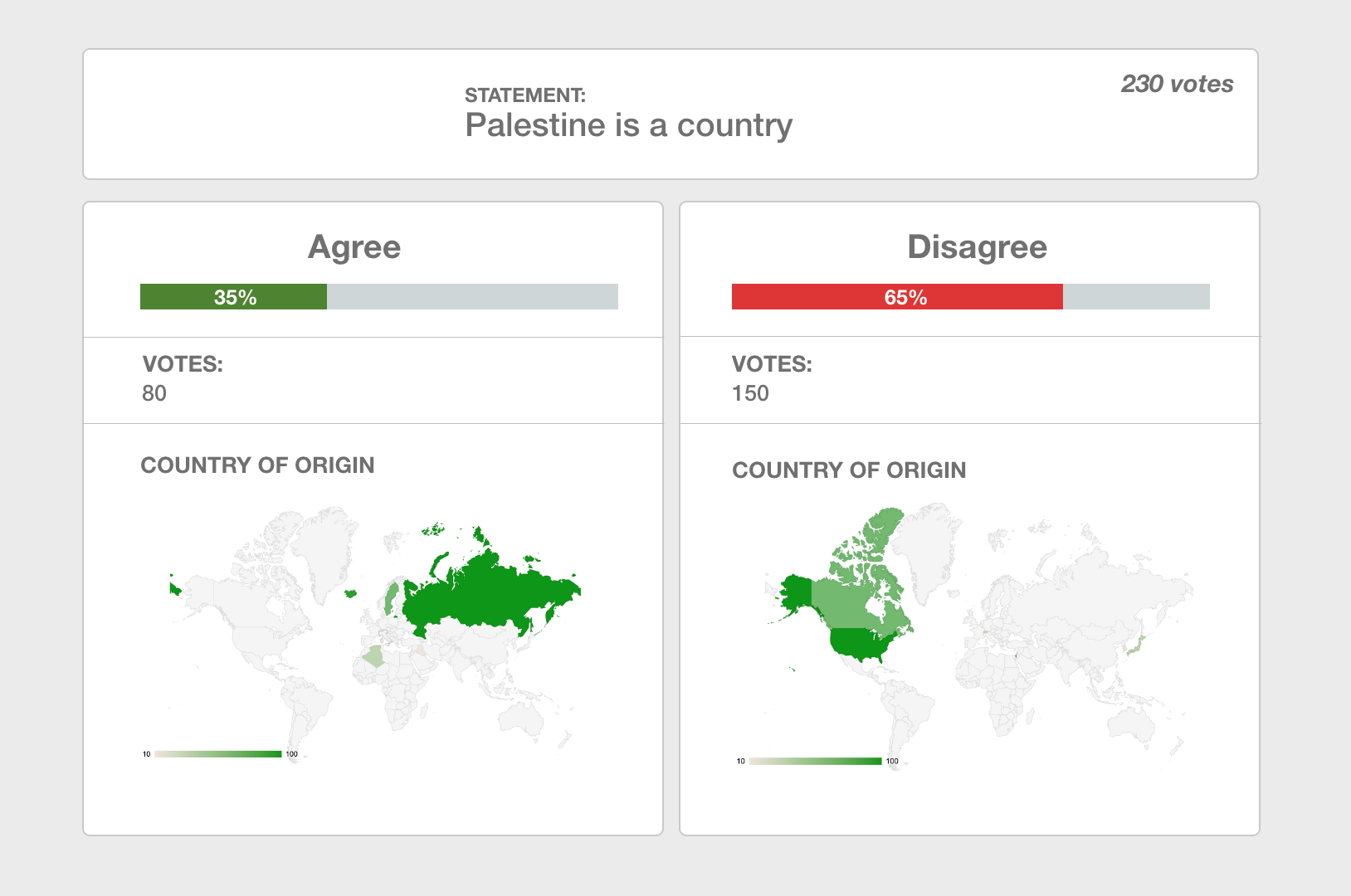}
    \caption{Example of Palestine statement provenance visualization}
    \label{fig:example-visualization}
\end{figure}

Statements with a high metric value are shown to reviewers in another crowdsourcing cycle. Different visualizations  are provided to allow reviews to see metadata associated to the triples in the KG.
Figure \ref{fig:example-visualization} shows a possible visualization of provenance metadata associated to the Palestine statement. In this example, 35\% agrees on the statement, the other 65\% disagrees. In addition to the percentages, the absolute amount of votes and the countries of respondents are displayed.

Reviewers evaluate if the statement is a solvable bias by answering the question ``is this bias solvable?". According to their response, the KG is refined by adding metadata indicating if the statement is a solvable bias or not. 

\subsection{Refinement}
Not all biased statements can be solved. In this case, the statement ``Palestine is a country" is likely not to be solvable, since some countries do consider it as a country and others not (and we expect the reviewers in the previous step to acknowledge this difference in points of view). No refinement actions are taken in this case. In the case of solvable bias, the solutions proposed by reviewers are aggregated together. Solutions can be new links to add to the KG or existing links to be removed from the KG.
According to the aggregation results, the solution mostly proposed by reviewers is actually propagated into the KG for refinement.

\section{Discussion and Conclusions}
\label{sec:conclusions-klingon}

\noindent
In this technical report, we proposed a four-step workflow to collect, analyze, review and refine the biases in crowd-sourced Knowledge Graphs. This workflow allows us, after every step, to add back the results to the KG and improve it while adding extra information to preserve the KG. 

We presented an experimental protocol evaluation with the ``Palestine is a country" statement and how this kind of unsolvable biased statement could be represented in the KGs. The existing KGs were improved for the management of bias by keeping track of provenance information and background knowledge, e.g. cultural and religious background, age, educational level etc. of users who contribute to the crowd-sourced knowledge graphs. The provenance information collected is then analysed to identify biases in existing KG. This work also describe how involving human-in-the-loop factors throughout the process helps to identify and improve the presentation of biases in KGs.

Further directions for this work include the experimental evaluation with potential biased statements in KGs and how end-users review the statements. Metrics on how to refine and update the crowd-sourced knowledge graphs are also one of the directions of future work. Also, a method to identify which part of the KG could go through this iterative process and how many iterations could be sufficient to control bias.



\chapter*{Acknowledgement}
We would like to thank everyone who contributed to the organisation of ISWS, the students who are its soul and motivating engine, and the sponsors. 

Please visit \texttt{\url{http://www.semanticwebschool.org}}

\printbibliography
\end{document}